\documentclass[11pt,twoside]{article}

\usepackage{geometry}

\usepackage[linesnumbered, algoruled, boxed, lined]{algorithm2e}
\usepackage{amsmath,amssymb,fullpage,graphicx,mathtools,amsthm,enumitem,subfig,dsfont,xspace}
\usepackage{hyperref,cleveref}
\usepackage{comment}
\usepackage[dvipsnames]{xcolor}

\newtheorem{theorem}{Theorem}
\newtheorem{proposition}{Proposition}
\newtheorem{lemma}{Lemma}
\newtheorem{corollary}{Corollary}

\newcommand{\Expect}{\mathbb{E}}
\newcommand{\Prob}{\mathbb{P}}
\newcommand{\sample}{\sim}
\newcommand{\Uniform}{\textrm{Unif}}
\let\uniform\Uniform

\newcommand{\reals}{\mathbb{R}}
\newcommand{\naturals}{\mathbb{N}}

\newcommand{\given}{\mid}

\newcommand{\defn}{:=}
\newcommand{\setcomplement}[1]{\overline{#1}}

\DeclareMathOperator*{\argmin}{argmin}

\DeclarePairedDelimiter\abs{\lvert}{\rvert}
\DeclarePairedDelimiter\norm{\lVert}{\rVert}
\DeclarePairedDelimiter\ceil{\lceil}{\rceil}
\DeclarePairedDelimiter\floor{\lfloor}{\rfloor}

\newcommand{\normtwo}[1]{\norm*{#1}_2}
\newcommand{\normone}[1]{\norm*{#1}_1}

\newcommand{\indicator}{\mathds{1}}

\newcommand{\intersect}{\cap}

\newcommand{\stepone}{\text{(i)}\xspace}
\newcommand{\steptwo}{\text{(ii)}\xspace}
\newcommand{\stepthree}{\text{(iii)}\xspace}

\newcommand*\dd{\mathop{}\!\mathrm{d}}
\newcommand{\textth}{\textrm{th}}

\newcommand{\lessorder}{\lesssim}
\newcommand{\gtrorder}{\gtrsim}

\newcommand{\worker}{worker\xspace}
\newcommand{\workers}{workers\xspace}
\newcommand{\Workers}{Workers\xspace}

\newcommand{\pos}{t}
\newcommand{\posalt}{\pos'}
\newcommand{\numitems}{n}
\newcommand{\idxitem}{i}
\newcommand{\idxitemalt}{{\idxitem'}}
\newcommand{\idxworker}{m}

\newcommand{\report}{y}
\newcommand{\setperms}{\Pi}

\newcommand{\rankrel}{r}
\newcommand{\rankrelalt}{\rankrel'}
\newcommand{\rankreltrue}{\rankrel^*}

\newcommand{\param}{x}

\newcommand{\const}{c}
\newcommand{\Const}{C}
\newcommand{\noise}{\epsilon}

\newcommand{\textlow}{\textrm{L}}
\newcommand{\texthigh}{\textrm{H}}
\newcommand{\set}{S}
\newcommand{\setlow}{\set_\textlow}
\newcommand{\sethigh}{\set_\texthigh}

\newcommand{\estscore}{\widehat{\tau}}
\newcommand{\rankrand}{V}

\newcommand{\losssf}{d_{\textrm{SF}}}
\newcommand{\losskt}{d_{\textrm{KT}}}
\newcommand{\lossentryat}[1]{d_{#1}}

\newcommand{\invperm}{\gamma}
\newcommand{\invpermtrue}{\invperm^*}

\newcommand{\invpermrestrict}[1]{\invperm_{[#1]}}
\newcommand{\estinvperm}{\widehat{\invperm}}
\newcommand{\estinvpermrestrict}[1]{\estinvperm_{[#1]}}
\newcommand{\estinvpermls}{\widehat{\invperm}_{\textrm{LS}}}
\newcommand{\estinvperminduced}{\widehat{\invperm}_0}

\newcommand{\noiseunifparam}{\delta}

\newcommand{\mapreltoabs}{\rho}

\newcommand{\arr}{a}
\newcommand{\estrankrel}{\widehat{\rankrel}}

\newcommand{\event}{\mathcal{E}}
\newcommand{\posplusone}{\pos'}

\newcommand{\identityrank}{\textrm{id}}
\newcommand{\obj}{v}
\newcommand{\objpartial}{\overline{\obj}}
\newcommand{\objtrue}{\obj^*}
\newcommand{\objest}{\widehat{\obj}}
\newcommand{\ceilnoiseposplueone}{\ceil*{\noiseunifparam \posplusone}}

\newcommand{\invpermtruerestrict}[1]{\invpermtrue_{[#1]}}

\newcommand{\numitemsfourth}{m}

\newcommand{\sourcerand}{\xi}
\newcommand{\classdet}{\mathcal{C}_\textrm{det}}
\newcommand{\setsource}{\Xi}

\newcommand{\vardiff}[2]{T_{#1}(#2)} 

\newcommand{\varinsert}{X}
\newcommand{\varinsertpos}{\varinsert^+}
\newcommand{\numitemslb}{\numitems_0}

\newcommand{\rankrelminusonetrue}{\widetilde{\rankrel}^*}

\newcommand{\funcrank}{\phi}

\newcommand{\term}{T}
\newcommand{\lossinvvec}{d_\rankrel}
\newcommand{\rankrelvec}{\boldsymbol{\rankrel}}

\newcommand{\prior}{\numitems_0}

\let\citep\cite
\let\citet\cite
\let\citealp\cite

\begin{document}

\begin{center}

  {\bf{\LARGE{Modeling and Correcting Bias in Sequential Evaluation}}}

\vspace*{.2in}

{\large{
\begin{tabular}{ccc}
Jingyan Wang$^{\dagger}$, Ashwin Pananjady$^{\dagger, \star}$
\end{tabular}
}}
\vspace*{.2in}

\begin{tabular}{c}
Schools of Industrial and Systems Engineering$^\dagger$ and
Electrical and Computer Engineering$^\star$ \\
Georgia Institute of Technology
\end{tabular}

\vspace*{.2in}

July 2022; \quad \quad Revised: November 2023

\vspace*{.2in}

\begin{abstract}
We consider the problem of sequential evaluation, in which an evaluator observes candidates in a sequence and assigns scores to these candidates in an online, irrevocable fashion. Motivated by the psychology literature that has studied sequential bias in such settings---namely, dependencies between the evaluation outcome and the order in which the candidates appear---we propose a natural model for the evaluator's rating process that captures the lack of calibration inherent to such a task. We conduct crowdsourcing experiments to demonstrate various facets of our model. We then proceed to study how to correct sequential bias under our model by posing this as a statistical inference problem. We propose a near-linear time, online algorithm for this task and prove guarantees in terms of two canonical ranking metrics. We also prove that our algorithm is information theoretically optimal, by establishing matching lower bounds in both metrics. Finally, we perform a host of numerical experiments to show that our algorithm often outperforms the de facto method of using the rankings induced by the reported scores, both in simulation and on the crowdsourcing data that we collected.

\end{abstract}
\end{center}

\section{Introduction}\label{sec:intro}
Consider your favorite sports or performance competition in which judges evaluate candidates; in keeping with a topical theme, say this is a figure skating competition at the 2022 Winter Olympics. A setting such as this one has some signature characteristics: Candidates (also called items in the sequel) perform one by one, in some predetermined sequence. A panel of judges (also called evaluators in the sequel) scores each candidate\footnote{In the sequel, we use the terms judge/evaluator interchangeably, and the terms candidate/item interchangeably.} 
in some quantitative fashion based on the performance.
These scores to each candidate are finalized before the next candidate performs. Similar characteristics are found in other settings besides sports, such as court decisions~\citep{chen2016gambler} and various other competitions (including music~\cite{flores1996elisabeth,ginsburgh2003musical} and Idol series contests~\citep{page2010idol}).

It is well known that \emph{sequential bias} is a hallmark of such settings, by which the evaluation of a candidate depends intricately on the position within the sequence that the candidate arrives. Following the example of sports competitions, sequential bias has been empirically observed in figure skating~\citep{bruin2006skating}, gymnastics~\citep{damisch2006olympic}, diving~\citep{kramer2017diving} and synchronized swimming~\citep{wilson1977swimming}. Indeed, in figure skating, performing just one position later increases the resulting percentile of the candidate by $2\%$ on average~\citep[Table 1]{bruin2006skating}, which is a non-trivial amount of deviation considering that the total number of candidates is typically around $10$ or $20$. Trusting evaluations in the presence of sequential bias is clearly problematic, because the outcomes of such competitions may have a long-term effect on candidates’ careers, similarly for other high-stakes applications such as court decisions~\citep{chen2016gambler}. It is thus important to alleviate sequential bias and draw correct conclusions in such applications. Moreover, the primitive approach of using a randomized ordering is not sufficient in one-shot competitions: While randomization improves fairness in some aggregate sense, one still needs to commit to a fixed realization of this random ordering in a particular competition. 

In this paper, we present new approaches to modeling and mitigating sequential bias. In order to motivate our model, let us begin by defining three characteristics of the setting that we would like to capture: sequential evaluation, relative rating, and prior knowledge. We use sports competitions as our running example, but these characteristics apply to other applications where sequential bias has been empirically observed, including music competitions~\citep{flores1996elisabeth,ginsburgh2003musical,antipov2017eurovision}, hiring~\citep{vives2021erosion}, grading~\citep{unkelbach2012oral}, residency matching at medical schools~\citep{elbeaino2019orthopaedic,harahan2021interview} and grant review~\citep{wieczorkowska2021grant}. 

\begin{enumerate}[label=(\arabic*)]
\item \textbf{Evaluation is carried out in a sequential manner.} As mentioned before, some sports competitions are inherently sequential, and judges score candidates immediately after their performances. A clear challenge in this setting is that we have to evaluate each candidate, without knowing information about future candidates. Consequently, evaluating candidates who appear earlier in the ordering—when judges have little information about the overall quality of the pool—is more challenging than evaluating candidates appearing later in the ordering. This aligns with the common belief---and empirical evidence (e.g.,~\citealp{flores1996elisabeth,bruin2006skating})---that candidates who perform earlier are at a disadvantage. There are two possible remedies to this state of affairs, but both come with challenges. The first is to allow the evaluator to amend the scores on previous candidates as and when they see new candidates. However, such a procedure can be hard to implement in practice. For example, judges cannot change the court decisions on previous cases. Such a procedure also runs the risk of introducing strategic behaviors. For example, if a competition judge is in favor of a particular candidate, then they can amend the scores of previous competitive candidates in order to decrease their chance of winning. Amending scores of previous candidates also increases cognitive burden on judges, since it requires that they faithfully recall and compare later candidates with the previous ones. A second possible remedy is to refrain from scoring sequentially, and instead evaluate all candidates simultaneously at the end of the process. This strategy once again increases cognitive effort, requiring judges to memorize a lot more information than what is practical. It could also introduce other types of sequential bias. For example, it is well known in psychology that the earliest and latest items in a sequence are easier to recall than the others~\citep{murdock1962recall}, and people positively perceive items that they recall better. In summary, both these natural remedies have drawbacks, which  motivates careful modeling of sequential bias in our settings of interest.

\item \textbf{Candidates are evaluated on a relative scale.} Another challenge with evaluation is that the scale may not be based on objective and directly quantifiable measures. For quantifiable measures (e.g., ``how tall is this candidate'', ``how fast does this candidate run''), a universally acceptable scale may be used. On the other hand, such a universal scale may not exist for more complex evaluation tasks (e.g., ``how good is this performance’’). Indeed, this is the primary reason why we have human judges in these settings! The most convenient way to define such a scale is in a relative fashion; to quote Unkelbach et al.~\citep{unkelbach2012oral}, ``objects have no categorical properties on their own, but only in reference to their context (i.e., heavy vs. light, large vs. small)''. For example, it is common to elicit scores according to some relative measure (e.g., a score of 5 means that the candidate is in the top 10\% percent) or relative language (e.g., defining the scales of 1, 2, and 3 as ``below average, above average, and exceptional’’ respectively). In addition to being natural and commonly used, a relative scale has the following advantages. First, it often aligns with the end goal of the evaluation task, which is to choose the top few candidates within our pool. Second, people may have different interpretations of absolute terms such as ``good'' or ``bad'', leading to the commonly observed phenomenon of miscalibration. A relative scale improves consistency among the evaluators, by providing a common scale according to which evaluators can calibrate themselves. Third, a relative scale is adaptive, in that one can use the same relative scale (e.g., ``selecting the top $10\%$) for widely different candidate pools of the evaluation task (say at the junior and senior levels of the same sports competition), whereas separately writing precise language to define the absolute scale for each pool is not only time-consuming but also requires a priori expert knowledge about the quality of each pool.

\item \textbf{Evaluators' prior knowledge plays a crucial role.} As alluded to before, a challenge in this setting is the lack of information about future candidates. This lack of information is particularly pronounced for inexperienced evaluators, who have little expertise or prior knowledge in judging the task at the specified level. At the same time, an expert evaluator’s scores may still exhibit sequential bias, albeit to a lesser extent, due to psychological effects such as the propensity to remember certain candidates~\citep{murdock1962recall}, and possible misalignment between the expert's prior knowledge and the distribution of the current pool.
Mismatches of this form can occur when: (a) The sample size is small---as is often the case in sports competitions---where the judges may not be able to rate accurately at the beginning of the task; (b) The sample size is relatively large, but the distribution of candidates changes over time.
For example, in hiring, the quality of the candidates may change from year to year (e.g., improving over the years due to technological advances in education, or changing suddenly due to  the effects of a global pandemic), so that experts still need to acquire information specific to this year's candidates in a sequential manner. 
In both of the cases (a) and (b), the fact that evaluators do not have a comprehensive view on the current pool of candidates explains their empirical tendency to be conservative in the beginning.\footnote{%
We make a qualifying note that there are applications in which the sample size is large \emph{and} the distribution changes very slowly, and this alleviates the need for careful bias correction. An example is grading GRE essays: Since graders evaluate these essays full-time, they see a large number of these essays from which they gradually develop an accurate sense of the distribution, and more importantly, they do not need to consistently keep making adjustments to their grading scale. In such applications, standard methods (e.g., occasionally inserting calibration essays and giving feedback to the graders when they deviate) are effective.
}
\end{enumerate}


In this work, we model and analyze sequential bias in ratings, paying particular attention to capturing the three defining characteristics described above. We propose a general class of models that describes how the rating of an item changes as a function of its position and its relative ranking with respect to the previous items (Section~\ref{sec:model_nonparametric}), where we use ``position'' to refer to the place of occurrence, and ``ranking/ordering'' to refer to the value comparison. To support various facets of this model, we conduct crowdsourcing experiments on a toy sequential evaluation task. These experiments suggest the existence of sequential bias (Section~\ref{sec:expt_existence}), the relative nature of the ratings (Section~\ref{sec:expt_relative}), and specific structure in the mistakes made when inferring comparisons from the scores (Section~\ref{sec:conflicts}).
We further study a subclass of our model described by certain natural \emph{parametric} assumptions (Section~\ref{sec:model_parametric}). In particular, we motivate this modeling choice by showing that scoring according to our parametric model is the theoretically optimal response if the evaluator’s goal is to minimize the squared error in reporting the (normalized) ranking of the items. Under the parametric model, we consider the statistical problem of estimating the underlying true ranking of the items under noisy observations. We propose a least squares estimator that is computable in near-linear time, and has a natural interpretation as an insertion algorithm (Section~\ref{sec:mle}). We study this this estimator and provide guarantees on its performance in two canonical ranking metrics. In a complementary direction, we show that these guarantees are optimal up to logarithmic factors, by proving matching information-theoretic lower bounds. In an effort to model more realistic settings, we then show that the methodology can be naturally extended to incorporate prior knowledge of the evaluators (\Cref{sec:prior}). We corroborate our theoretical results by examining the performance of the algorithm using numerical simulations and also on the crowdsourcing data we have collected (Section~\ref{sec:simulation}). We conclude with a discussion on the limitations and open problems (Section~\ref{sec:discuss}).

Overall, our work addresses the important problem of correcting sequential bias in a data-dependent manner. On the modeling front, our proposed formulation lays out a principled framework for understanding the mathematical foundations of the problem. On the methodological front, we combine tools from coding theory and ranking estimation with novel (to our knowledge) techniques in analyzing rankings, which may be of independent interest. 
All experiments conducted in this paper were approved by the Institutional Review Board (IRB) at Georgia Institute of Technology. The crowdsourcing data as well as all code to reproduce our results is available at \url{https://github.com/jingyanw/sequential-bias}.

\section{Related work}\label{sec:related_work}
In this section, we discuss related work, categorizing it under two verticals for convenience.

\subsection{Bias and evaluation in social science}

The first thread of related work motivates our models, and arises in the social science literature.

\paragraph{Sources of sequential bias.}
The existence of sequential bias has been widely observed in applications. Most commonly, a negative correlation between the position and the rank is observed~\citep{bruin2005dance,bruin2006skating}: The later an item appears in a sequence, the more likely the item is to be ranked as better. There also exist correlations between the assigned rank of an item and those of previous items. On the one hand, a positive correlation may arise due to affective priming, such as in emotion recognition~\citep{shen2019priming}. On the other hand, negative correlation may arise due to gambler's fallacy (or the so-called ``law of small numbers''~\citep{tversky1971small}). That is, people tend to overestimate how small-size samples are representative of the population characteristics, believing that ``early draws of one signal increase the odds of next drawing other signals''~\citep{rabin2002small}. Such negative correlations are empirically observed in court decisions and loan reviews~\citep{chen2016gambler}, where the current decision is negatively correlated with previous decisions made. In addition, effects of assimilation (focusing on similarities with previous items) and contrast (focusing on differences compared to previous items) have been observed in sports competitions~\citep{damisch2006olympic,kramer2017diving} and Idol series~\citep{page2010idol}. The generosity-erosion effect---by which the current candidate is graded more harshly to compensate for grading previous candidates generously---has been observed in hiring~\citep{vives2021erosion}.
Finally, people are more likely to remember items that appear at the beginning (primacy effect) and at the end (recency effect). As a result, the items that are remembered better also get graded more positively~\citep{page2010idol,collins2019talent}.

\paragraph{Relative evaluation.}
Using relative scales as supposed to absolute scales has been shown to be more effective in various judgment tasks such as job performance, attitude, and person perception~(see \citealp{goffin2011relative} and references therein). There are a few documented reasons for this.
First, people naturally develop a relative scale over time. In social comparison theory (e.g.,~\citealp{festinger1954social,kruglanski90comparison}), ``people are spontaneously inclined, and as a result of evolution perhaps even predisposed, to evaluate others and themselves in a comparative manner rather than an absolute manner''~\citep{goffin2011relative}. Second, absolute judgment is defined vaguely and has no explicit reference points, causing different raters to interpret the scale differently and exhibit miscalibration~\citep{griffin2008calibration}.

There are many different forms in which relative information can be elicited. A relative scale may be defined by using percentage information (e.g., ``a score of 5 corresponds to being top $10\%$'') or language of a relative nature (e.g., ``above average or below average'')~\citep{goffin2011relative,garg2020informative}. One may also directly elicit ordinal data in the form of a total ranking~\citep{shah2017design} or pairwise comparisons~\citep{shah2013mooc}.
Eliciting ordinal data has been shown to be more effective in domains such as grading~\citep{jones2015peer}, but is not naturally applicable to the sequential setting, where the items appear fleetingly and it is not possible to have them appear repeatedly for pairwise comparisons.

\subsection{Statistical methodology and coding}

The second thread of related work is methodological research in statistics.

\paragraph{Probabilistic ranking models.}
The main idea of our proposed model (to be presented formally in Section~\ref{sec:model_nonparametric}) is to have the ratings reported by the evaluators be akin to a ``noisy'' execution of an insertion algorithm. There is a statistical literature on probabilistic ranking models based on insertion sorting. Doignon et al.~\citep{doignon2004repated} propose the repeated insertion model (RIM) as a generalization of the popular Mallows ranking model. In RIM, an ordering of items is constructed by an insertion procedure, where at each step, item $i$ is inserted to position $t$ with some known probability $p_{it}$. This insertion procedure induces a distribution over the final ordering of the items. RIM makes the assumption that each insertion is independent of how previous insertions have been carried out, which is an assumption that we also make. However, the goal in RIM is typically to compute the marginal probability of partial orderings~\citep{kenig2018repeated} (such as the probability of item $i$ being ranked higher than item $j$) given a known true ranking, while our goal is to infer the true ranking from the noisy observed scores.
In a similar spirit to RIM, other probabilistic ranking models have been proposed~\citep{biernacki2013generative,mesaoudi-paul2018noisy}, in which the distribution of the observed ranking is induced by an insertion sort procedure. In particular, each item is compared pairwise with all existing items from left to right (with some given probability of winning each comparison), and inserted to the position at which it first loses a comparison. By contrast, we consider the rating component where the information is not just ordinal but is observed as  a cardinal score.

\paragraph{Inversion tables for rankings.}
Our proposed model uses the relative rank of each item, defined as the rank of each item with respect to the items so far. Relative ranks have a close connection to inversion tables~\citep[Section 5.1.1]{knuth1998art} for rankings and the so-called Lehmer code~\citep{lehmer1960tricks}. Inversion tables can be defined in many equivalent forms. Let $\sigma:[n]\rightarrow [n]$ be any permutation of $n$ items. One definition of an inversion table is an $n$-dimensional vector whose $i^\textth$ component is the number of elements in $\sigma$ to the right of item $i$ (whose value is $\sigma(i)$) that have values smaller than item $i$~\citep{lehmer1960tricks}.
Motivated by applications such as flash memory devices, a series of work~\citep{jiang2010modulation,mazumdar2013modulation,wang2015compression} studies the construction of codes of permutations. Some of these constructions use inversion tables, and in particular, a relation between the $\ell_1$ distance of inversion tables and the Kendall--tau distance between rankings. From this perspective, our work can be viewed as ranking estimation under a specific noise model on the inversion tables relevant to the sequential evaluation setting. The relation between inversion tables and the Kendall--tau distance is also used in proving one of our lower bounds (Theorem~\ref{thm:lb_sf}).

\paragraph{Detecting drifts.}
A number of methods have been proposed to detect raters' drift on large-scale grading tasks. A well-known method is DRIFT~\citep{wolfe2001drift}, using which rater drift has been shown in music assessment~\citep{wesolowski2017solo}, AP essay scoring~\citep{myford2009drift}, elementary school writing scoring~\citep{congdon2000assessment}, and clinical skill assessment~\citep{harik2009generalizability}. As explained in the introduction (see footnote 2), large-scale tasks do not align with the setting we consider here. Moreover, the prior methods here address detection but not estimation or correction.

\paragraph{Permutation-based models.}
Finally, there is a line of work on using permutation-based models---a general class of models in which an unknown latent permutation governs the observations---for statistical inference.
These models have the benefit of not making overly restrictive parametric assumptions for applications spanning choice modeling~\citep{shah2017sst,pananjady2022isotonic}, crowd-labeling~\citep{shah2021permutation}, network modeling~\citep{gao2021minimax}, correspondence estimation~\citep{hsu2017linear,pananjady2017denoising}, seriation~\citep{flammarion2019optimal}, and evaluation problems~\citep{wang2021evaluation}. The proposed non-parametric constraints for our model follow this line of work.

\section{Modeling ratings: Setup and intuition}\label{sec:model_nonparametric}

Consider a set of $\numitems$ items, and let $\setperms_\numitems$ denote the set of all permutations on $\numitems$ items. We assume that there exists an underlying true ranking of the $\numitems$ items; let $\invpermtrue \in \setperms_\numitems$ denote this true ranking (in the competition example, the true ranking is based on the quality of the candidates' actual performance).  In particular, the rank of item $\idxitem$ is $\invpermtrue(\idxitem)$, where a higher value of rank corresponds to a better item in the total ordering. For illustration, $\invpermtrue = [1, 3, 2]$ implies that the worst item appears first and the second-worst appears last.

The items are presented in a sequential order, with each item $\pos\in [\numitems]$ being revealed at timestep $\pos$.
At each timestep $\pos$, the evaluator gives a rating of $\report_\pos\in \reals$ to item $\pos$ based on how this item fares in comparison to the $(\pos-1)$ previous items.
For any ranking $\invperm$, we define the ``relative rank'' of each item $\pos$ as the rank of the current item $\pos$ relative to the $(\pos-1)$ previous items according to $\invperm$, given by
\begin{align}\label{eq:def_rel_rank}
    \rankrel_\pos(\invperm) \defn \abs*{\{\idxitem\in [\pos]: \invperm(\idxitem) \le \invperm(\pos)\}},
\end{align}
where the larger $\rankrel_\pos(\invperm)$ is, the better item $\pos$ is in comparison to the previous $(\pos - 1)$ items according to $\invperm$.
We also refer to $\invperm(\pos)$ as the ``absolute rank'' of item $\pos$ to distinguish it from the relative rank $\rankrel_\pos(\invperm)$. For notational simplicity, we use the shorthand $\rankreltrue_\pos \defn \rankrel_\pos(\invpermtrue)$ to denote the relative rank of item $\pos$ according to the true ordering $\invpermtrue$. We assume that the score given to item $\pos$ depends on its position $\pos$ and its relative rank $\rankreltrue_\pos$. Specifically, we assume that the evaluator gives a score $\report_\pos$ to item $\pos$ following
\begin{align}
    \report_\pos & = \param(\pos, \rankreltrue_\pos) + \noise_\pos \label{eq:model_nonparametric}
\end{align}
where $\param: [\numitems]\times [\numitems]\rightarrow \reals$ is a function describing the dependence of the reported score on the position $\pos$ of the item and its relative rank $\rankreltrue_\pos\in [\pos]$. We use $\report\in \reals^\numitems$ to denote the vector $\{\report_\pos\}_{\pos\in [\numitems]}$. The term $\noise_\pos$ denotes sub-Gaussian noise that is independent across $\pos\in [\numitems]$.
The causal nature of this model is natural: Since the evaluator scores items without knowing what items will arrive in the future, the information useful for this evaluation is precisely what the evaluator has seen about the items so far.

\begin{figure}[tb]
    \centering
    \includegraphics[width=0.35\linewidth]{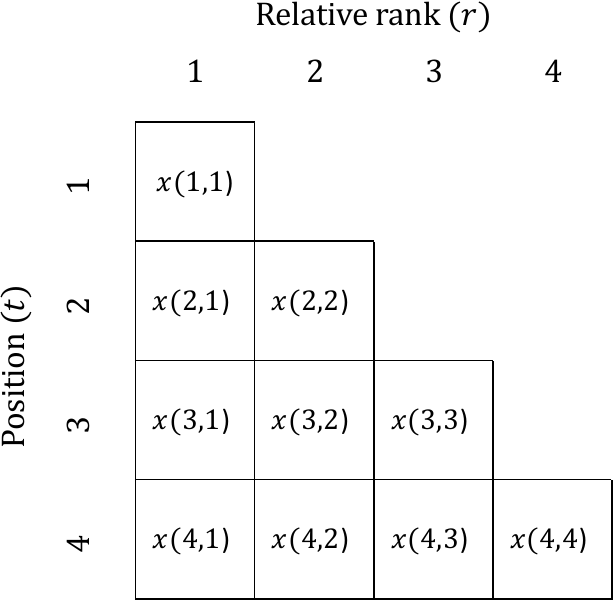}

    \caption{The model~\eqref{eq:model_nonparametric} is characterized by the set of parameters $\{\param(\pos, \rankrel)\}_{\pos\in [\numitems], \rankrel\in [\pos]}$.
    \label{fig:model}}
\end{figure}

The model~\eqref{eq:model_nonparametric} is described by the set of parameters $\{\param(\pos, \rankrel)\}_{\pos \in [\numitems], \rankrel \in [\pos]}$, as shown in Figure~\ref{fig:model}. We assume the ordering constraints
\begin{align}\label{eq:model_constraint}
    \param(\pos, \rankrel) < \param(\pos, \rankrel') \qquad \text{ for all } \pos\in [\numitems] \text{ and all } \rankrel, \rankrel'\in [\pos] \text{ with }\rankrel < \rankrel'.
\end{align}
The constraints~\eqref{eq:model_constraint} capture a natural desideratum for good-faith evaluators: 
At each fixed position $\pos\in [\numitems]$, the evaluator's (mean) score is monotone in the item's relative rank.
Note that we assume strict inequality in the constraints~\eqref{eq:model_constraint} to avoid pathological cases, e.g., in which all items are given an identical score independently of the pair $(\pos, \rankrel)$.

Before studying model~\eqref{eq:model_nonparametric}, we make a few remarks about our modeling choice. First, model~\eqref{eq:model_nonparametric} assumes for simplicity that the reported scores depend only on the current evaluation task, without external or prior knowledge influencing the evaluation. This model can be naturally extended to incorporate prior knowledge. This extension is formally presented in Section~\ref{sec:prior} and experimentally evaluated in \Cref{sec:expt_crowdsourcing}.
Second, we make a particular choice of how relative ranks affect scores. Specifically, the evaluator reasons on a relative scale through the true relative rank $\rankreltrue_\pos$ of the current item comparing to previous items. An extension—as suggested by an anonymous referee—is to posit that $y_t = x(t, \rankreltrue_\pos, \{\report_\idxitem\}_{\idxitem\in [\pos-1]}) + \epsilon_\pos$ (or more generally $y_t = x(t, \{\rankreltrue_\idxitem\}_{\idxitem\in [\pos]}, \{\report_\idxitem\}_{\idxitem\in [\pos-1]}) + \epsilon_\pos$, so that the score for each item also depends on the noisy scores  $\{\report_\idxitem\}_{\idxitem\in [\pos-1]}$ that are already given to previous items. Section~\ref{sec:discuss} discusses other possible extensions, and formalizing these extensions is an interesting direction for future work.

In the rest of this section, we collect a crowdsourced dataset, and describe experimental evidence to motivate various facets of the proposed model. We first describe the set up of the experiments, and then present our results.

\subsection{Experimental set up}

\begin{figure}[tb]
    \centering
    \includegraphics[width=0.7\linewidth]{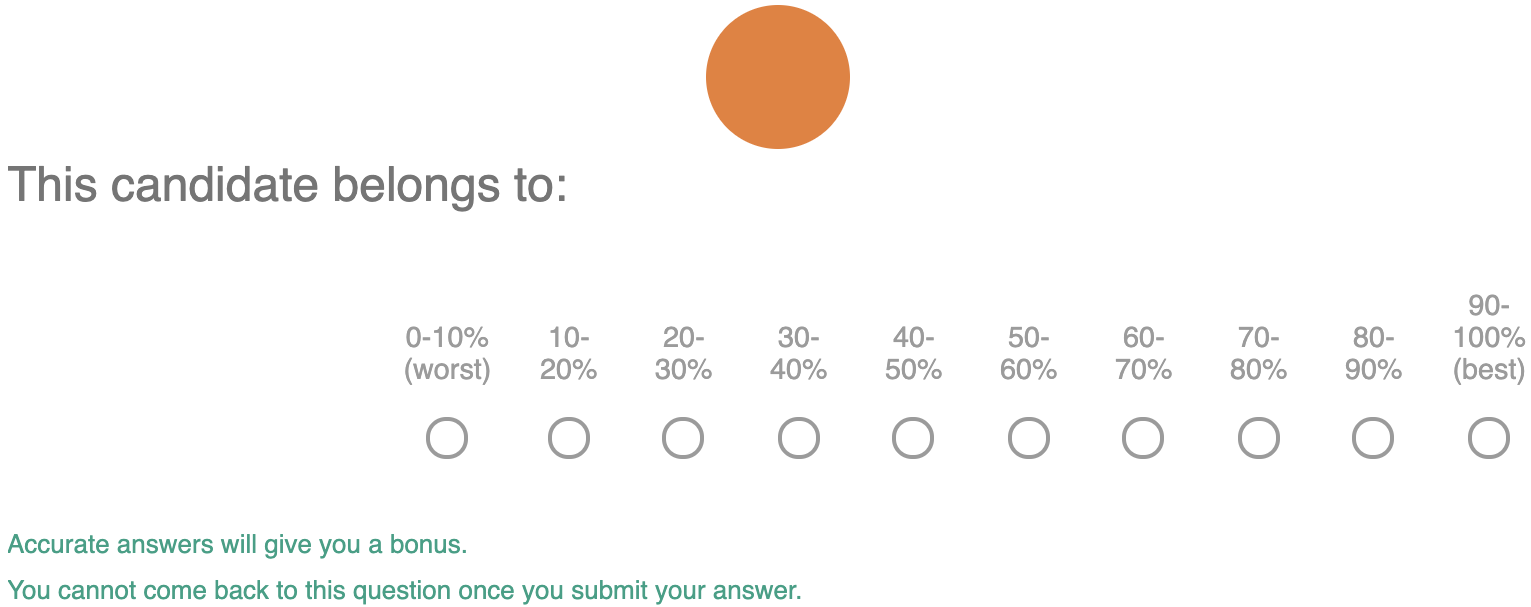}
    \caption{The interface shown to the \workers for evaluating each candidate in sequence.}
    \label{fig:worker}
\end{figure}

We create a simple evaluation task of a sequential and relative nature, as follows. We recruit \workers 
from the online crowdsourcing platform Prolific. The \workers are instructed that they will be presented a set of items one-by-one, in an ordering that they do not know in advance. To align with common real-world scenarios, the \workers are informed in advance the total number of items $\numitems$ (in our experiments, we use $\numitems=5$ or $10$).
Each item is represented by a circle shown on the \worker's screen. The circles have different sizes, and the sizes are chosen such that any two circles of most similar size are still easily distinguishable if they were presented on the same page. The evaluation task is to estimate how large the circles are in comparison to the entire pool of the $\numitems$ circles. The \workers are instructed not to hold prior beliefs about how large the circles might be, but instead to gradually learn the scale from the items presented to them.
For each item, the \workers are given a decile scale, consisting of $10$ bins of $0\%\text{--}10\%, 10\text{--}20\%$ etc. up to $90\text{--}100\%$. The \workers are instructed that the bin of $0\text{--}10\%$ means that the item falls within the lowest $10\%$ among the $\numitems$ items, etc. For each item, the task for the \workers is to classify the item into one of the $10$ bins, and they are instructed that their eventual reward (payment) for the task will be based on the accuracy of their answers. Once the \workers report their evaluation for an item, a new page is shown to present the next item, and the \workers are not allowed to go back and amend their answers to previous items. Figure~\ref{fig:worker} shows an example page presented to the \workers.

A few remarks on this setup are in order. For complicated and possibly subjective evaluation tasks in realistic scenarios, errors arise due to both the ambiguity or difficulty in inferring the true ranking, and additional errors made by evaluators even if the true ranking were clear, due to evaluators' limited memory, lack of calibration, etc. For simplicity and following prior work~\citep{shah2016topology}, we choose a cognitively straightforward task of comparing circle sizes to focus on studying the latter type of error. Additionally, the design choices in our experiments are made to align with the characteristics of sequential evaluation in real-world scenarios (see \Cref{sec:intro}); let us give two illustrative examples. First, we do not allow workers to review previous items or amend their scores, because judges in competitions rely solely on their memory in terms of the performances of previous candidates, and are often not allowed to amend their scores. Second, we choose decile bins to simplify the relative scale, and the workers are instructed not to hold prior beliefs (although we observe that they still do --- see the results to follow in \Cref{sec:expt_existence} and \Cref{sec:prior}).
We now present three crucial properties implied by our proposed model~\eqref{eq:model_nonparametric}, and show that these are borne out in the experimental results.

\subsection{Existence of sequential bias} \label{sec:expt_existence}
First, we make the immediate observation that our model~\eqref{eq:model_nonparametric} captures sequential bias by virtue of having $\param(\pos,\rankreltrue_\pos)$ be dependent on the position $\pos$ (directly and through $\rankreltrue_\pos$). We now experimentally examine whether the scores of the items do depend on their positions.

\paragraph{Set up.} We consider $n=5$ items, yielding $5!=120$ possible orderings in total.  We recruit $240$ \workers. Each ordering is assigned to two \workers uniformly at random.

\begin{figure}[tb]
    \centering
    \includegraphics[width=0.4\linewidth]{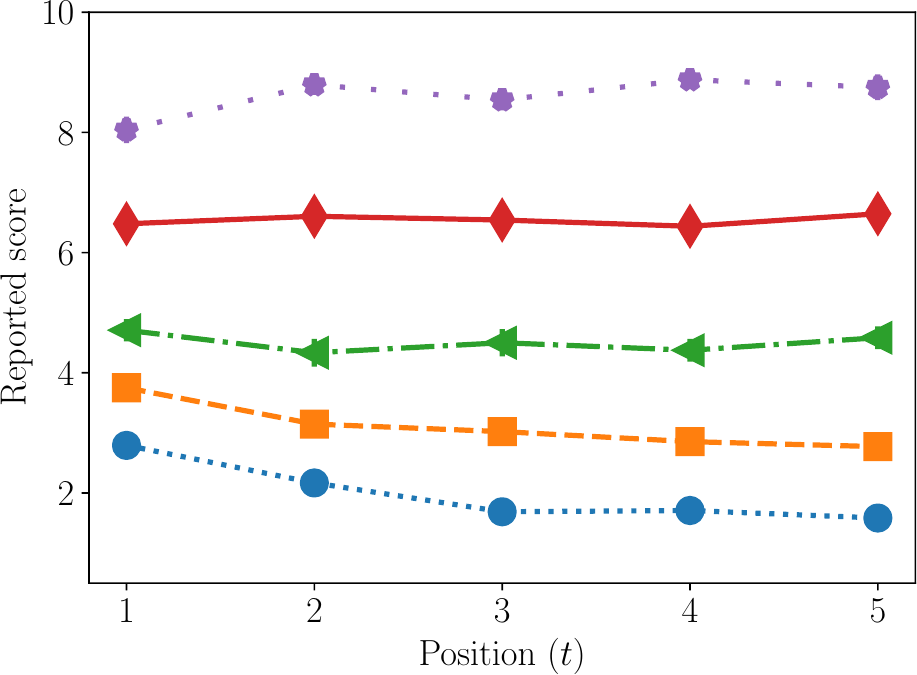}
     \hspace{1cm}
        \begin{minipage}[t]{0.175\linewidth}
            \raisebox{1.25cm}{\includegraphics[width=\linewidth]{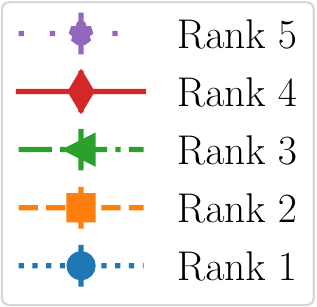}}
        \end{minipage}
   
    \caption{Mean score of each item appearing at different positions. Error bars (barely visible) represent standard error of the mean.\label{fig:bias_position}}
    
\end{figure}

\paragraph{Result.} In Figure~\ref{fig:bias_position}, we plot the mean score (y-axis) received by the same item at different positions (x-axis), where we map the bins from $0\text{--}10\%$ through $90\text{--}100\%$ to values from $1$ through $10$. For the lowest two items (rank 1 and rank 2), we observe a decreasing curve; for the highest item (rank 5) we observe an increasing curve (albeit to a lesser extent). This confirms the existence of sequential bias. More specifically, at the beginning of the sequence, the mean scores of different items are closer, because the \workers may anticipate more extreme items to appear in future, and hence rate conservatively. At the end of the sequence, the \workers have gradually collected information about the pool, and are able to more accurately distinguish different items.
These results align with existing literature that ``being first is bad when you are good, and good when you are bad''~\citep{unkelbach2014judgments}, confirming the validity of our experimental design. 

In Figure~\ref{fig:bias_position}, we also observe that at the first position, different items still receive different scores positively correlated with their quality, despite the fact that the evaluator has not seen any items yet. This reveals a limitation of our experiment that \workers still carry some prior knowledge about the evaluation and apply a combination of absolute and relative scales: We show \workers some shapes as part of the task instructions, and \workers may use these as anchor points. They may also calibrate under the reasonable expectation that the circle sizes do not exceed their screen size.

\subsection{Relative nature of scores}\label{sec:expt_relative}
Next, we conduct an experiment to verify that the scores given to the items are indeed relative. In particular, the constraints~\eqref{eq:model_constraint} imply that the scores of different items are negatively correlated.
To see this, consider the relative rank $\rankreltrue_\pos$ of item $\pos$. If the previous items have high ranks, then $\rankreltrue_\pos$ is small. If the previous items have low ranks, then $\rankreltrue_\pos$ is large. By the monotonicity constraints~\eqref{eq:model_constraint}, item $\pos$ will receive a higher score in the former case than in the latter case, and hence there should be a negative correlation between the scores of this item and those of previous items. 
This negative correlation arises precisely from relativity: The non-parametric model~\eqref{eq:model_nonparametric} assigns scores based on the rank computed \emph{relatively} to all previous items.

\paragraph{Set up.} We consider $\numitems=10$ items. Let $\setlow$ and $\sethigh$ denote the subset of the $5$ smallest items and the subset of the $5$ largest items, respectively. We recruit $50$ \workers in total, and divide them into two groups with $25$ \workers each. 
For each $\idxworker\in [25]$, we sample two i.i.d. rankings $\invpermtrue_{1,\idxworker}, \invpermtrue_{2,\idxworker}: [5]\rightarrow [5]$ uniformly at random. For any ranking $\invperm \in \setperms_{\numitems}$, and a set $\set$ of $\numitems$ items, we slightly abuse notation and let $\invperm(\set)$ denote the permutation of the items in $\set$ such that the permuted sequence follows the same ranking as $\invperm([\numitems])$. 
Then we present the $\idxworker^\textth$ \worker in group 1 the items in the order of
\begin{align*}
\invpermtrue_{1,\idxworker}(\setlow), \invpermtrue_{2,\idxworker}(\sethigh),
\end{align*}
and present the $\idxworker^\textth$ \worker in group 2 the items in the order of
\begin{align*}
    \invpermtrue_{1,\idxworker}(\sethigh), \invpermtrue_{2,\idxworker}(\sethigh). 
\end{align*}
In words, we present both groups the highest $5$ items at positions $6\text{--}10$, but present to them different items at positions $1\text{--}5$ (small for group $1$ and large for group $2$). If the evaluation were solely based on an absolute scale (i.e. the size of the circles alone), then the two groups should give similar ratings to the last $5$ items. On the other hand, if the ratings given to the last $5$ items are significantly different, then it suggests that the \workers adaptively learn the scale from observing the first $5$ items, suggesting that the scale is relative to previous items.

\paragraph{Result.} We collect all the scores given by group 1 to the last 5 items (totaling $25\times 5 = 125$ scores), and collect all the scores given by group 2 to the last 5 items (totaling another $125$ scores). We again map the bins from $0\text{--}10\%$ through $90\text{--}100\%$ to values from $1$ through $10$. The sample means (plus-minus standard error of the mean) in the two groups are $8.99 \pm 0.10$ and $6.78 \pm 0.21$, respectively. 
This result suggests that different grading scales are constructed by the two groups of \workers.
Since group 2 is presented large items to start with, they learn to calibrate down and hence give lower scores to the last 5 items than group 1.
We perform a univariate permutation test on these two collections of scores, using the sample mean of each group as the test statistic. We reject the null hypothesis that these two collections of scores are generated from the same distribution (with one-sided $p$-value ${<}$ 0.01).

As an aside, we note that this setup also provides a second piece of evidence for the existence of sequential bias. To make this concrete, consider group 2 alone. By construction, the first $5$ items and the last $5$ items are identical. Using a similar procedure to collect $125$ scores for either of the two subsets, we find that the sample means of the ratings are $7.94 \pm 0.18$ and $6.78 \pm 0.21$, respectively. Performing a univariate permutation test, we reject the null hypothesis that they are generated from the same distribution (with one-sided $p$-value ${<} 0.01$). This shows (once again) the existence of sequential bias in this setting.

\subsection{Conflicts between ratings and rankings}\label{sec:conflicts}

Finally, we verify that sequential bias does indeed have adverse effects, in that it renders the ranking induced by the reported scores different from the true ranking of the items. 
To illustrate, we consider pairwise ``conflicts'' between the scores, which we describe in detail below. 

Consider any ordering $\invpermtrue$ of $\numitems$ items, and consider any two items $\idxitem, \idxitemalt\in [\numitems]$. Suppose for the moment that we operate in the noiseless setting, in which the scores assigned to the two items are $\param(\idxitem, \rankreltrue_\idxitem)$ and $\param(\idxitemalt, \rankreltrue_\idxitemalt)$. If the pairwise comparison induced by the reported ratings and the true comparison have opposite signs (namely, if $\param(\idxitem, \rankreltrue_\idxitem)> \param(\idxitemalt, \rankreltrue_\idxitemalt)$ and $\invpermtrue(\idxitem) < \invpermtrue(\idxitemalt)$, or if $\param(\idxitem, \rankreltrue_\idxitem)< \param(\idxitemalt, \rankreltrue_\idxitemalt)$ and $\invpermtrue(\idxitem) > \invpermtrue(\idxitemalt)$), then we say that the pair of items $(\idxitem, \idxitemalt)$ form a ``conflict'' under ordering $\invpermtrue$.
In what follows, we show that the existence of conflicts is a property of the constraints~\eqref{eq:model_constraint}, and present an experiment that corroborates this property.
\begin{proposition}\label{prop:conflict_exist_nonparametric}
Let $\numitems \geq 4$, and consider the model~\eqref{eq:model_nonparametric} in the noiseless setting. Consider any instantiation of the parameters $\{\param(\pos, \rankrel)\}_{\pos\in [\numitems], \rankrel\in [\pos]}$ that satisfies the constraints~\eqref{eq:model_constraint}. There always exists some ranking $\invpermtrue \in \setperms_\numitems$ such that a conflict exists under $\invpermtrue$.
\end{proposition}

The proof of this proposition is provided in Section~\ref{sec:proof_prop_conflict_exist_nonparametric}. In the proof, we construct a simple example of a pair of rankings consisting of $4$ items each. We assume that there is no conflict, and use the constraints~\eqref{eq:model_constraint} to derive contradictions. We now conduct the following experiment to examine if conflicts are indeed observed in practice.

\paragraph{Set up.}
We consider $\numitems=5$ items, and consider the two orderings:
\begin{subequations}\label{eq:conflict_ranking}
\begin{align}
    & \invpermtrue_1 = [1, 3, 4, 5, 2]\label{eq:conflict_ranking_one}\\
    & \invpermtrue_2 = [2, 3, 4, 5, 1].
\end{align}
\end{subequations}
Note that these two rankings exchange the items at the first position and the last position, keeping the other three items the same. We recruit $100$ \workers, and divide them into two groups with $50$ \workers each. 
\Workers in group 1 are presented items ordered according to $\invpermtrue_1$, and \workers in group 2 are presented items ordered according to $\invpermtrue_2$.
Our goal is to examine whether the first item and the last item form a conflict. According to the definition of a conflict, for each \worker we inspect the comparison between these two items induced by the \worker's ratings, and compare it against the true comparison. If these two items form a conflict, the comparison of the \worker's ratings and the true comparison differ.

However, note that conflicts are defined based on noiseless scores, whereas in reality noise is inevitable, so the two comparisons may differ purely due to noise. We reason that a major source of noise is the imprecise memory of the \workers. Specifically, when evaluating the last item, if the \worker cannot recall the first item and compare them correctly, then the \worker is likely to report very noisy ratings compared to the \workers who are able to compare the two items correctly. In order to reduce this type of noise, at the end of the survey, we ask the \workers if they recall the first item or the last item is better (``Based on your memory, is Candidate 1 (first) better or Candidate 5 (last) better?'').
If \workers correctly recall the true ordering but still actively choose to give ratings opposite to this comparison, then we reason that this discrepancy likely arises because their changing relative scale of calibration results in conflicts.

\paragraph{Result.}
We classify the \workers into categories based on whether they correctly recall the comparison between the first item and the last item in the last question asked in the survey (``correct/wrong recall''), and whether the comparison induced by their scores agrees with the true ranking (``correct/wrong rating''). The number of workers in the four categories is shown in Table~\ref{tab:conflict} for the two groups.

Since the last item follows three large items (of ranks $3, 4$ and $5$) in both groups, the relative nature of the scores as shown in Section~\ref{sec:expt_relative} suggests a negative bias on the last item. Intuitively, group 2 has an easier evaluation task, because the last item is the smallest, and the negative bias only further lowers its score. In Table~\ref{tab:conflict}, most participants ($47/50$) in group 2 both correctly recall the comparison between the first and last items (``correct recall''), and provide correct ratings consistent with this comparison (``correct rating''). 

On the other hand, group 1 is faced with a more difficult task. The last item is actually larger than the first one, but the negative bias lowers the score of the last item, making it easier to confuse it with the first item. Following the intuition from above, workers that wrongly recall the comparison between the first and last items are noisy, so information about conflicting behavior is better gleaned from looking at the first row of Table~\ref{tab:conflict}. Indeed, $9/25$ workers with correct recall also provide the correct rating, indicating that this category provides more reliable data than workers with wrong recall ($1/25$). Furthermore, within the ``correct recall'' category, a significant fraction of the \workers ($16/25$) still exhibit a disagreement (``correct recall \& wrong rating''), hence providing evidence for the existence of a conflict between the first and the last item in the ordering $\invpermtrue_1$.

\begin{table}[tb]
    \centering
    \begin{tabular}{c|c|c}
        \textbf{group 1} & correct rating & wrong rating\\\hline
        correct recall & $9$ & $\mathbf{ 16}$\\\hline
        wrong recall & $1$ & $24$
    \end{tabular}
    
    \vspace{0.5cm}
    \begin{tabular}{c|c|c}
        \textbf{group 2} & correct rating & wrong rating\\\hline
        correct recall & $47$ & $0$\\\hline
        wrong recall & $2$ & $1$
    \end{tabular}
    \caption{The number of \workers in each category, where categories are defined based on reported comparisons in the survey (``correct/wrong recall'') and score-induced comparisons (``correct/wrong rating'') between the first item and the last item in the ordering~\eqref{eq:conflict_ranking} assigned to that group.}
    \label{tab:conflict}
\end{table}

\section{A parametric model}\label{sec:model_parametric}

Having verified that various properties of our general model are borne out in our crowdsourcing experiments, we now turn to the question of whether one can \emph{correct} sequential bias in such tasks. We do so for a  special case of the general model in which we make further parametric assumptions. 
In particular, we suppose that
\begin{align}\label{eq:model_parametric_param}
    \param(\pos, \rankrel) = \frac{\rankrel}{\pos + 1},
\end{align}
so that
\begin{align}\label{eq:model_parametric}
    \report_\pos = \frac{\rankreltrue_\pos}{\pos + 1} + \noise_\pos
\end{align}
models the evaluation of item $\pos$.
For example, when the first item arrives, its relative rank $\rankreltrue_1$ is trivially $1$, and hence in the noiseless setting the item receives a score of $\frac{1}{2}$, regardless of its quality. Intuitively, this is because we have no information to evaluate this item relative to other items, simply because no other item has arrived yet. Then item $2$ arrives. Depending on whether item $2$ is smaller or greater than item $1$, its relative rank $\rankreltrue_2$ is $1$ or $2$, and hence its score in the noiseless setting, namely $\param(2, \rankreltrue_2)$, is either $\frac{1}{3}$ or $\frac{2}{3}$ correspondingly, and so on for the later items.

Let us provide a few further nuggets of intuition for this model. First, we inspect the behavior of our observations as $\pos \to \infty$. The noiseless score $\frac{\rankreltrue_\pos}{\pos+1}$ of item $\pos$ converges to its true percentile as $\pos\rightarrow \infty$. Formally, if each item $\pos$ has a true value $x_\pos\in \reals$ drawn i.i.d. from some distribution $F$, then the noiseless score $\frac{\rankreltrue_\pos}{\pos + 1}$ converges to the true percentile of the current item, namely the inverse c.d.f. $F^{-1}(x_\pos)$. Our parametric model thus captures the intuition that the evaluator's empirical estimate of the percentiles (based on the items seen so far) becomes more accurate as the number of items increases. Moreover, in the limit of an infinite number of items, the percentile estimate is perfect, as one would have gained perfect knowledge about the distribution.

For a second nugget of intuition, we inspect the comparison between adjacent items $\pos$ and $(\pos+1)$ in the noiseless setting. If item $\pos$ is ranked lower than item $(\pos + 1)$, then we have relative ranks $\rankreltrue_\pos < \rankreltrue_{\pos+1}$, yielding the relation $\frac{\rankreltrue_\pos}{\pos+1} < \frac{\rankreltrue_{\pos+1}}{\pos+2}$ (where we have also used the fact that $\rankreltrue_\pos, \rankreltrue_{\pos + 1}$ and $\pos$ are integers). Consequently,
\begin{subequations}\label{eq:adjacent_comparison}
\begin{align}
    \param(\pos, \rankreltrue_\pos) < \param(\pos+1, \rankreltrue_{\pos+1}).
\end{align}
Likewise, if item $\pos$ is ranked higher than item $\pos+1$, we have $ \rankreltrue_\pos \geq \rankreltrue_{\pos+1}$, yielding the relation $\frac{\rankreltrue_\pos}{\pos+1} > \frac{\rankreltrue_{\pos+1}}{\pos+2}$. Consequently,
\begin{align}
    \param(\pos, \rankreltrue_\pos) > \param(\pos+1, \rankreltrue_{\pos+1}).
\end{align}
\end{subequations}
Combining the two cases in Eq.~\eqref{eq:adjacent_comparison}, the parametric model implies that comparisons between adjacent items are always correct in the noiseless setting. This aligns with the intuition that people compare adjacent items well because of their immediate contrast. To take the example of our crowdsourcing experiment, when \workers flip a page, it is easy to tell whether the circle has grown larger or smaller, but it gets difficult to compare circles that are shown farther apart in the sequence.

Finally, let us re-examine the scores given by the parametric model to the first item and the last item in the ordering~\eqref{eq:conflict_ranking_one}, noting that the experiment in Section~\ref{sec:conflicts} shows a conflict on this pair. The mean scores of these two items predicted by the parametric model are $\param(1, 1)=\frac{1}{2}$ and $\param(5, 2) = \frac{2}{6}$, respectively. The comparison of the scores ($\frac{1}{2} > \frac{2}{6}$) and the true ranking ($1<2$) are opposite. Hence, the parametric model correctly predicts a conflict on these two items.

As indicated by these properties above---and given its simplicity---the parametric model is a natural choice for capturing how sequentially arriving items are evaluated. Next, we provide a further theoretical justification for studying this model by appealing to an incentive-based argument.

\subsection{Optimal response}\label{sec:opt_response}

In this section, we take the perspective of the evaluator and show that if they intend to minimize their error in estimating the (normalized) true ranking of the items, then the parametric model characterizes their ``best response'' in the sequential setting.

To set the stage, consider $\numitems$ items, and suppose that the evaluator assumes a uniform prior on the ranking, i.e., $\invpermtrue\in \setperms_\numitems$ is chosen uniformly at random over all $\numitems!$ permutations.
For any ordering $\invperm$, we let $\invpermrestrict{\pos}$ denote the ordering restricted to the first $\pos$ items (in a relative sense). For example, if $\invperm = [1, 4, 3, 2]$, then we have $\invpermrestrict{2} = [1, 2]$ and $\invpermrestrict{3} = [1,3,2]$. With this definition, for the relative rank we have the relation $\rankrel_\pos(\invperm) = \invpermrestrict{\pos}(\pos)=\rankrel_\pos(\invpermrestrict{\pos})$.
At each timestep $\pos\in [\numitems]$, we assume the evaluator observes the ranking $\invpermtruerestrict{\pos}$ restricted to the first $\pos$ items. This is equivalent to observing noiseless scores from the parametric model~\eqref{eq:model_parametric}. Then the evaluator reports their score for item $\pos$ according to some response function $\estscore_\pos: \setperms_\pos\rightarrow \reals$. In particular, the evaluator reports a score of $\estscore_\pos(\invpermtruerestrict{\pos})$ upon seeing item $\pos$. 

We assume that the goal of the evaluator is to estimate the true ranking $\invpermtrue$ of the items, divided by a normalization factor of $(\numitems + 1)$. For this task, we consider the squared error defined as
\begin{align}\label{eq:err_est_score}
    \Expect \left[\sum_{\pos\in [\numitems]}\left(\estscore_\pos(\invpermtruerestrict{\pos}) - \frac{\invpermtrue(\pos)}{\numitems+1}\right)^2\right],
\end{align}
where the expectation is taken over the uniform prior on $\invpermtrue$ and possible randomness in the estimator $\estscore$.
The following proposition shows that if the evaluator aims to minimize the loss~\eqref{eq:err_est_score}, then their optimal response is to  rate according to the parametric model~\eqref{eq:model_parametric}.

\begin{proposition}\label{lem:incentive_rank} 
Consider $\numitems$ items, and suppose that the true ordering $\invpermtrue$ is chosen uniformly at random from $\setperms_n$. Consider (possibly randomized) estimators of the form $\estscore=\{\estscore_\pos(\invpermrestrict{\pos})\}_{\pos\in [\numitems]}$, where $\numitems$ can either be known or unknown to the estimator. Then among all estimators, the estimator $\estscore_\pos(\invpermrestrict{\pos})= \frac{\rankrel_\pos(\invpermrestrict{\pos})}{\pos + 1}$ minimizes the squared loss~\eqref{eq:err_est_score}.
\end{proposition}

The proof of this proposition is provided in \Cref{sec:lem_incentive_rank}. Intuitively, at each timestep $\pos$, we have sampled $\pos$ numbers from $[\numitems]$ uniformly at random without replacement. 
One should thus expect that the $\pos$ numbers are evenly spread out from $1$ to $\numitems$, inducing the expected $\rankrel^\textth$ order statistics to be $\frac{\numitems+1}{\pos+1}\cdot \rankrel$ (this is formally described by Lemma~\ref{lem:expect_dist} in Appendix~\ref{app:proof_lem_expect_dist}). After normalizing by $(\numitems+1)$ and given that we are estimating under the squared loss, the rating $\frac{\rankrel}{\pos+1}$ is a natural choice for minimizing the loss.

In spite of this transparent intuition, we note that the result is nontrivial for several reasons. First, since the loss in Eq.~\eqref{eq:err_est_score} is computed by summing up the error over $\numitems$ items, one might expect that shrinkage estimators can trade off variance for bias, akin to examples in which Stein's paradox is observed~\citep{stein1956inadmissibility}. This turns out not to be possible because shrinkage is a frequentist phenomenon, whereas in our case the true ordering is random and has a known distribution. Second, the random quantities (normalized true ranks of each item) that we are estimating are all dependent. One might reasonably expect that the optimal estimator can take advantage of this dependence to reduce the error, but this turns out not to be the case.
Third, the quantity we are estimating involves a normalization factor of $\frac{1}{\numitems+1}$, i.e., the true estimands of interest depend on $\numitems$. However, the optimal response is independent of $\numitems$, and the evaluator does not need to know the value of $\numitems$ a priori. 

Zooming out, Proposition~\ref{lem:incentive_rank} shows that if the evaluators behave in good faith and attempt to maximize the accuracy of the predictions as defined by the squared loss~\eqref{eq:err_est_score}, then it is in their best interest to rate items according to the parametric model~\eqref{eq:model_parametric}. 
The parametric model thus describes ``reasonable" behavior in sequential evaluation tasks.

\section{Theoretical results}\label{sec:mle}

So far, we have seen that bias exists in sequential evaluation, and proposed a parametric model for the evaluation process justified through a best-response argument. This motivates the problem of designing estimators that \emph{correct} for sequential bias under the parametric model. We consider the class of estimators $\estinvperm:\reals^\numitems\rightarrow \setperms_\numitems$ that take as input the scores $\report\in \reals^\numitems$ of the $\numitems$ items, and output an estimate ranking of the $\numitems$ items. This is a general class of offline estimators that are not required to commit to an evaluation of the current item at each timestep $\pos$. For any ranking estimator $\estinvperm$, we use the shorthand $\estrankrel_\pos \defn \rankrel_\pos(\estinvperm)$ to denote the relative rank of item $\pos$ according to the estimated ranking $\estinvperm$.

We assess the estimators in terms of how effective they are at estimating the ground truth ranking $\invpermtrue$, through two ranking error metrics. The first standard metric is the normalized Spearman's footrule error between any ranking $\invperm\in \setperms_\numitems$ and the ground truth ranking $\invpermtrue\in \setperms_\numitems$, given by 
\begin{align*}
\losssf(\invperm, \invpermtrue) \defn \frac{1}{\numitems^2}\normone{\invperm - \invpermtrue}.
\end{align*}
In words, this measures the mean difference of ranks over all the items, and is bounded within a factor $2$ to the normalized Kendall--Tau distance between two rankings~\citep{diaconis1977disarray} (formally described by Eq.~\eqref{eq:kt} in \Cref{sec:proof_notation}). The second metric is the normalized entry-wise absolute error at each position $\pos\in [\numitems]$, given by
\begin{align*}
    \lossentryat{\pos}(\invperm, \invpermtrue) \defn \frac{1}{\numitems}\abs*{\invperm(\pos) - \invpermtrue(\pos)}.
\end{align*}
Note that the Spearman's footrule error is the mean of the entry-wise absolute error over all positions:
\begin{align}\label{eq:relation_err_sf_entrywise}
    \losssf(\invperm, \invpermtrue) = \frac{1}{\numitems}\sum_{\pos=1}^\numitems \lossentryat{\pos}(\invperm, \invpermtrue).
\end{align}

\subsection{Reported scores are inconsistent even in the noiseless setting}

Before we present our estimators for bias correction, let us  motivate the need for bias correction by revisiting (arguably) the most natural estimator for this problem.
This baseline estimator---commonly used in practice---does not attempt any bias correction; it simply uses the ranking induced by the raw scores. As described in the introduction, the problem with such an approach is the lack of calibration in earlier rounds, which induces errors in the estimated ranking. We begin by making this intuition formal.

Let $\estinvperminduced$ denote the ranking induced by the reported scores $\report$.
For example, if the scores are $\report = [0.5, 0.2, 0.7]$, then the induced ranking is $\estinvperminduced(\report) = [2, 1, 3]$. The following result shows that even in the noiseless case, there exist ``bad'' rankings such that this naive estimator incurs constant error.
\begin{proposition}\label{prop:worst_case_induced}
    There exists a universal constant $\const > 0$ such that the following is true.
    Consider the parametric model~\eqref{eq:model_parametric} in the noiseless setting. For any $\numitems \geq 8$, there exists a true ranking $\invpermtrue \in \setperms_\numitems$, such that the error of the ranking $\estinvperminduced$ induced by reported scores is lower bounded as
    \begin{subequations}
    \begin{align}
        \losssf(\estinvperminduced, \invpermtrue) & \ge \const,\label{eq:err_noiseless_induce_sf}\\
        \lossentryat{\pos}(\estinvperm_0, \invpermtrue) & \ge \const\qquad \text{for some position }\pos\in [\numitems].\label{eq:err_noiseless_induce_entrywise}
    \end{align}
    \end{subequations}
\end{proposition}

The proof of this proposition is provided in \Cref{sec:proof_prop_worst_case_induced}. In the proof, we carefully construct a ``bad'' ranking $\invpermtrue$ where the number of conflicts between the induced ranking $\estinvperminduced$ and the true ranking $\invpermtrue$ is quadratic in $\numitems$. The normalized Kendall--Tau distance (which is a constant factor away from the normalized Spearman's footrule distance) incurred by this ``bad'' ranking is thus bounded below by a constant $\const$. The lower bound~\eqref{eq:err_noiseless_induce_entrywise} on the worst-case entry-wise error follows directly from~\eqref{eq:err_noiseless_induce_sf}.
Since the setting is noiseless, we see transparently that the core issue with the induced ranking $\estinvperminduced$ is that it does not take sequential bias into account.

\subsection{Our proposed estimator}

To correct the issue above, we propose an estimator for this problem that attempts to output the ranking that is both consistent with our model and closest to the vector of observations $\report \in \reals^\numitems$. When the noise in the problem is sub-Gaussian, a natural measure of closeness is given by the squared error. Abusing notation slightly, let $\param(\invpermtrue)\in \reals^\numitems$ denote the sequence of noiseless scores given to the items under the true ranking $\invpermtrue$. Formally, we define $\param(\invpermtrue) \defn \{\param(\pos, \rankrel_\pos(\invpermtrue))\}_{\pos\in [\numitems]}$, so that in the parametric model~\eqref{eq:model_parametric}, we have $\param(\invpermtrue) = \big\{\frac{\rankrel_\pos(\invpermtrue)}{\pos+1}\big\}_{\pos\in [\numitems]}$  The least squares estimator is given by
\begin{align}\label{eq:ls}
    \estinvpermls \in \argmin_{\invperm\in \setperms_\numitems}{\normtwo{\report - \param(\invperm)}^2},
\end{align}
where ties are broken arbitrarily.
An immediate advantage of the least squares estimator is that it always yields the correct ranking in the noiseless case under the parametric model, in contrast to the constant error incurred by the induced ranking. To see this, note that in the noiseless setting, the squared loss in Eq.~\eqref{eq:ls} is equal to zero $0$ if and only if the permutation $\invperm$ is the true permutation.

Before analyzing its error performance, let us first address the question of whether the least squares estimator is computable in polynomial time. This is not immediately obvious, since a naive search over permutations to compute the minimizer in Eq.~\eqref{eq:ls} requires $n!$ time. We show that a natural insertion-style algorithm, presented in Algorithm~\ref{algo:ls}, exactly computes the minimizer $\estinvpermls$. Let us describe it in words. Algorithm~\ref{algo:ls} inserts items one-by-one in the same order as they appear, and can therefore be computed in a fully online fashion. The algorithm keeps a sequence $\arr$ of the items so far that are estimated to be in increasing order. At each step $\pos\in [\numitems]$, the algorithm estimates the relative rank $\estrankrel_\pos$ of the current item $\pos$ as a function of its score $\report_\pos$ (Line~\ref{line:compute_rank}), and inserts the item to position $\estrankrel_\pos$ to the sequence $\arr$ (Line~\ref{line:insert}). Such an online insertion step is natural since the squared error decomposes over items. The following result formally establishes this equivalence along with the near-linear time complexity.
\begin{algorithm}[ht]
    \DontPrintSemicolon
    \KwIn{Reported scores of the items $\report \in \reals^\numitems$.}
    \KwOut{Ranking of the items $\estinvperm\in \setperms_\numitems$.}
    Initialize an empty array $\arr=[\;]$.\;
    \ForEach{$\pos\in [\numitems]$\label{line:for_start}}{
       Estimate the relative rank $\estrankrel_\pos\in \argmin_{\rankrel\in [\pos]} \abs*{\report_\pos - \frac{\rankrel}{\pos + 1} }$, where ties are broken arbitrarily.\label{line:compute_rank}\;
       Insert item $\pos$ to be position $\estrankrel_\pos$ in the array $\arr$.\label{line:insert}\;
    }\label{line:for_end}
    Output the ranking of items $\estinvperm=\arr^{-1}$.\label{line:sort}\;
    \caption{Insertion algorithm to compute the least squares estimator.}\label{algo:ls}
\end{algorithm}

\begin{proposition}\label{prop:equivalence}
    For any $\report\in \reals^\numitems$, 
    Algorithm~\ref{algo:ls} runs in $O(\numitems \log \numitems)$ time and computes the least squares estimator~\eqref{eq:ls} exactly.
\end{proposition}
The proof of this proposition is provided in~\Cref{sec:proof_prop_equivalence}. The correctness comes from decomposing the error~\eqref{eq:ls} into an individual term for each timestep $\pos$. The time complexity comes from using an order-statistics tree~\cite[Chapter 14.1]{algorithms2009cormen} as the data structure to perform insertions (Line~\ref{line:insert} of Algorithm~\ref{algo:ls}).

\subsection{Guarantees in the noisy setting}\label{sec:theory_noisy}
Having shown that the least squares estimator is practically computable and outputs the correct ranking in the noiseless setting, we now proceed to analyze its performance in the noisy setting.
We assume that the noise in the model~\eqref{eq:model_parametric} is bounded as $\noise_\pos\in[-\noiseunifparam, \noiseunifparam]$ for some parameter $\noiseunifparam \in [0, 1]$, but is allowed to take arbitrary values unless otherwise specified. Bounded noise is a natural assumption for a finite grading scale.

We first consider the Spearman's footrule error.
Consider any ranking $\invperm$. Recall that 
$\invpermrestrict{\pos}$ denotes the relative ranking restricted to the first $\pos$ items, and let $\mapreltoabs^\pos_\rankrel(\invperm)\defn \invperm\big(\invpermrestrict{\pos}^{-1}(\rankrel)\big)$. Let us describe this quantity in words: at each timestep $\pos$, we identify the position of the item whose relative rank is $\rankrel$, namely $\invpermrestrict{\pos}^{-1}(\rankrel)$. Then we obtain this item's absolute rank $\invperm(\invpermrestrict{\pos}^{-1}(\rankrel))$. The following theorem provides an upper bound on the Spearman's footrule error of the least squares estimator, as a function of any given true ranking $\invpermtrue$. The ceiling function $\ceil*{\cdot}$ means rounding to the least integer greater than or equal to the input value. Recall the shorthand $\rankreltrue_\pos = \rankrel_\pos(\invpermtrue)$.

\begin{theorem} \label{thm:mle_ub}
Let the function $\mapreltoabs^\pos_\rankrel$ be defined as above. There exists a universal constant $\Const > 0$ such that the following is true. Suppose the noise is bounded in the range $[-\noiseunifparam, \noiseunifparam]$. Then for any $\numitems \ge 1$ and any ranking $\invpermtrue\in \setperms_\numitems$, the Spearman's footrule error incurred by the least squares estimator is upper bounded as
\begin{align}\label{eq:ls_ub_adaptive}
    \losssf(\estinvpermls, \invpermtrue) \le \frac{\Const}{\numitems^2} \cdot \sum_{t=1}^\numitems \left(\mapreltoabs^t_{\rankreltrue_\pos + \ceil*{\noiseunifparam (\pos + 1)}}(\invpermtrue) - \mapreltoabs^\pos_{\rankreltrue_\pos - \ceil*{\noiseunifparam (\pos+1)}}(\invpermtrue)\right),
\end{align}
where we define $\mapreltoabs_\rankrel^\pos = \mapreltoabs_\pos^\pos$ for $\rankrel > \pos$, and $\mapreltoabs_\rankrel^\pos = \mapreltoabs_1^\pos$ for $\rankrel < 1$.
\end{theorem}
The proof of this theorem is provided in~\Cref{sec:proof_thm_mle}.
At first glance, the bound~\eqref{eq:ls_ub_adaptive} looks quite intuitive: Each term $\big(\mapreltoabs^\pos_{{\rankreltrue_\pos + \noiseunifparam (\pos + 1)}} - \mapreltoabs^\pos_{{\rankreltrue_\pos - \noiseunifparam (\pos + 1)}}\big)$ (omitting the ceiling and the dependency on $\invpermtrue$ for readability) bounds the error introduced in step $\pos$, by translating from a relative rank error to an absolute rank error. We then sum this error up over $\pos\in [\numitems]$. 
However, the challenge comes from tracking the errors which are intertwined across the steps. To see this, let us first consider a hypothetical scenario where all the insertions were correct (as in the noiseless setting) except for item $\pos$ of relative rank $\rankrel_\pos$. In the setting of bounded noise, item $\pos$ is erroneously inserted to some other position $\rankrelalt\in [\rankreltrue_\pos - \noiseunifparam(\pos + 1), \rankreltrue_\pos + \noiseunifparam(\pos + 1)]$. For simplicity, let us assume $\rankrelalt > \rankreltrue_\pos$. By definition of $\mapreltoabs^\pos_{\rankrelalt}$, item $\pos$ takes the place of the item whose absolute rank is $\mapreltoabs^\pos_{\rankrelalt}$. If all other insertions were correct, then item $\pos$ ends up at position $\mapreltoabs^\pos_{\rankrelalt}$, incurring an error of $(\mapreltoabs^\pos_{\rankrelalt} - \mapreltoabs^\pos_{\rankreltrue_\pos})$ on item $\pos$, which is in turn bounded by $\big(\mapreltoabs^\pos_{\rankreltrue_\pos + \noiseunifparam (\pos + 1)} - \mapreltoabs^\pos_{\rankreltrue_\pos - \noiseunifparam (\pos + 1)}\big)$.
In other words, $\big(\mapreltoabs^\pos_{\rankreltrue_\pos + \noiseunifparam (\pos + 1)} - \mapreltoabs^\pos_{\rankreltrue_\pos - \noiseunifparam (\pos + 1)}\big)$ bounds the error introduced at each timestep $\pos$ by the item being inserted in isolation, \emph{assuming all previous steps were correct}. In reality, the previous insertions and the future insertions can both be noisy. Moreover, each erroneous insertion of an item causes other items to shift. Hence, a more intricate argument is needed to analyze these errors jointly, and it is surprising that the total error stays within a constant of the error as if each insertion were analyzed in isolation. In the proof, we carefully construct intermediate objects that allow us to decompose and track the error over each individual timestep.
Let us now showcase a consequence of Theorem~\ref{thm:mle_ub} by considering a uniform prior on $\invpermtrue$, where we evaluate the so-called Bayes' risk of estimation. We note that some prior work has also jointly considered worst-case and average-case error in ranking problems~\citep{pananjady2020worst}.
\begin{corollary}\label{cor:ls_ub_uniform_prior}
There is a universal constant $\Const > 0$ such that the following is true. Suppose the true ranking $\invpermtrue$ is sampled from $\setperms_\numitems$ uniformly at random, and that the noise is bounded as $\noise_\pos\in[-\noiseunifparam, \noiseunifparam]$. Then the expected Spearman's footrule error of the least squares estimator is bounded as
\begin{align*}
    \Expect[\losssf(\estinvpermls, \invpermtrue)] \le \Const\noiseunifparam,
\end{align*}
where the expectation is taken over the uniform prior on $\invpermtrue$.
\end{corollary}
The proof of this corollary follows straightforwardly from Theorem~\ref{thm:mle_ub} by taking an expectation over~\eqref{eq:ls_ub_adaptive}, and is presented for completeness in~\Cref{sec:cor-proof}.
Interpreting this result a bit more, note that we expect to make an order of $\noiseunifparam$ normalized error even on the last few items, since there is noise in the problem and we are bound to confuse them with their adjacent $\Theta(\numitems\noiseunifparam)$ neighbors. Corollary~\ref{cor:ls_ub_uniform_prior} shows that this scaling behavior remains the same even though the errors compound from previous steps, provided the underlying permutation $\invpermtrue$ is uniform.

The above intuition already suggests that the result of Corollary~\ref{cor:ls_ub_uniform_prior} ought to be optimal in some sense, and this brings us to our next theoretical result: a lower bound on the Spearman's footrule error.
For this lower bound, we consider the class of estimators $\estinvperm:\reals^\numitems\rightarrow \setperms_\numitems$ that take as input the scores $\report\in \reals^\numitems$ of the $\numitems$ items, and output an estimated ranking of the $\numitems$ items.
\begin{theorem}\label{thm:lb_sf}
There exist strictly positive universal constants $(\const, \Const, \noiseunifparam_0)$ such that the following is true. Suppose the true ranking $\invpermtrue$ is sampled uniformly at random from the set of all rankings $\setperms_\numitems$. Suppose the noise terms $\{\noise_\pos\}_{\pos\in [\numitems]}$ are sampled i.i.d. as $\noise_\pos\sample\uniform[-\noiseunifparam, \noiseunifparam]$.
    Then for all $\noiseunifparam < \noiseunifparam_0$ and all $\numitems\ge \frac{\Const}{\noiseunifparam}$, the expected Spearman's footrule error over the class of all estimators is lower bounded as 
        \begin{align*}
            \inf_{\estinvperm}\Expect[\losssf(\estinvperm, \invpermtrue)] \ge \const \noiseunifparam,
        \end{align*}
        where the expectation is taken over the uniform prior on $\invpermtrue$, the uniform noise  $\{\noise_\pos\}_{\pos=1}^\numitems$, and potential randomness in the estimator $\estinvperm$.
\end{theorem}
The proof of this theorem is provided in \Cref{sec:proof_thm_lb_sf}. In the proof, we use inversion vectors (see Appendix \cref{sec:proof_notation} for the formal definition) to show that it suffices to lower bound the error in estimating the relative rank $\estrankrel_\pos$ at each timestep~\citep{knuth1998art}. A lower bound on the error incurred by the relative rank $\estrankrel_\pos$ is derived by viewing the score $\report_\pos$ as a noisy observation of the true relative rank $\rankreltrue$ in the model~\eqref{eq:model_parametric}.
A few comments are in order. First, note that the condition $\numitems\ge \frac{\Const}{\noiseunifparam}$ is necessary, because when  $(\numitems+1)\noiseunifparam<0.5$, namely when $\numitems+1 < \frac{0.5}{\noiseunifparam}$, it can be verified using the parametric model~\eqref{eq:model_parametric} that the least squares estimator incurs an error of $0$.
Second, and as alluded to earlier, the intuition for this lower bound is that each item deviates on an order of $n\noiseunifparam$ positions with respect to its true position, and the error is summed up over timesteps. However, similarly to the upper bound for the least squares estimator, the challenge of formalizing this intuition is to handle the sequential setting, and to show that insertion algorithms cannot leverage the dependencies between different timesteps to reduce the error. Finally, we emphasize that this result is an information-theoretic lower bound that holds unconditionally for all possible estimators.

Having provided bounds on the Spearman's footrule error of our estimator and established their optimality among the insertion algorithms, we now present theoretical results on the entry-wise absolute error.
\begin{theorem}\label{thm:ub_inf} 
There exists a universal positive constant $\Const > 0$ such that the following is true. Suppose the true ranking $\invpermtrue$ is sampled uniformly at random from the set of all rankings $\setperms_\numitems$. Suppose the noise terms $\{\noise_\pos\}_{\pos\in [\numitems]}$ are bounded as $\noise_\pos\in [-\noiseunifparam, \noiseunifparam]$. Then the expected entry-wise absolute error of the least squares estimator is bounded as
\begin{align*}
    \Expect\left[\lossentryat{\pos}(\estinvpermls, \invpermtrue)\right] \le \Const \noiseunifparam\log \numitems\qquad \text{ for all } \pos\in [\numitems].
\end{align*}
\end{theorem}
The proof of this theorem is provided in \Cref{sec:proof_thm_ub_inf}. In the proof, we decompose the error over each individual timestep by constructing another set of intermediate objects that are different from the ones used in the proof of \Cref{thm:mle_ub}. Due to the relation~\eqref{eq:relation_err_sf_entrywise} between the Spearman's footrule error and the entry-wise error, an immediate consequence of Theorem~\ref{thm:lb_sf} is that under the assumptions of that theorem, there exists a universal constant $\const > 0$ such that
\begin{align*}
    \inf_{\estinvperm}\max_{\pos\in [\numitems]} \Expect[\lossentryat{\pos}(\estinvperm, \invpermtrue)]\ge \const\noiseunifparam.
\end{align*}
Thus, the least squares estimator is not only optimal among the class of insertion algorithms with respect to its performance in Spearman's footrule error, but also incurs the optimal error (up to a log factor) on each individual item. Intuitively, the error of our estimator is evenly spread out across all items, instead of being particularly large or small at a small fraction of ``bad'' positions.

Let us provide some intuition for the proof. The main challenge is to show that the error introduced by flips at early insertions does not propagate to later insertions. For example, consider the first two items. If the two items are flipped by the insertion algorithm, then in the final sequence, we expect item $1$ to end up at roughly the true position of item $2$, and item $2$ to end up at roughly the true position of item $1$, incurring a constant error on both items in expectation. However, it is important to note that errors in the insertion only occurs when $(\pos+1)\noiseunifparam \ge 0.5$, namely when $\pos \ge \frac{0.5}{\noiseunifparam}-1$. We show that each insertion error only gives an error of the order $\frac{1}{\pos} \approx \const\noiseunifparam$, and carefully track the propagation of this insertion error over all later timesteps. 

To contextualize Theorem~\ref{thm:ub_inf}, we present our final theoretical result: a lower bound on the entry-wise error incurred by the ranking $\estinvperminduced$ induced by the reported scores.

\begin{proposition} \label{thm:lb_inf} 
There exist strictly positive universal constants $(\numitemslb, \noiseunifparam_0, \const)$ such that the following is true. Suppose the true ranking $\invpermtrue$ is sampled uniformly at random from the set of all rankings $\setperms_\numitems$. Then for all $\noiseunifparam< \noiseunifparam_0$ and $\numitems \ge \numitemslb$, the expected entry-wise error of the ranking $\estinvperminduced$ induced by the reported scores is lower bounded as
\begin{align*}
    \Expect\left[\lossentryat{\pos}(\estinvperminduced,\invpermtrue)\right] \ge \const\qquad\text{ at }\pos=1.
\end{align*}
\end{proposition}
The proof of this theorem is provided in \Cref{sec:proof_thm_lb_inf}. According to the parametric model~\eqref{eq:model_parametric}, the noiseless score of item $1$ is $\frac{1}{2}$, so its observed score is in $[\frac{1}{2} - \noiseunifparam, \frac{1}{2} + \noiseunifparam]$, roughly speaking ``in the middle''. Asymptotically, we expect that a constant fraction of the items have scores higher than the score of item $1$, and a constant fraction of the items have scores lower than the score of item $1$. This incurs a constant error if the true rank of item $1$ is not ``in the middle''. Again, the sub-optimality of the score-induced ranking arises from not taking account into the sequential bias, and the error is particularly problematic at items that are presented earlier; this aligns with what one would expect to see in real world scenarios (e.g.,~\citealp{flores1996elisabeth,bruin2006skating}).

\section{Incorporating prior knowledge}\label{sec:prior}

So far, we have considered the model~\eqref{eq:model_parametric}, where the evaluator has no information about the distribution of the items to appear. However, in many applications, one may expect that evaluators have expectations about the quality of the items based on their prior knowledge. For example, consider an evaluator who has been judging a competition for many years. The evaluator has formed an impression about how good the candidates are expected to be, despite the fact that the quality can still fluctuate from year to year. To incorporate prior knowledge into the model, we assume that the first $\prior$ items represent prior knowledge for some $\numitems_0$, and suppose that the actual evaluation starts from timestep $(\numitems_0+1)$. In other words, the $\pos^\textth$ item in the actual evaluation now arrives at timestep $(\numitems_0+\pos)$, and its relative rank in model~\eqref{eq:model_parametric} is computed with respect to all previous items, consisting of $(\pos-1)$ previous items from the actual evaluation, and $\prior$ items representing prior knowledge. This model for prior knowledge aligns in spirit with the exemplar model of memory~\citep{medin1978context}, which represents knowledge as a collection of previously seen items. This model also captures miscalibration of the evaluators, by letting items from prior knowledge and items from the current evaluation have different quality distributions. For example, if the evaluator has previously seen many items of high quality, then this model suggests that the evaluator is inclined to give lower scores to the current items (for contrast, compare this with an evaluator who has previously seen many items of low quality).

The insertion algorithm presented in Algorithm~\ref{algo:ls} can be naturally generalized to incorporate the prior as well. This generalized algorithm is presented in Algorithm~\ref{algo:prior}. 
We now highlight the difference between Algorithm~\ref{algo:prior} and Algorithm~\ref{algo:ls}. Algorithm~\ref{algo:prior} is given a prior size $\prior \ge 0$. Instead of starting with an empty array representing no prior knowledge, we now initialize an array consisting of the $\prior$ prior items (Line~\ref{line:prior_init}). When finding the insertion position for each item $\pos\in [\numitems]$ in the actual evaluation, we estimate its relative rank $\estrankrel_\pos$ by accounting the fact that noiseless score should be $\frac{\estrankrel_\pos}{\prior+\pos + 1}$ as opposed to $\frac{\estrankrel_\pos}{\pos+1}$ (Line~\ref{line:prior_est}). Finally, after all items are inserted, we have a length-$(\prior+\numitems)$ array. To derive a ranking from this algorithm, we ignore the prior items and only report the ranks of the $\numitems$ items of interest. Note that Algorithm~\ref{algo:prior} reduces to Algorithm~\ref{algo:ls} when there is no prior (i.e., when $\prior=0$).

\newcommand{\dummyitem}{\emptyset}
\begin{algorithm}[htb]
    \DontPrintSemicolon
    \KwIn{Reported scores of the items $\report \in \reals^\numitems$, and size of the prior $\prior \in \naturals$.}
    \KwOut{Ranking of the items $\estinvperm\in \setperms_\numitems$.}
    Initialize a length-$\prior$ array $\arr=[\,\emptyset, \ldots, \emptyset\,]$, where each $\emptyset$ represents a prior item.\label{line:prior_init}\;
    \ForEach{$\pos\in [\numitems]$}{
       Estimate the relative rank $\estrankrel_\pos\in \argmin_{\rankrel\in [\pos]} \abs*{\report_\pos - \frac{\rankrel}{\prior + \pos + 1} }$, where ties are broken arbitrarily.\label{line:prior_est}\;
       Insert item $\pos$ to be position $\estrankrel_\pos$ in the array $\arr$.\;
    }
    Remove all prior items $\emptyset$ from $\arr$.\label{line:removal}\;
    Output the ranking of items $\estinvperm=\arr^{-1}$.\;
    \caption{Insertion algorithm generalized to incorporate prior knowledge.}\label{algo:prior}
\end{algorithm}

It is important to note that Algorithm~\ref{algo:prior} does not require knowing the quality of any prior item, or their comparison to the items in the actual evaluation. Consequently,  Algorithm~\ref{algo:prior} can be applied to multiple evaluators and can accommodate the case where evaluators differ in their prior knowledge. Algorithm~\ref{algo:prior} only requires knowing the prior size $\prior$. In practice, the prior size $\prior$ may be estimated from controlled laboratory experiments where the true ordering is known. In the case where the true ordering is unknown, we envisage that it is still possible to estimate the prior size $\prior$ by cross-validation style approaches. For example, if a pair of evaluators see the same set of items in the same sequence, such as in competitions, one may tune the prior size $\prior$, such that the estimated rankings given by the two evaluators as computed by Algorithm~\ref{algo:prior} are close. We leave it for future work to formalize such a cross-validation procedure.

The prior size $\prior$ controls the the amount of correction performed by Algorithm~\ref{algo:prior}. Specifically, a large prior size means that the evaluator is more experienced and calibrated, and consequently the algorithm performs less correction to the evaluator’s scores. For intuition, consider the extreme case where the prior size $\prior$ approaches infinity. Then each item $\pos$ is inserted where a fraction of $\report_\pos$ items are on its left. The ordering computed by the insertion algorithm thus follows the ordering of the scores $\{\report_\idxitem\}_{\idxitem\in [\numitems]}$ (when ignoring tie-breaking where multiple items receive the same score), and this in turn corresponds to performing little to no correction to the raw scores.

\section{Numerical experiments}\label{sec:simulation}

\begin{figure}[htb]
    \centering
    \includegraphics[width=0.4\linewidth]{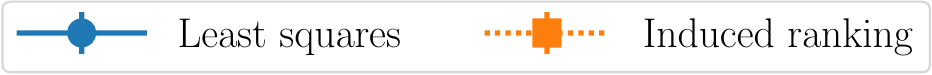}
    
    \subfloat[]{\includegraphics[width=0.4\linewidth]{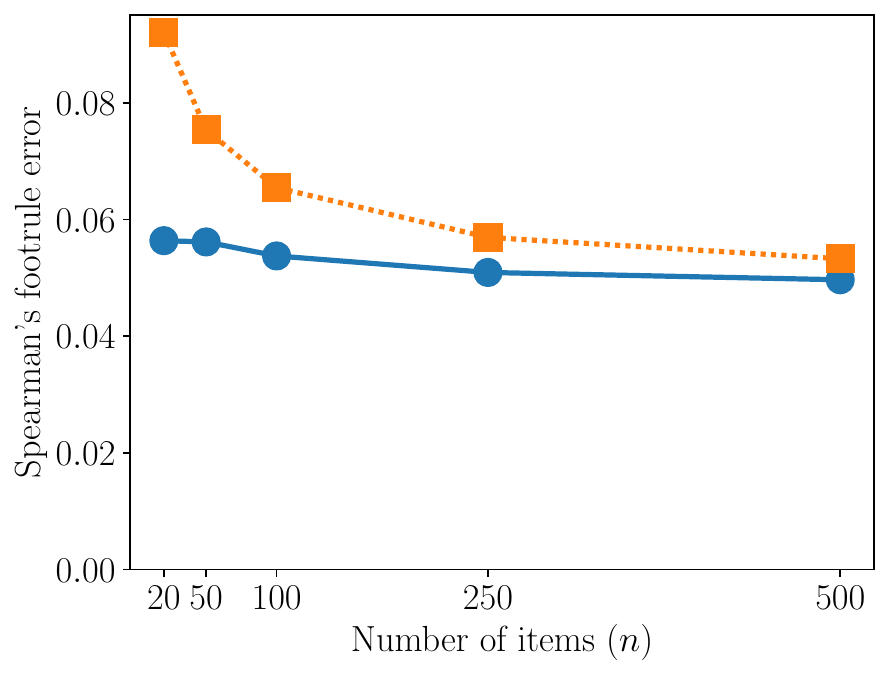}\label{float:vary_n_sf}}
    ~~
    \subfloat[]{\includegraphics[width=0.4\linewidth]{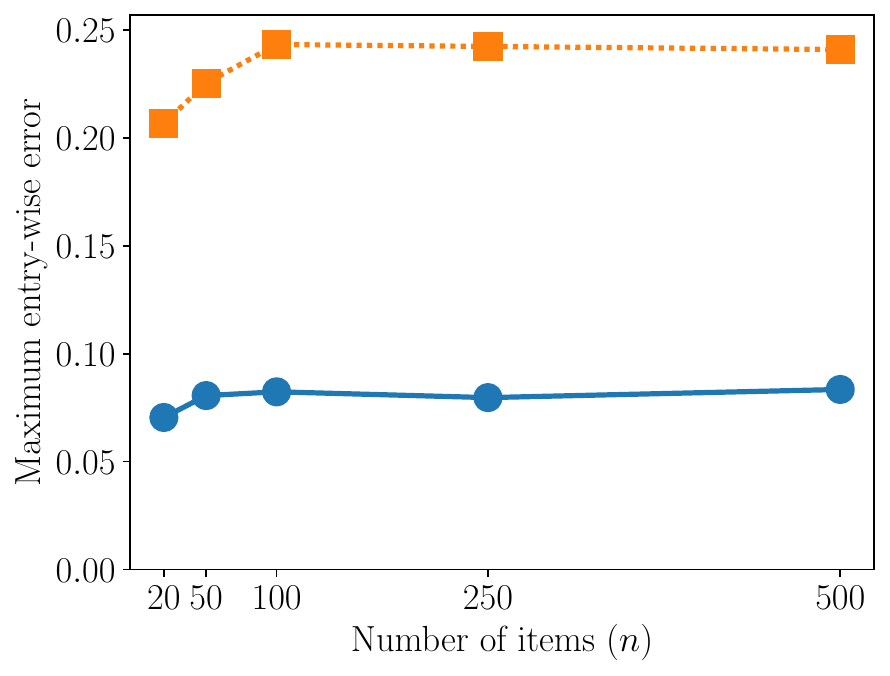}\label{float:vary_n_entrywise}}
    
    \subfloat[]{\includegraphics[width=0.4\linewidth]{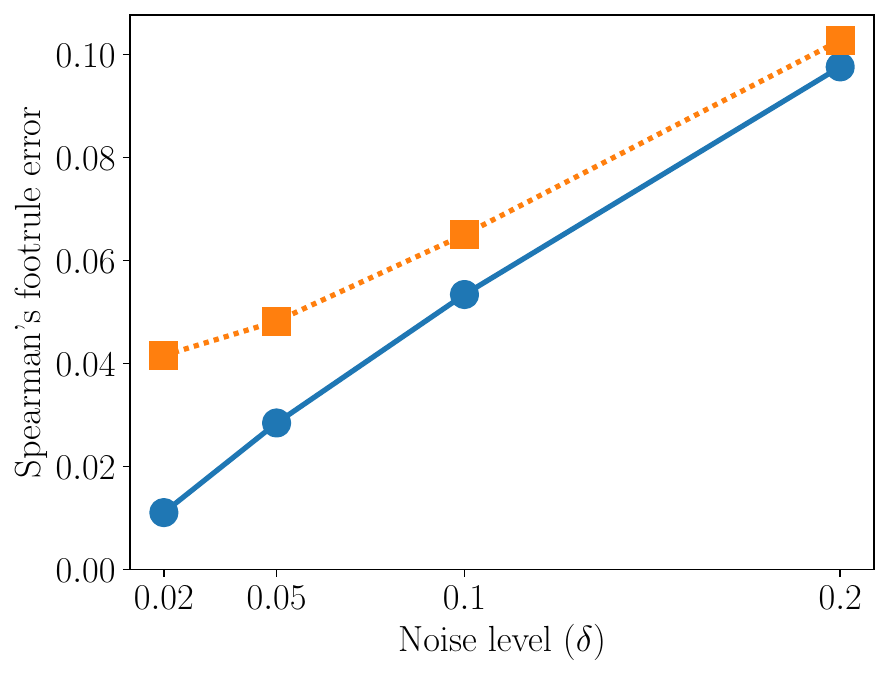} \label{float:vary_delta_sf}}
    ~~
    \subfloat[]{\includegraphics[width=0.4\linewidth]{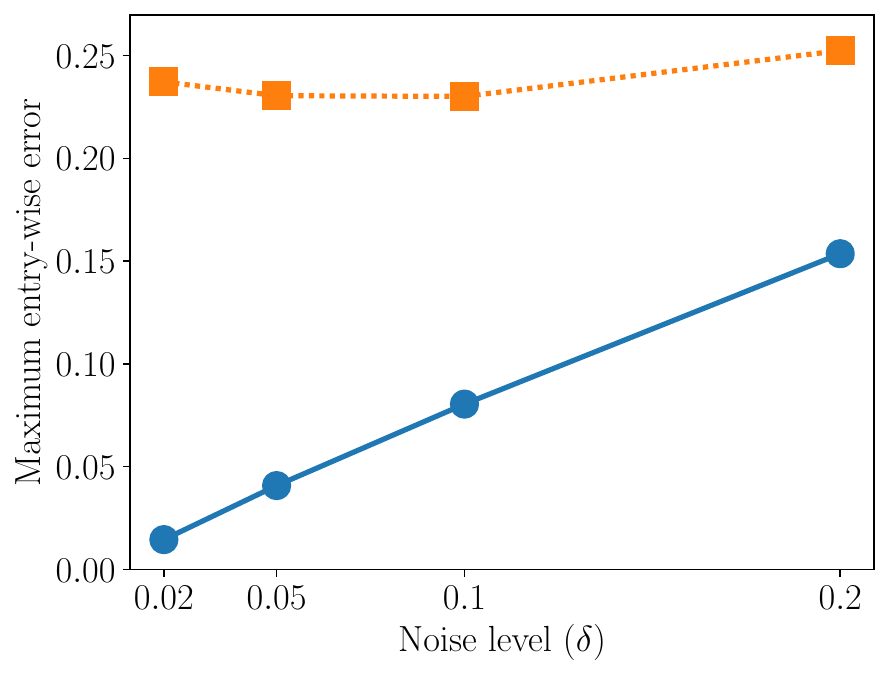} \label{float:vary_delta_entrywise}}

    \caption{The Spearman's footrule error and the maximum entry-wise error of our estimator and the score-induced ranking, varying the values of $\numitems$ and $\noiseunifparam$. Each point is computed over $1000$ runs. Error bars (barely visible) represent standard error of the mean.\label{fig:simulation_vary}}
\end{figure}

\begin{figure}[htb]
    \centering
    \includegraphics[width=0.4\linewidth]{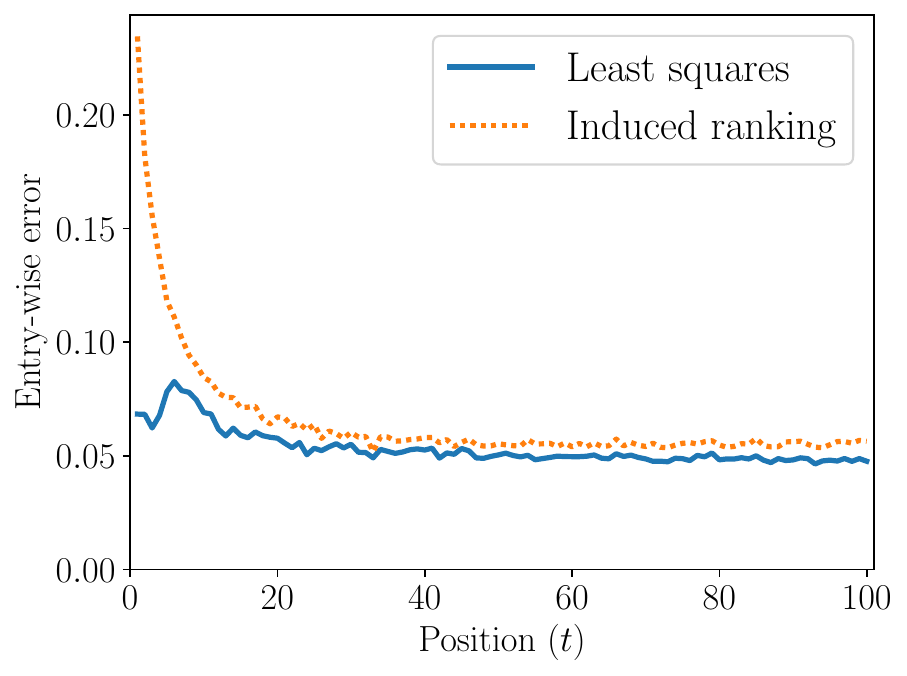}
    \caption{The entry-wise error of our estimator and the score-induced ranking at individual positions, with $\numitems=100$ and $\noiseunifparam=0.1$. Each point is computed over $1000$ runs. Error bars are very small and omitted to avoid clutter.}
    \label{fig:simulation_per_item}
\end{figure}

We now empirically inspect the behavior of our insertion algorithm and compare it with the baseline of using the ranking induced by the reported scores. We first use simulated data and then data from our crowdsourcing experiments that were described in Section~\ref{sec:model_nonparametric}.  Our simulations are carried out assuming prior size zero (i.e., we run Algorithm~\ref{algo:ls}, or equivalently, set $n_0 = 0$ in Algorithm~\ref{algo:prior}), but we equip our algorithm with different prior sizes when using it on crowdsourcing data.

\subsection{Simulation}\label{sec:sim}

In our simulations, we sample an ordering of the $\numitems$ items uniformly at random and generate the scores according to the parametric model~\eqref{eq:model_parametric}, where the noise is sampled i.i.d. from $\uniform[-\noiseunifparam, \noiseunifparam]$.

\paragraph{Dependence on $\numitems$.} We set $\noiseunifparam = 0.1$, and vary the value of $\numitems$. The Spearman's footrule error and entry-wise error of both our estimator and the score-induced ranking are shown in Figure~\ref{fig:simulation_vary}(\subref*{float:vary_n_sf}-\subref*{float:vary_n_entrywise}). To compute the maximum entry-wise error over all positions, we choose the position giving the maximum empirical mean error (over all runs), and plot the error (and its associated standard error of the mean) at this position. We observe that the Spearman's footrule error of both estimators decreases as the number of items $\numitems$ increases. On the other hand, we observe that the maximum entry-wise error increases as a function of $\numitems$ (because the maximum is taken over more items). We observe that our estimator empirically performs better than the score-induced ranking across different values of $\numitems$, especially on the entry-wise error.

\paragraph{Dependence on $\noiseunifparam$.}
We set $\numitems=100$, and vary the value of $\noiseunifparam$. The Spearman's footrule error and the entry-wise error are shown in Figure~\ref{fig:simulation_vary}(\subref*{float:vary_delta_sf}-\subref*{float:vary_delta_entrywise}). The regime of interest is when the noise level $\noiseunifparam$ is low (the left side of each plot). For the Spearman's footrule error (Figure~\ref{fig:simulation_vary}\subref{float:vary_delta_sf}), we observe that the error decreases for both estimators as the noise level $\noiseunifparam$ decreases. Our estimator outperforms the score-induced ranking  especially when the noise level is low, and appears to align with the linear decay indicated by Corollary~\ref{cor:ls_ub_uniform_prior}. For the maximum entry-wise error (Figure~\ref{fig:simulation_vary}\subref{float:vary_delta_entrywise}), the error of our estimator decreases as the noise level $\noiseunifparam$ decreases, whereas the error for the score-induced ranking remains large. These behaviors are consistent with Theorem~\ref{thm:ub_inf} and Proposition~\ref{thm:lb_inf}.

\paragraph{The entry-wise error at each individual position.}
Finally, we inspect the entry-wise error $\Expect[\lossentryat{\pos}]$ at each individual position $\pos\in[\numitems]$. We set $\numitems=100$ and $\noiseunifparam=0.1$. The error at different positions incurred by our estimator and the score-induced ranking is shown in Figure~\ref{fig:simulation_per_item}. For the score-induced ranking, we observe that items appearing early in the sequence incur a larger error compared to items appearing later in the sequence. This large error at earlier positions is significantly reduced by our estimator, which sheds light on its smaller entry-wise error in Figure~\ref{fig:simulation_vary}. We also empirically observe that the maximum error of our estimator is not incurred by the first item as in the case of the score-induced ranking, but instead slightly later (around $\pos=6$). This is an interesting phenomenon worthy of follow-up investigation, because one may have otherwise speculated that the largest error would still be incurred at the first item.

\subsection{Crowdsourcing experiments}\label{sec:expt_crowdsourcing}
\newcommand{\statsbias}{entry-wise difference\xspace}

\begin{figure}[tb]
    \centering
    \hspace{2mm}
    \includegraphics[width=0.3\linewidth]{figures/legend.pdf}\hspace{6mm}
    \includegraphics[width=0.6\linewidth]{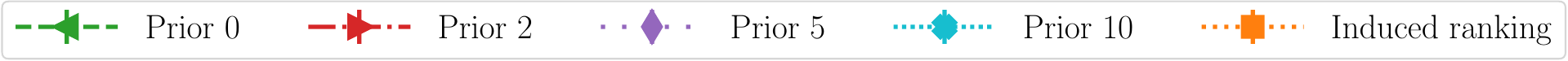}\vspace{3mm}
    
    \includegraphics[width=0.32\linewidth]{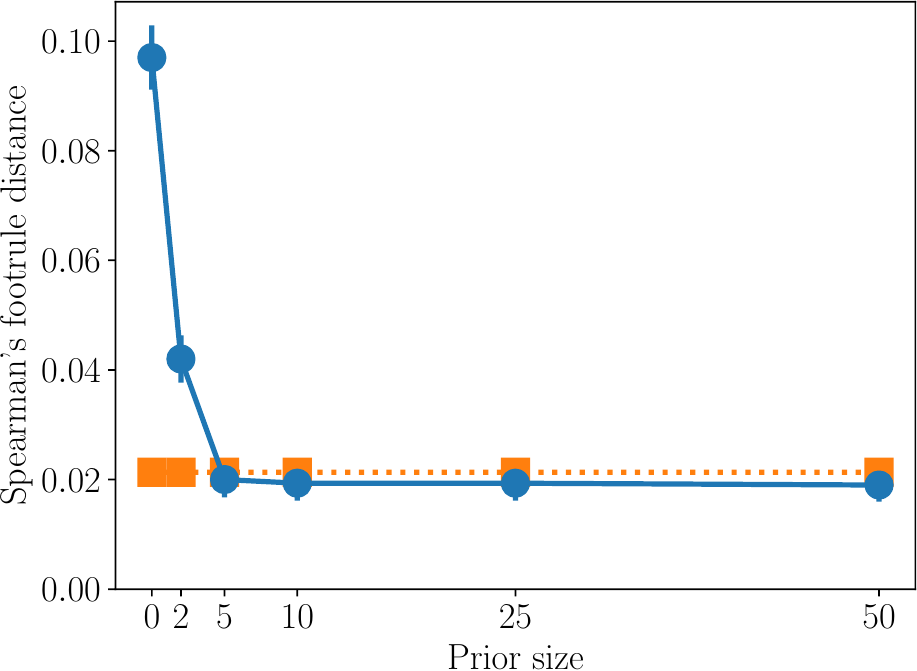}
    \includegraphics[width=0.32\linewidth]{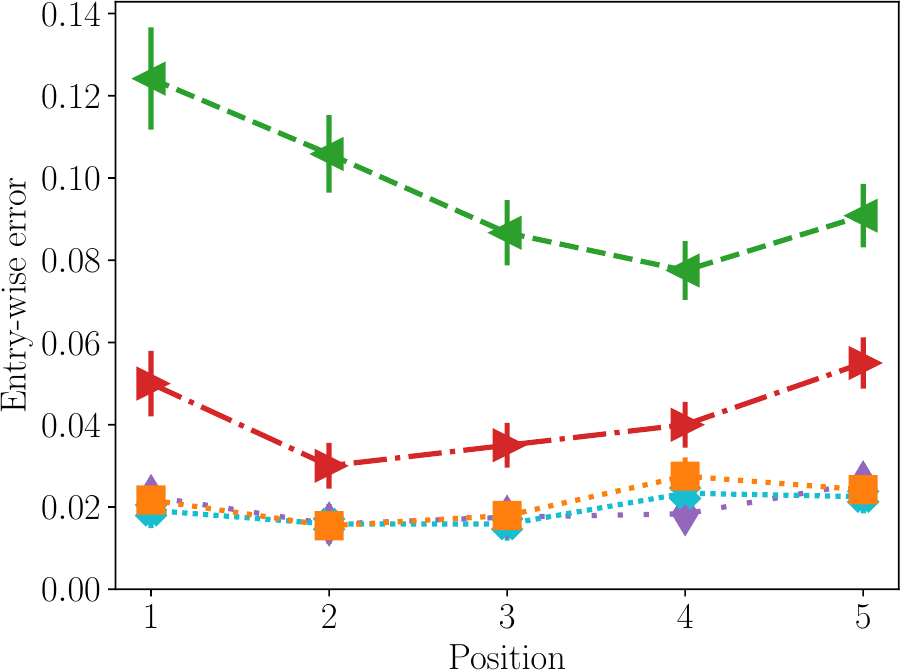}
    \includegraphics[width=0.32\linewidth]{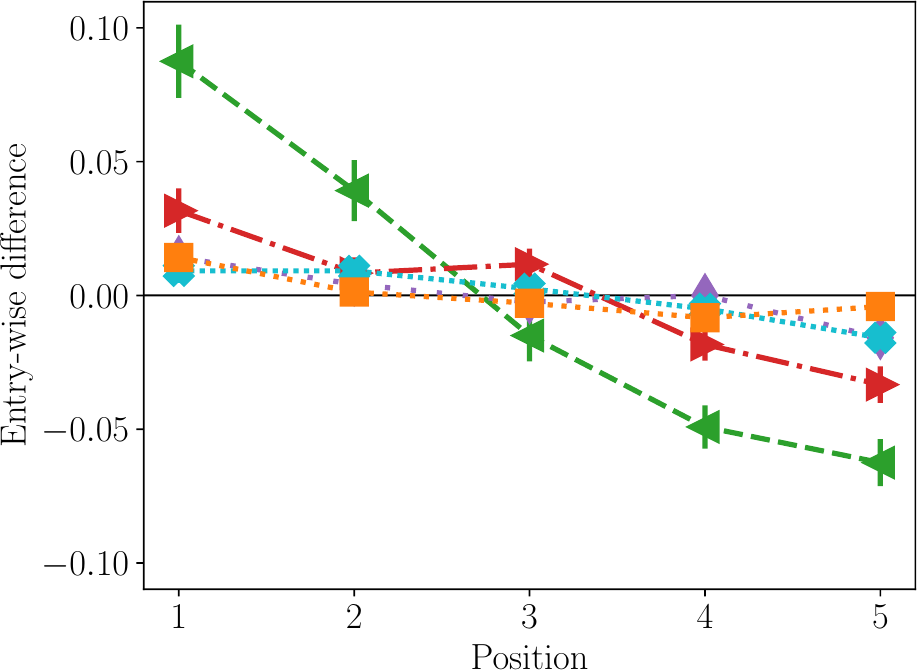}
    \caption{Comparison of of our estimator with various prior sizes and the score-induced ranking, on the first crowdsourcing experiment collected in Section~\ref{sec:expt_existence}. Error bars represent standard error of the mean.\label{fig:expt_existence}}
\end{figure}

\begin{figure}[tb]
    \centering
    \hspace{2mm}
    \includegraphics[width=0.3\linewidth]{figures/legend.pdf}\hspace{6mm}
    \includegraphics[width=0.6\linewidth]{figures/expt_position_legend.pdf}
    \subfloat[Workers in group 1 are presented $5$ small items followed by $5$ large items]{
    \includegraphics[width=0.33\linewidth]{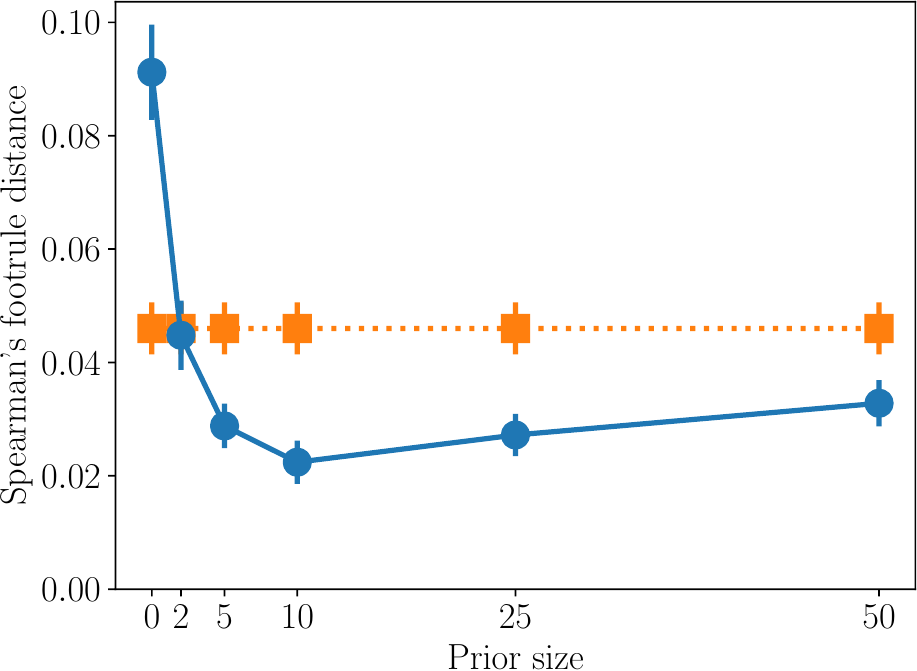}
    \includegraphics[width=0.33\linewidth]{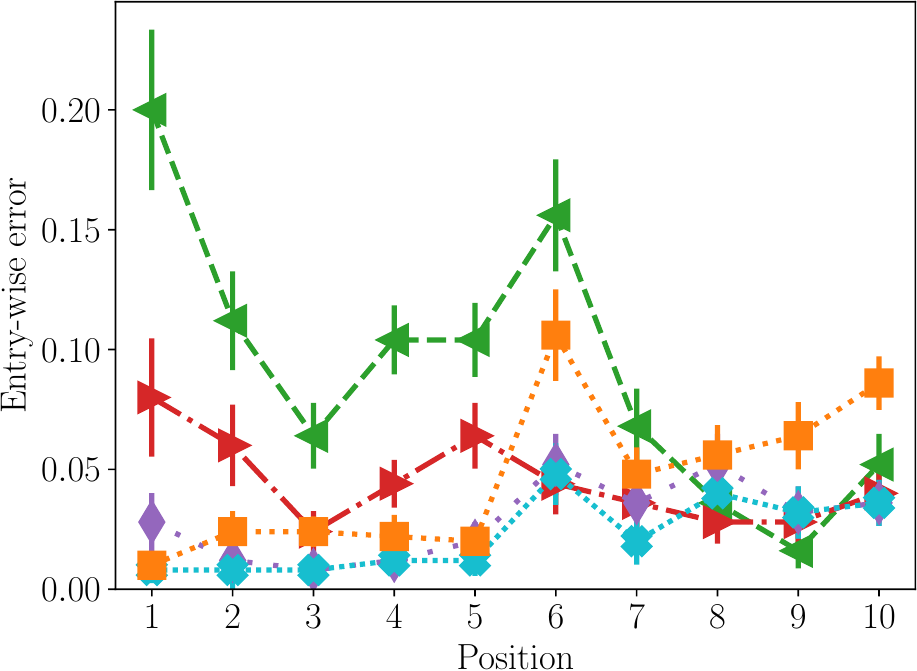}
    \includegraphics[width=0.33\linewidth]{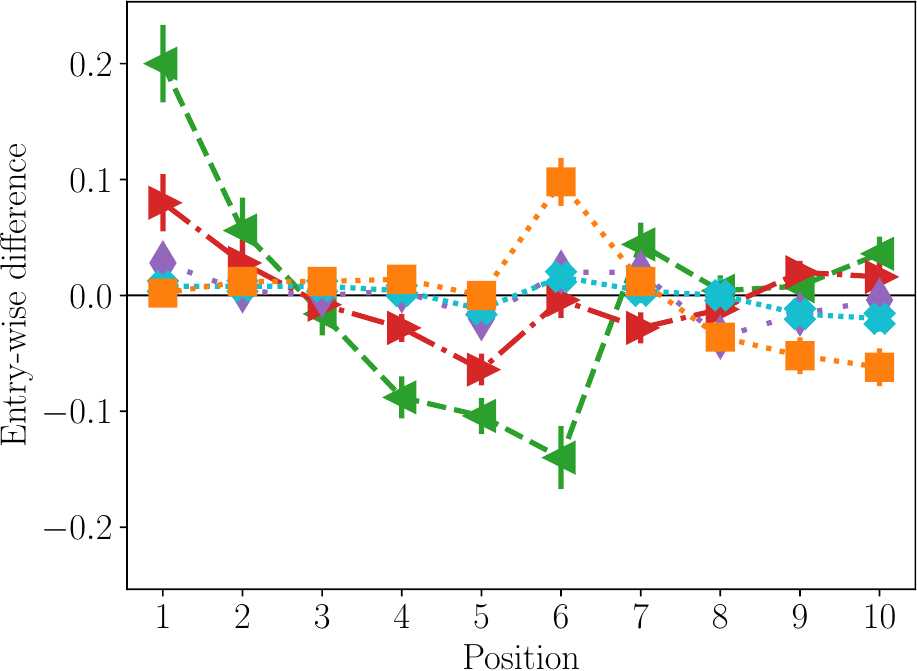}\label{float:expt_relative_lh}}

    \subfloat[Workers in group 2 are presented $10$ large items]{
    \includegraphics[width=0.33\linewidth]{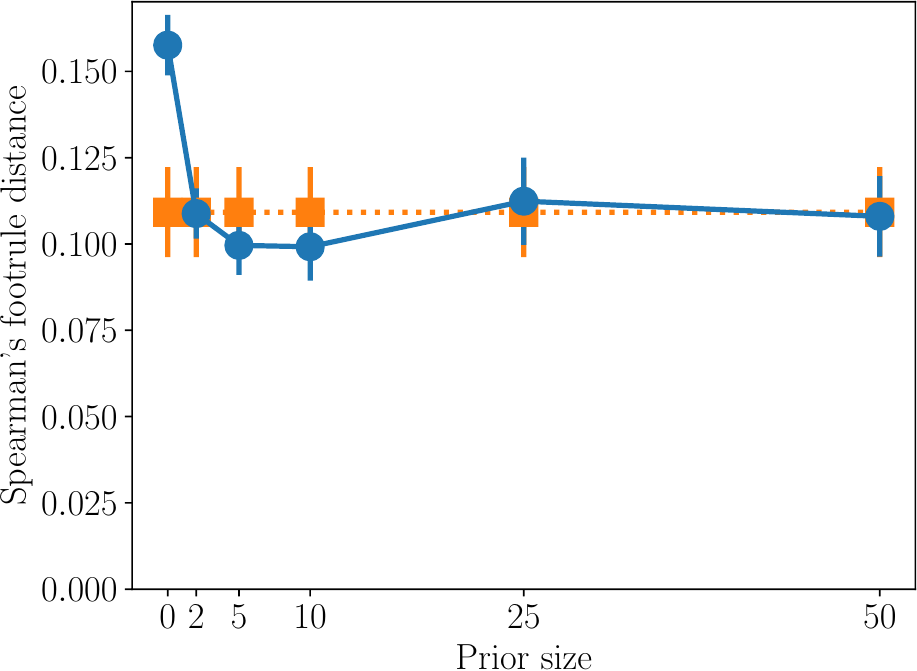}
    \includegraphics[width=0.33\linewidth]{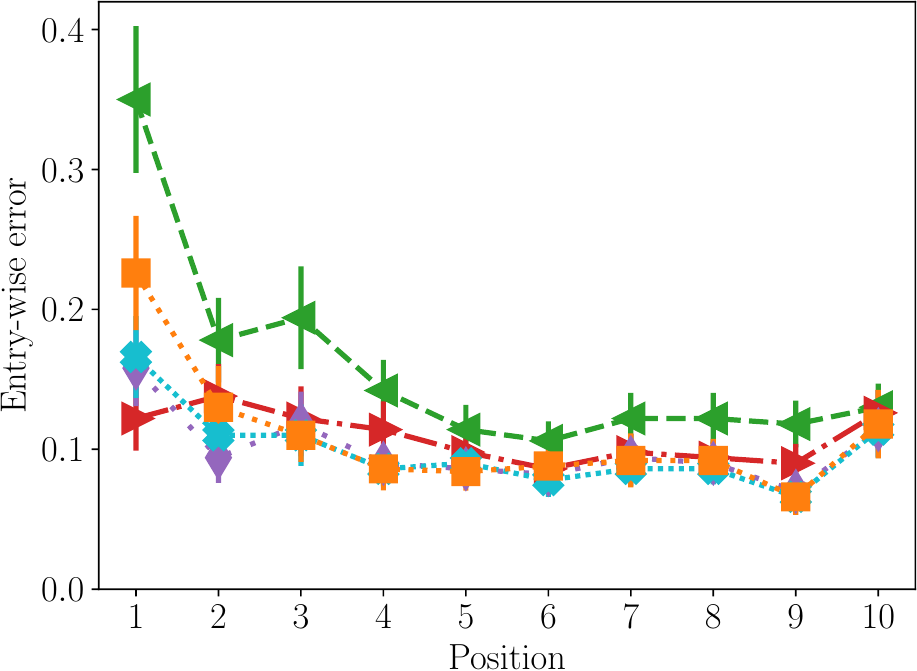}
    \includegraphics[width=0.33\linewidth]{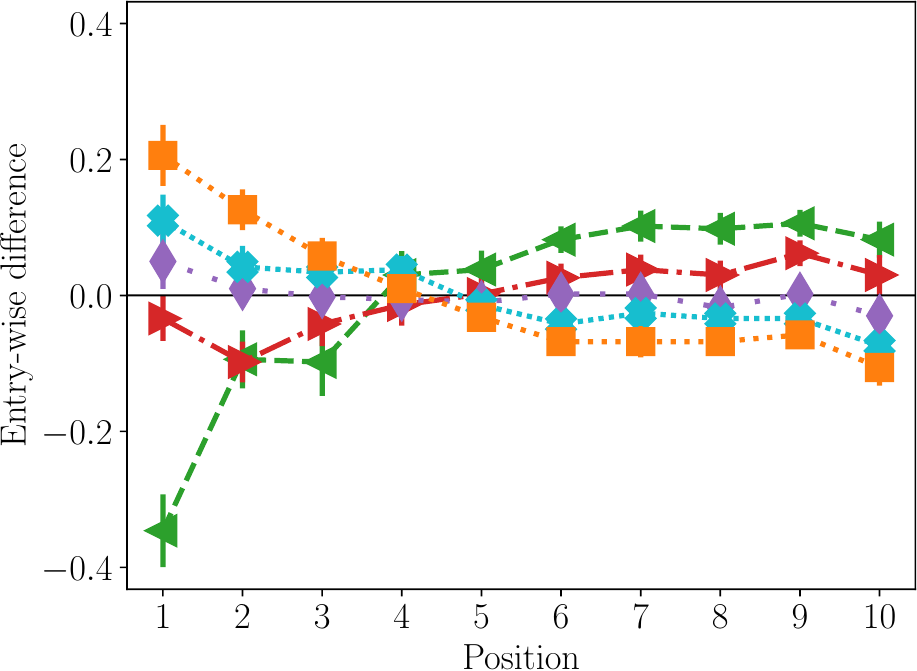}\label{float:expt_relative_hh}}
    \caption{Comparison of of our estimator with various prior sizes and the score-induced ranking, on the second crowdsourcing experiment collected in Section~\ref{sec:expt_relative}. Error bars represent standard error of the mean.\label{fig:expt_relative}}
\end{figure}

\begin{figure}[tb]
    \centering
    \hspace{2mm}
    \includegraphics[width=0.3\linewidth]{figures/legend.pdf}\hspace{6mm}
    \includegraphics[width=0.6\linewidth]{figures/expt_position_legend.pdf}
    \subfloat[Workers in group 1 are presented the ordering $\lbrack 1, 3, 4, 5, 2 \rbrack$ with a conflict]{
    \includegraphics[width=0.33\linewidth]{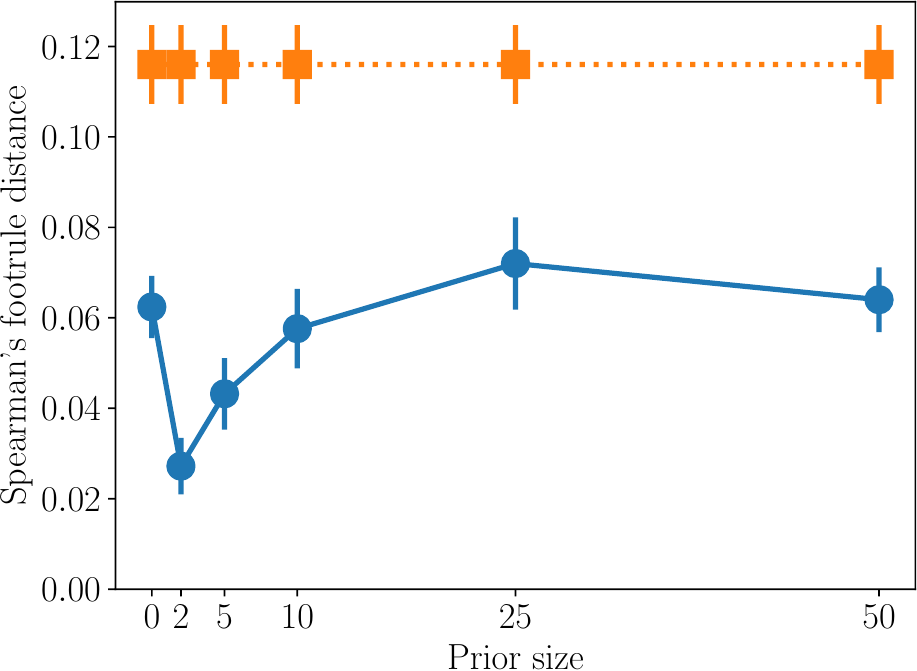}
    \includegraphics[width=0.33\linewidth]{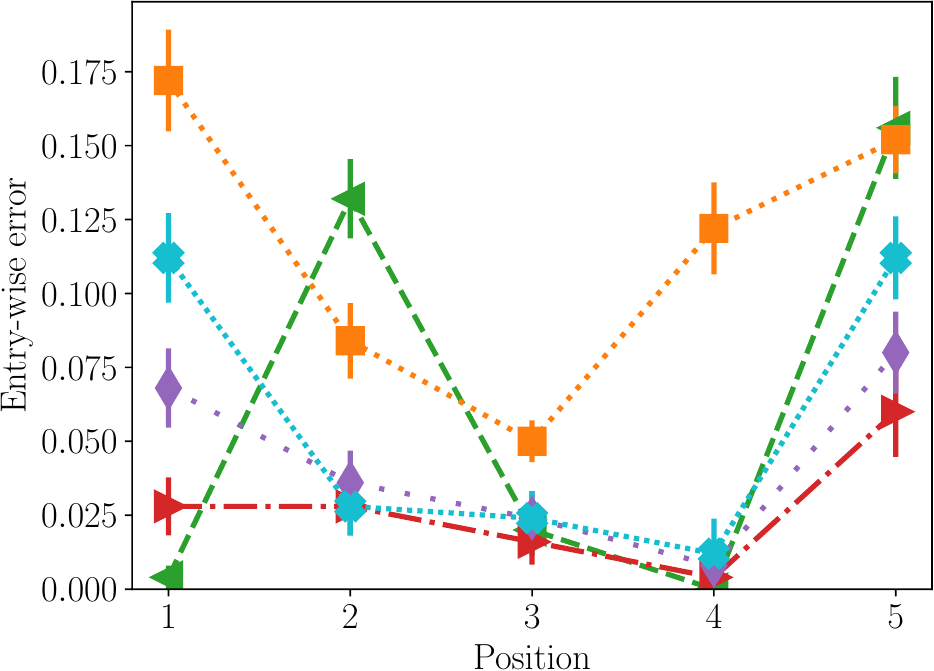}
    \includegraphics[width=0.33\linewidth]{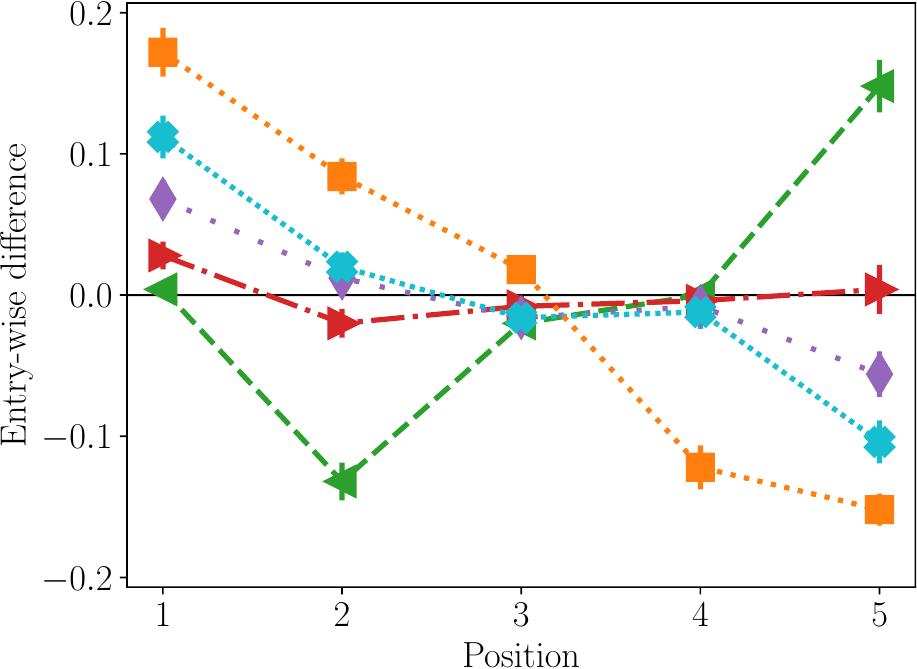}\label{float:expt_conflict_conflict}}
    
    \subfloat[Workers in group 2 are presented the ordering $\lbrack 2, 3, 4, 5, 1\rbrack$ without any conflict]{
    \includegraphics[width=0.33\linewidth]{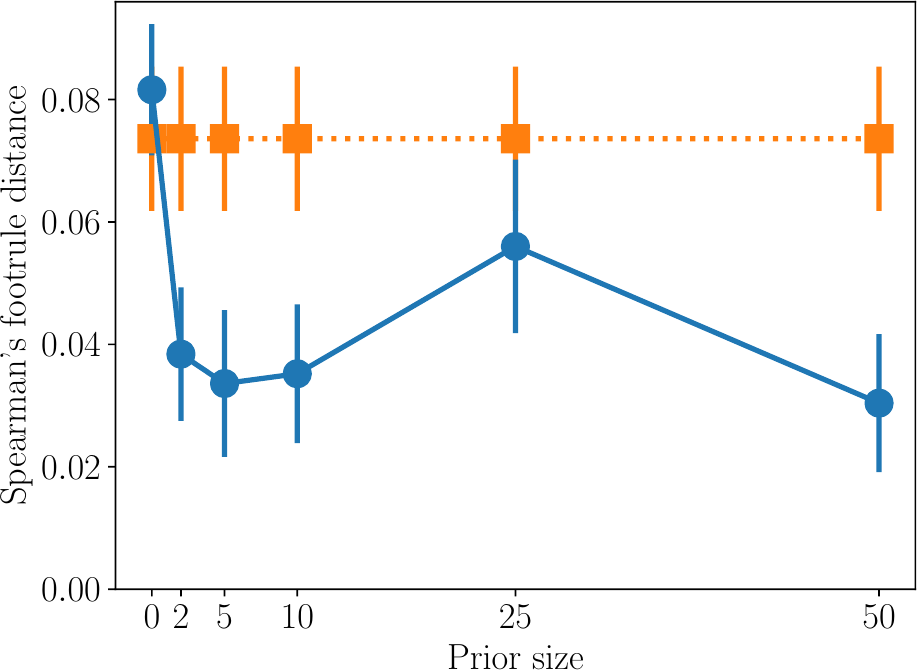}
    \includegraphics[width=0.33\linewidth]{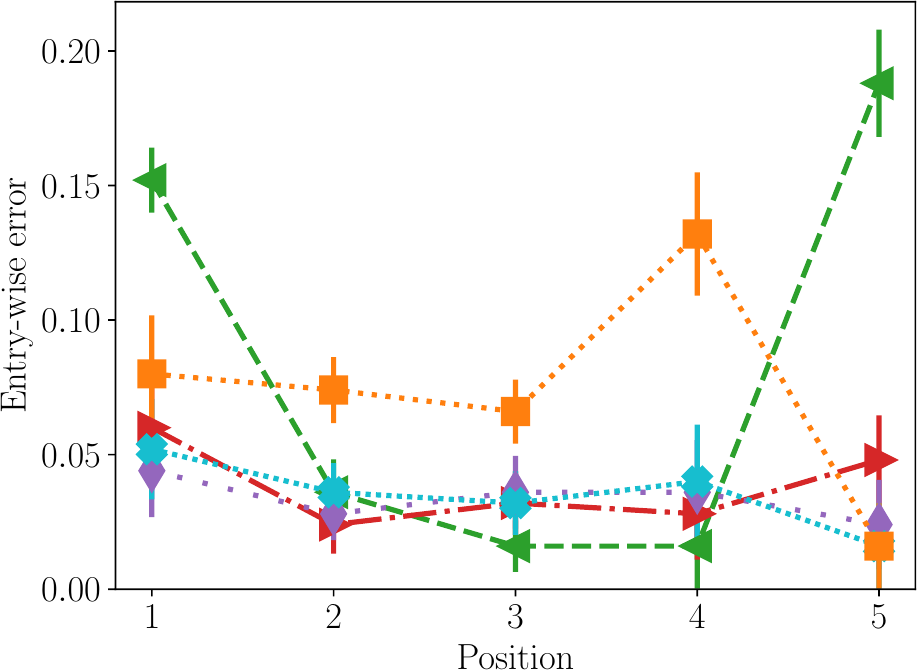}
    \includegraphics[width=0.33\linewidth]{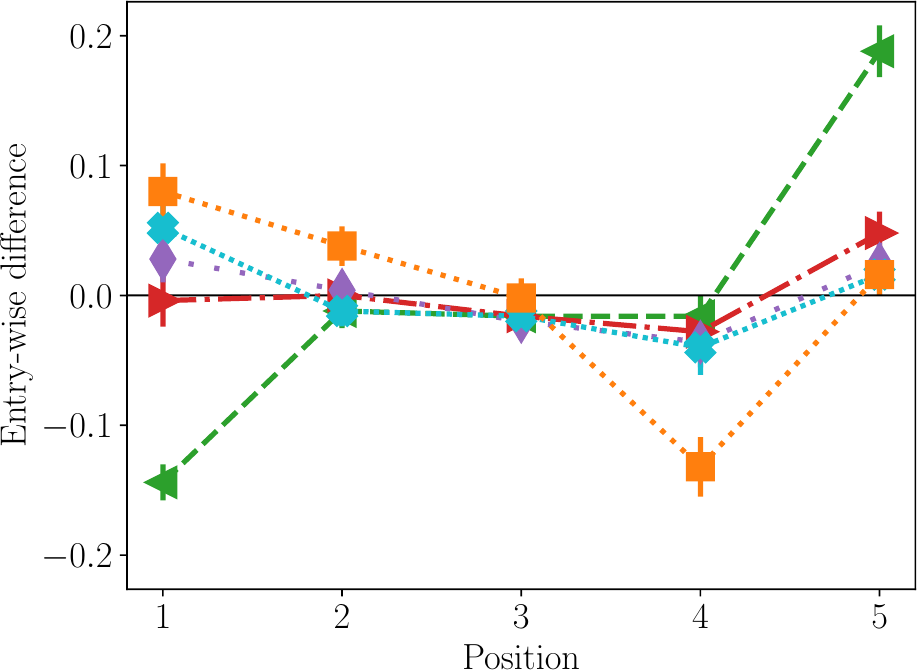}\label{float:expt_conflict_no_conflict}}

    \caption{Comparison of of our estimator with various prior sizes and the score-induced ranking, on the third crowdsourcing experiment collected in Section~\ref{sec:expt_existence}. Error bars represent standard error of the mean.\label{fig:expt_conflict}}
\end{figure}

We now evaluate Algorithm~\ref{algo:prior} using data from the three crowdsourcing experiments described in Section~\ref{sec:model_nonparametric}, and examine the effect of different prior sizes. Recall that the workers answer questions by selecting one of the ten bins representing deciles. To compute the input scores $\report$ to the insertion algorithm, we map these deciles to numerical scores $\frac{1}{11}, \frac{2}{11}, \ldots, \frac{10}{11}$. We then generate ranking estimates $\invperm$ using Algorithm~\ref{algo:prior} for various values of prior size.
The results for the three crowdsourcing experiments are shown respectively in Figure~\ref{fig:expt_existence}, Figure~\ref{fig:expt_relative}, and Figure~\ref{fig:expt_conflict}. In these figures, we examine the Spearman's footrule distance as a function of the prior size (left column), the entry-wise error $\abs*{\invperm(\pos) - \invpermtrue}$ at each position $\idxitem$ (middle column), and the \statsbias $\invperm(\pos) - \invpermtrue(\pos)$ (which could be positive or negative) at each position $i$. 

\paragraph{First experiment -- existence of sequential bias (Figure~\ref{fig:expt_existence}).} We apply Algorithm~\ref{algo:prior} to the data collected in \Cref{sec:expt_existence}, where we uniformly present workers all possible $5!=120$ orderings of $5$ items. We observe that the induced ranking baseline performs well in general, and the insertion algorithm leads to over-correction, especially when the prior size is small. This result aligns with Figure~\ref{fig:bias_position}, where we observe that although workers' scores depend on the position, the variation introduced is relatively small when averaged over all possible rankings.

\paragraph{Second experiment -- Relative nature of scores (Figure~\ref{fig:expt_relative}).} Starting from the left column, we observe that having no prior ($\prior=0$) still yields high error. However, with an appropriate prior size ($\prior=5$ or $\prior=10$), the insertion algorithm incurs smaller Spearman's footrule distance than the induced ranking baseline. Next, move to the middle and right columns of Figure~\ref{fig:expt_relative}\subref{float:expt_relative_hh}. Recall that the workers in group 1 are presented $5$ small items followed by $5$ large items. Hence, the workers incur a high error for the first large item (item 6) by overestimating its quality. The workers in group 2 are presented $10$ items that are all large. Hence, the workers over-estimate the quality of items at earlier positions. For both groups, the algorithm with no prior flips the sign of the error (see the right column). As the prior size increases, the curve for the \statsbias gradually changes from no prior (green curve) towards the induced ranking baseline (orange curve). Hence, the error first decreases, and once the sign becomes the same as the sign of the induced ranking baseline, the error starts to increase.

\paragraph{Third experiment -- Conflicts between ratings and rankings (Figure~\ref{fig:expt_conflict}).}
Starting from the left column once again, we observe that the insertion algorithm with various prior sizes outperforms the induced ranking baseline in Spearman's footrule distance. Next, move to the middle and right columns of Figure~\ref{fig:expt_conflict}\subref{float:expt_conflict_conflict}. Recall that the workers in group 1 are presented the ordering $[1, 3, 4, 5, 2]$ where there is a conflict between the first and the last item. We observe that the error is the highest at these two positions, as expected. For both groups, we again observe that the algorithm with no prior over-corrects in the opposite direction for the \statsbias, compared to the induced ranking baseline.

\section{Discussion}\label{sec:discuss}

Sequential evaluation is a prevalent problem in many real-world scenarios, but is known to be influenced by the sequential bias of evaluators. Motivated by our experimental and theoretical observations, we formulated this problem by modeling non-parametric and parametric relations between the ratings from the evaluators and the relative comparison of the items so far. We highlight our best-response justification for the parametric model, and hope that arguments like these can also be used in other related contexts. Based on the proposed model, we designed a simple and efficient estimator for bias correction and showed that it enjoys several desirable theoretical properties under both noiseless and noisy settings, and in two different metrics.

\paragraph{Limitations and open problems.} The desirable properties of our estimator notwithstanding, we emphasize that these results do not imply that practitioners should unconditionally apply our estimator in lieu of using score-induced rankings (or other existing methods) in every setting. In particular, care is required to verify that the simplifications made by our model are indeed reasonable.
A first instance of our simplifications is alluded to in \Cref{sec:model_nonparametric}. We assume that the evaluator ``inserts'' each item according to its true relative rank $\rankreltrue_\pos$.
In reality, the rating may not only depend on such underlying ranking information about the items, but also the scores the evaluator \emph{has already given} to the previous items. For example, at a later point of the evaluation, the evaluator may realize mistakes they have made when evaluating earlier items, but decide nevertheless to follow a grading scale consistent with these mistakes in the interest of fairness. 
A second simplification of our model was that we ignored other sources of biases (such as the effect of anchoring to previous items discussed in Section~\ref{sec:related_work}) that compound the calibration issues at play.
More generally, even in a non-sequential setting, there are challenges such as the heterogeneity of the evaluators (where evaluators' rating behaviors exhibit inter-personal differences, such as different levels of noise or domains of expertise) and subjectivity (where evaluators may not unanimously agree with an underlying ranking even without noise, and hence their opinions need to be aggregated in a voting fashion). Our model captures one type of heterogeneity due to different prior knowledge of the evaluators, and it would be useful to derive principled aggregation methods that take as input the scores from multiple evaluators, and jointly infer these priors and the true ranking of the items.

Due to these reasons, we suggest using our estimator as an auxiliary tool to guide practical evaluation instead of ``substituting'' it in place of existing methods. For example, in hiring, the ranking produced by our estimator can be used as extra information to guide discussion: If our estimator is inconsistent with the candidates' raw scores, it may indicate that the case should be inspected further. For competitions, it may be perceived as unfair not to give awards based on the raw scores, but the information gleaned from our estimator can be useful in contextualizing the results of such competitions in the  future, e.g., when recruiting new candidates based on their performance history in such competitions.

Orthogonal to the design of estimators, another important aspect in improving accuracy and fairness in evaluation processes is the design of mechanisms. For instance, in many applications, selection is decomposed hierarchically into multiple rounds (e.g., preliminary and final). In such settings, the earlier rounds can provide rough, partial information about the ordering of the items (say, computed from our estimator), which may prove helpful in adaptively designing a sensible sequence of presentation in later rounds. A second design question is to construct the reporting scale in data elicitation jointly with the sequential model. In our experiments for example, we divide the bins into deciles of $0\text{--}10\%$, $10\text{--}20\%$, etc. for simplicity. However, other scales may be more effective, and different scales may be designed for different goals~\citep{garg2019binary}, such as selecting the top-K candidates or estimating a total ranking among them. Moreover, in light of the sequential nature of the task, one can also consider an adaptive scale that changes over the course of the task as evaluators accumulate more information.

\section{Proofs}

In this section, we present the proofs of our theoretical results.

\subsection{Notation and preliminaries}\label{sec:proof_notation}
For notational simplicity, we let $\posplusone\defn \pos+1$, and use the shorthand $x_{1:n} = \{x_i\}_{i\in [n]}$ and $x^{(1:n)} = \{x^{(i)}\}_{i\in [n]}$. We use the notation $f(x) \lessorder g(x)$ to denote that there exists some universal positive constant $\const > 0$, such that $f(x) \le \const\cdot g(x)$, and use the notation $f(x) \gtrorder g(x)$ when $g(x) \lesssim f(x)$.

Let $\invperm, \invpermtrue\in \setperms_\numitems$ be any two rankings of $\numitems$ items. The normalized Kendall--Tau distance between $\invperm$ and $\invpermtrue$ is defined as
\begin{align}
    \losskt(\invperm, \invpermtrue) \defn \frac{1}{\numitems^2}\sum_{\substack{
        \idxitem, \idxitemalt\in [\numitems]\\
        \invpermtrue(\idxitem) < \invpermtrue(\idxitemalt)
    }} \indicator\{\invperm(\idxitem) > \invperm(\idxitemalt)\}.\label{eq:kt}
\end{align}
In words, the Kendall--Tau distance counts the number of pairwise comparisons on which the rankings $\invperm$ and $\invpermtrue$ do not agree. We call each such pairwise comparison a ``flip''. The normalized Spearman's footrule distance and the normalized Kendall--tau distance are bounded within a factor of $2$ from each other~\citep{diaconis1977disarray}. Hence, we consider the Kendall--tau distance in some of our proofs.

We now define another metric between a pair of rankings. Recall that $\rankrel_\pos(\invperm)$ denotes the relative rank of item $\pos$ (out of $\pos$ items) according to the ranking $\invperm$. We consider the $\numitems$-dimensional vector $\rankrelvec(\invperm) \defn \{\rankrel_\pos(\invperm)\}_{\pos\in [\numitems]}$. This vector $\rankrelvec(\invperm)$ can be seen as a version of the ``inversion vector'' (also called the ``inversion table'')~\citep{knuth1998art}. We define the (normalized) $\ell_1$ distance on the inversion vectors between two rankings as:
\begin{align}
    \lossinvvec(\invperm, \invpermtrue) \defn \frac{1}{\numitems^2} \sum_{\pos=1}^\numitems \normone{\rankrelvec(\invperm) - \rankrelvec(\invpermtrue)}.\label{eq:dist_inv_vec}
\end{align}

Let $\invpermtrue$ be a permutation sampled uniformly at random from the set of all permutations $\setperms_\numitems$. Recall from Section~\ref{sec:opt_response} that $\invpermtrue_{[\pos]}$ denotes the ranking restricted to the first $\pos$ items according to $\invpermtrue$. 
It can be verified that marginally for each $\pos\in [\numitems]$, the distribution of $\invpermtruerestrict{\pos}$ is equivalent to sampling the relative ranks $\rankreltrue_{1:\pos}$, where each $\rankreltrue_\idxitem$ is sampled uniformly at random from $[\idxitem]$ and independent from all else.

Next, we describe the construction of some intermediate quantities that are used in the proofs of Theorem~\ref{thm:mle_ub} and Theorem~\ref{thm:ub_inf} for our estimator $\estinvpermls$. For an explicit example, assume that the true ranking $\invpermtrue$ of the items follows
\begin{subequations}\label{eq:ub_mle_example}
\begin{align}\label{eq:ub_mle_example_true}
    \invpermtrue(1) = 40, 
    \quad \invpermtrue(2) = 10, 
    \quad \invpermtrue(3) = 60,
\end{align}
then we have
\begin{align*}
    \invpermtrue_{[3]} = [2, 1, 3],
\end{align*}
meaning the true ranks of the first three items (restricted to the three items) are $2, 1$ and $3$, respectively. Suppose that (eventually) an estimator $\estinvperm$ outputs the absolute ranks for these items as
\begin{align*}
    \estinvperm(1) = 75,\quad \estinvperm(2) = 85, \quad\estinvperm(3) = 25.
\end{align*}
Then the estimated ranking restricted to the first three items are given by
\begin{align*}
    \estinvpermrestrict{3} & = [2, 3, 1].
\end{align*}
\end{subequations}
At each timestep $\pos$, we construct the vectors $\objtrue_{[\pos]}$ and $\objest_{[\pos]}: [\pos]\rightarrow [\numitems]$ as follows:
\begin{subequations}\label{eq:ub_construct_intermediate}
\begin{align}
    \objtrue_{[\pos]} & = \invpermtrue \circ (\invpermtruerestrict{\pos})^{-1},\\
    \objest_{[\pos]} & = \invpermtrue \circ (\estinvpermrestrict{\pos})^{-1}.\label{eq:ub_construct_intermediate_est}
\end{align}
\end{subequations}
In words, the vector $\objtrue_{[\pos]}$ collects the absolute ranks of the $\pos$ items sorted in increasing order. In the example~\eqref{eq:ub_mle_example_true}, we have
\begin{align*}
    \objtrue_{[3]} & = [10, 40, 60].
\end{align*}
On the other hand, the vector
$\objest_{[\pos]}$ is constructed by sorting the $\pos$ items according to $\estinvpermrestrict{\pos}$, and then taking their true absolute ranks. In the example~\eqref{eq:ub_mle_example}, the smallest item in $\estinvpermrestrict{3}$ is item $3$ whose true absolute rank is $60$; the second smallest item in $\estinvpermrestrict{3}$ is item $1$ whose true absolute rank is $40$, etc.:
\begin{align*}
    \objest_{[3]} & = [60, 40, 10].
\end{align*}
Note that we have $\invpermtrue_{[\numitems]} = \invpermtrue$ and $\estinvperm_{[\numitems]} = \estinvperm$. Hence, by the definitions~\eqref{eq:ub_construct_intermediate} we have
\begin{subequations}\label{eq:obj_n}
\begin{align}
    \objtrue_{[\numitems]} & = \identityrank_\numitems\\
    \objest_{[\numitems]} & = \invpermtrue \circ (\estinvperm)^{-1},
\end{align}
\end{subequations}
where $\identityrank_\numitems$ denotes the identity ranking $[1, 2, \ldots, \numitems]$.
\qed

\subsection{Proof of Proposition~\ref{prop:conflict_exist_nonparametric}}\label{sec:proof_prop_conflict_exist_nonparametric}
Note that the mean score of an item $\param(\pos, \rankreltrue_\pos)$ depends only on its position $\pos$ and its relative rank $\rankreltrue_\pos$ among previous items, independent of later items. Accordingly, to prove the claim for $\numitems\ge 4$, it suffices to consider $\numitems=4$. We consider the following two rankings:
\begin{align*}
    \invpermtrue_1 = & [1, 3, 4, 2]\\
    \invpermtrue_2 = & [4, 1, 2, 3].
\end{align*}
Assume for contradiction that there is no conflict in either ranking. In ranking $\invperm_1$, the first item (of relative rank $\rankrel_1(\invpermtrue_1) = 1$) and the fourth item (of relative rank $\rankrel_4(\invpermtrue_1) = 2$) suggests:
\begin{subequations}
\begin{align}\label{eq:conflict_contradiction_one}
    \param(1,1) < \param(4,2).
\end{align}
On the other hand, in ranking $\invpermtrue_2$, the first item (of relative rank $\rankrel_1(\invpermtrue_2) = 1$) and the fourth item (of relative rank $\rankrel_4(\invpermtrue_2) = 3$) suggests: 
\begin{align}\label{eq:conflict_contradiction_two}
    \param(1,1) > \param(4,3).
\end{align}
\end{subequations}
Combining~\eqref{eq:conflict_contradiction_one} and~\eqref{eq:conflict_contradiction_two} yields $\param(4,3) < \param(4,2)$, contradicting the monotonic assumption~\eqref{eq:model_constraint}.
\qed

\subsection{Proof of Proposition~\ref{lem:incentive_rank}}\label{sec:lem_incentive_rank}
In this proof, we assume $\numitems$ is unknown to the evaluator (but non-random). The case where $\numitems$ is known follows from straightforward modifications to this proof. 

Denote the normalized absolute rank of item $\pos$ by
\begin{align}\label{eq:incentive_def_normalized_rank}
    \rankrand_\pos \defn \frac{\invpermtrue(\pos)}{\numitems+1}.
\end{align}
We define the class of deterministic estimators $\classdet$ as follows. Each estimator $\estscore\in \classdet$ has the form of $\estscore = \{\estscore_\pos\}_{\pos\in [\numitems]}$, where each $\estscore_\pos: \setperms_\pos \mapsto \reals$  is a deterministic function that takes as input the relative ranking $\invpermtruerestrict{\pos}$ restricted to the $\pos$ items so far, and outputs a real value. In what follows, we first derive the optimal deterministic estimator among the class $\classdet$ of all deterministic estimators. Then we show that additional randomness cannot improve the performance of these estimators.

\noindent\textbf{Step 1: Decompose the error of deterministic estimators by linearity of expectation.}
By assumption, the true ranking $\invpermtrue$ is drawn uniformly at random from the set of all permutations $\setperms_\numitems$. For any deterministic estimator $\estscore = \{\estscore_\pos\}_{\pos\in [\numitems]}$, we decompose its error~\eqref{eq:err_est_score} as
\begin{align}
    \Expect_{\invpermtrue} \left[\sum_{\pos=1}^\numitems \left(\estscore_\pos(\invpermtruerestrict{\pos}) - \rankrand_\pos\right)^2\right]
    & = \sum_{\pos=1}^\numitems\Expect_{\invpermtrue} \left(\estscore_\pos(\invpermtruerestrict{\pos}) - \rankrand_\pos \right)^2\nonumber \\
    & \stackrel{\stepone}{=} \sum_{\pos=1}^\numitems\Expect_{\rankrand_{1:\numitems}} \left(\estscore_\pos(\invpermtruerestrict{\pos}) - \rankrand_\pos\right)^2,\label{eq:incentive_expression_single_step}
\end{align}
where step~\stepone is true because the random objects $\invpermtrue$ and $\rankrand_{1:\numitems}$ are measurable with respect to each other.
Now we analyze each term in the expression~\eqref{eq:incentive_expression_single_step}.

\paragraph{Decoupling from the future.} For each $\pos$, we have
\begin{align*}
    \Expect_{\rankrand_{1:\numitems}} \left(\estscore_\pos (\invpermtruerestrict{\pos}) - \rankrand_\pos\right)^2 
    & = \Expect_{\rankrand_{1:\pos}}\; \Expect_{\rankrand_{(\pos + 1):\numitems} \given \rankrand_{1:\pos}} \left(\estscore_\pos (\invpermtruerestrict{\pos}) - \rankrand_\pos\right)^2 \\
    & \stackrel{\stepone}{=} \Expect_{\rankrand_{1:\pos}} \left(\estscore_\pos (\invpermtruerestrict{\pos}) - \rankrand_\pos\right)^2,
\end{align*}
where step~\stepone is true because the expression $\left(\estscore_\pos (\invpermtruerestrict{\pos}) - \rankrand_\pos\right)^2$ is independent of $\rankrand_{(\pos+1):\numitems}$ conditional on $\rankrand_{1:\pos}$, i.e., since the estimator only depends on the observations so far, its error at the current timestep is independent of the future conditional on the past.

\paragraph{Decoupling from the past.}
Proceeding from above, we have
\begin{align}
    \Expect_{\rankrand_{1:\pos}}\left(\estscore_\pos (\invpermtruerestrict{\pos}) - \rankrand_\pos\right)^2 & = \Expect_{\invpermtruerestrict{\pos}}\Expect_{\rankrand_{1:\pos}\given \invpermtruerestrict{\pos}} \left(\estscore_\pos(\invpermtruerestrict{\pos}) - \rankrand_\pos\right)^2 \nonumber\\
    & \stackrel{\stepone}{=} \Expect_{\invpermtruerestrict{\pos}}\Expect_{\rankrand_\pos \given \invpermtruerestrict{\pos}}\left(\estscore_\pos(\invpermtruerestrict{\pos}) - \rankrand_\pos\right)^2,\label{eq:incentive_expression_simplify}
\end{align}
where step~\stepone follows because the the expression $\left(\estscore_\pos (\invpermtruerestrict{\pos}) - \rankrand_\pos\right)^2$ is independent of $\rankrand_{1:\pos-1}$ conditional on $(\invpermtruerestrict{\pos}, \rankrand_\pos)$.

\noindent\textbf{Step 2: Compute the minimizer of~\eqref{eq:incentive_expression_simplify} for deterministic estimators.}
Note that $\estscore_\pos(\invpermtruerestrict{\pos})$ is a deterministic function of $\invpermtruerestrict{\pos}$ by the definition of the class $\classdet$. By completing the square, the minimizer to~\eqref{eq:incentive_expression_simplify} is
attained at
\begin{align}
    \estscore_\pos(\invpermtruerestrict{\pos}) & = \Expect_{\rankrand_\pos\given \invpermtruerestrict{\pos}}[\rankrand_\pos].\label{eq:decouple_from_past}
\end{align}
We now analyze the value of $\Expect_{\rankrand_\pos\given \invpermtruerestrict{\pos}}[\rankrand_\pos]$. Recall from~\eqref{eq:incentive_def_normalized_rank} that $\rankrand_\pos$ is a normalized version of $\invpermtrue(\pos)$. Hence, we equivalently consider \begin{align}
    \Expect_{\invpermtrue(\pos) \given \invpermtruerestrict{\pos}}[\invpermtrue(\pos)].\label{eq:incentive_expect_minimizer}
\end{align}
Recall that the true ranking $\invpermtrue$ is sampled uniformly at random from $\setperms_\numitems$. In words, to compute~\eqref{eq:incentive_expect_minimizer} we sample $\pos$ numbers uniformly at random without replacement from $[\numitems]$. Conditional on the ranking $\invpermtruerestrict{\pos}$ restricted to the $\pos$ items, the $\pos^\textth$ number sampled has a relative rank of $\invpermtruerestrict{\pos}(\pos)$, i.e., it is the $\invpermtruerestrict{\pos}(\pos)^\textth$ order statistics of these $\pos$ numbers.
Applying Lemma~\ref{lem:expect_dist} from Appendix~\ref{app:proof_lem_expect_dist}, we have
\begin{align*}
    \Expect_{\invpermtrue(\pos) \given \invpermtruerestrict{\pos}} [\invpermtrue(\pos)] = \frac{\numitems+1}{\pos+1}\cdot \invpermtruerestrict{\pos}(\pos),
\end{align*}
and normalizing yields
\begin{align}\label{eq:expect_sample_wo_replacement}
    \Expect_{\rankrand_\pos\given \invpermtruerestrict{\pos}}[\rankrand_\pos] = \frac{\invpermtruerestrict{\pos}(\pos)}{\pos+1}.
\end{align}
Substituting equation~\eqref{eq:expect_sample_wo_replacement} back into equation~\eqref{eq:decouple_from_past}, the minimizer can be written as
\begin{align}\label{eq:incentive_det_opt}
    \estscore_\pos(\invpermtruerestrict{\pos}) = \frac{\invpermtruerestrict{\pos}(\pos)}{\pos+1}.
\end{align}

\noindent \textbf{Step 3: Show that randomization does not improve any deterministic estimator.}
Let $\sourcerand_1, \ldots, \sourcerand_\pos$ denote random variables supported on some set $\setsource$. Each $\sourcerand_\idxitem$ is sampled by the randomized estimator at timestep $\idxitem$ independently from all else. The class of randomized estimators is thus any estimator $\estscore=\{\estscore_\pos\}_{\pos\in [\numitems]}$, where each $\estscore_\pos: \setperms_\pos \times \setsource^{\pos} \rightarrow \reals$ is a deterministic function that takes as input the ranking $\invpermtruerestrict{\pos}$ restricted to the $\pos$ items as well as the values of $\sourcerand_{1:\pos}$, and outputs a real value. The error of any such randomized estimator $\estscore$ at timestep $\pos$ can be written as
\begin{align*}
    \Expect_{\invpermtrue, \sourcerand_{1:\numitems}} \left(\estscore_\pos(\invpermtruerestrict{\pos}, \sourcerand_{1:\pos}) - \rankrand_\pos\right)^2
    & = \Expect_{\invpermtrue, \sourcerand_{1:\pos}} \left(\estscore_\pos(\invpermtruerestrict{\pos}, \sourcerand_{1:\pos}) - \rankrand_\pos\right)^2\\
    & = \Expect_{\invpermtruerestrict{\pos},\sourcerand_{1:\pos}}\; \Expect_{\rankrand_\pos\given \invpermtruerestrict{\pos}, \sourcerand_{1:\pos}} \left(\estscore_\pos(\invpermtruerestrict{\pos}, \sourcerand_{1:\pos}) - \rankrand_\pos\right)^2 \nonumber\\
    & \stackrel{\stepone}{=}\Expect_{\sourcerand_{1:\pos}} \Expect_{\invpermtruerestrict{\pos}}\; \Expect_{\rankrand_\pos\given \invpermtruerestrict{\pos}} \left(\estscore_\pos(\invpermtruerestrict{\pos}, \sourcerand_{1:\pos}) - \rankrand_\pos\right)^2, \nonumber\\
\end{align*}
where step~\stepone is true because the values of $\sourcerand_{1:\pos}$ are sampled independent from all else.
Note that each $\estscore_\pos(\invpermtruerestrict{\pos}, \sourcerand_{1:\pos})$ corresponds to a deterministic estimator for fixed $\sourcerand_{1:\pos}$, so we have reduced to the previous case. The minimizer over the class of randomized estimators is attained by the deterministic estimator~\eqref{eq:incentive_det_opt}.
\qed

\subsection{Proof of Proposition~\ref{prop:worst_case_induced}}\label{sec:proof_prop_worst_case_induced}

Due to relation~\eqref{eq:relation_err_sf_entrywise} between the Spearman's footrule error and the entry-wise error, it suffices to prove bound~\eqref{eq:err_noiseless_induce_sf}. We consider the Kendall--tau distance (see Eq.~\ref{eq:kt}), constructing a permutation and lower bounding the number of flips between it and the ranking induced by the noiseless scores of this permutation.
Furthermore, it suffices to consider any $\numitems$ that is divisible by $4$. To see this, consider any $\numitems\ge 8$ and any ranking $\invperm\in \setperms_\numitems$. In the parametric model~\eqref{eq:model_parametric_param}, it can be verified that the number of flips between $(\invperm, \invpermtrue)$ is at least the number of flips between $(\invpermrestrict{4\floor{\frac{\numitems}{4}}}, \invpermtruerestrict{4\floor{\frac{\numitems}{4}}})$. Accordingly, for the remainder of the proof, we consider $\numitems=4\numitemsfourth$ (with $\numitemsfourth \ge 2$).

\paragraph{Constructing a flip.} We now describe a construction of a flip between some item $\pos$ of relative rank $\rankreltrue_\pos$ and item $(\pos+2)$ of relative rank $\rankreltrue_{\pos+2}$. Assume that the following two conditions are satisfied:
\begin{subequations}\label{eq:worst_case_induced_disagreement_condition}
\begin{align}
    \rankreltrue_{\pos+2} & = \rankreltrue_\pos+1\\
    \rankreltrue_\pos & > \frac{\pos}{2} + \frac{1}{2}.
\end{align}
\end{subequations}
From these two conditions, it can be verified that
\begin{align*}
    \frac{\rankreltrue_\pos}{\pos+1} > \frac{\rankreltrue_\pos+1}{\pos+3} = \frac{\rankreltrue_{\pos + 2}}{\pos + 3}
\end{align*}
and hence by the definition of the parametric model~\eqref{eq:model_parametric_param}, we have
\begin{align*}
   \param(\pos, \rankreltrue_\pos) > \param(\pos+2, \rankreltrue_{\pos+2}).
\end{align*}
Consequently, if item $\pos$ is ranked lower than item $(\pos+2)$ in the true ranking $\invpermtrue$, then in the noiseless setting, the ranking $\estinvperminduced$ induced by the scores has a flip at the pair $(\pos, \pos+2)$.

\paragraph{Constructing a true ranking.}
Using the construction of a flip above, we now construct a true ordering $\invpermtrue$ via
\begin{align}\label{eq:worst_case_induce_construction}
    \invpermtrue(\pos) = \begin{cases}
        \pos & \text{if } \pos \le 2\numitemsfourth\\
        \frac{\pos+1}{2}+\numitemsfourth & \text{if } \pos > 2\numitemsfourth \text{ and } \pos \text{ is odd}\\
        \frac{\pos}{2}  +2\numitemsfourth & \text{if } \pos > 2\numitemsfourth \text{ and } \pos \text{ is even}.
    \end{cases}
\end{align}
In words, the first half of the items in $\invpermtrue$ are the lowest items (of absolute ranks $1$ through $2\numitemsfourth$). In the second half, the items at odd positions are of absolute ranks $(2\numitemsfourth+1)$ through $3\numitemsfourth$; the items at even positions are the highest items (of absolute ranks $(3\numitemsfourth+1)$ through $4\numitemsfourth$).

Now consider any pair of items $(\pos, \pos+2)$, with $\pos> 2\numitemsfourth$ and $\pos$ is odd. By the construction~\eqref{eq:worst_case_induce_construction} of $\invpermtrue$, it can be verified that the relative ranks of the two items are identical to their absolute ranks. That is,
\begin{subequations}\label{eq:worst_case_induced_rel_equals_abs}
\begin{align}
    & \rankreltrue_\pos = \invpermtrue(\pos) \\
    & \rankreltrue_{\pos+2} = \invpermtrue(\pos+2) .
\end{align}
\end{subequations}
Combining~\eqref{eq:worst_case_induced_rel_equals_abs} with the construction~\eqref{eq:worst_case_induce_construction} of the true ranking $\invpermtrue$, it can be verified that both conditions in~\eqref{eq:worst_case_induced_disagreement_condition} are satisfied. Hence, we have
\begin{align}\label{eq:worst_case_induced_single_pair}
    \param(\pos, \rankreltrue_\pos) > \param(\pos+2, \rankreltrue_{\pos+2}) \qquad \text{ for all } \pos > 2\numitemsfourth \text{ such that } \pos \text{ is odd}.
\end{align}
Concatenating Eq.~\eqref{eq:worst_case_induced_single_pair} over all $\pos$ yields
\begin{align*}
    \param(2\numitemsfourth+1, \rankreltrue_{2\numitemsfourth+1}) > \param(2\numitemsfourth+3, \rankreltrue_{2\numitemsfourth+3}) > \ldots > \param(4\numitemsfourth-1, \rankreltrue_{4\numitemsfourth-1}).
\end{align*}
On the other hand, the true ranking $\invpermtrue$ constructed in~\eqref{eq:worst_case_induce_construction} follows
\begin{align*}
    \invpermtrue(2\numitemsfourth+1) < \invpermtrue(2\numitemsfourth+3) < \ldots < \invpermtrue(4\numitemsfourth-1).
\end{align*}
Consequently, there is a flip between any pair $(\pos, \posalt)$ with $\pos\ne \posalt > 2\numitemsfourth$, and $(\pos,\posalt)$ are both odd. The total number of flips is thus lower bounded by
\begin{align*}
    \binom{\numitemsfourth}{2} \ge \frac{ \numitemsfourth^2}{4} \ge \frac{\numitems^2}{64}.
\end{align*}
Normalizing by $\numitems^2$ completes the proof.
\qed

\subsection{Proof of Proposition~\ref{prop:equivalence}}\label{sec:proof_prop_equivalence}

\paragraph{Correctness.}
We first show that Algorithm~\ref{algo:ls} computes the least squares estimator~\eqref{eq:ls} exactly.
The objective function of the least squares estimator~\eqref{eq:ls_ub_adaptive} can be decomposed as
\begin{align}
    \norm{\report - \param(\invperm)}_2^2 = \sum_{\pos=1}^\numitems \big[\report_\pos - \param(\pos, \rankrel_\pos(\invperm))\big]^2\label{eq:equivalence_obj}
\end{align}
Let $\estinvperm$ denote the ranking constructed by Algorithm~\ref{algo:ls}. According to the insertion procedure (Line~\ref{line:insert}), the relative rank of each item $\pos$ is $\estrankrel_\pos$ (among the $\pos$ items) in the ranking $\estinvperm$. By Line~\ref{line:compute_rank} in Algorithm~\ref{algo:ls}, we have that $\estrankrel_\pos$ minimizes
\begin{align*}
\min_{\rankrel\in [\pos]} \big(\report_\pos - \param(\pos, \rankrel)\big)^2.
\end{align*}
Summing over $\pos\in [\numitems]$, yields that $\estrankrel_{1:\numitems}\in [1] \times \cdots \times [n]$ minimizes~\eqref{eq:equivalence_obj}; consequently, $\estinvperm$ minimizes~\eqref{eq:equivalence_obj}.

\paragraph{Time complexity.}
We describe an implementation of Algorithm~\ref{algo:ls} using an order-statistics tree~\citep[Chapter 14.1]{algorithms2009cormen}. An order-statistics tree is a red-black tree~\citep{bayer1972tree}, where each insertion takes worst-case $O(\log \numitems)$ time~\cite[Chapter 13.3]{algorithms2009cormen}, with the augmented functionality that retrieving an element with a given rank takes worst-case $O(\log \numitems)$ time.

In  Algorithm~\ref{algo:ls}, we store the array $a$ in an order-statistics tree. For each timestep $\pos\in [\numitems]$, the insertion position $\estrankrel_\pos$ (Line~\ref{line:compute_rank}) has a closed-form solution in $\Theta(1)$ time, which is the closest integer within the range $[1, \pos]$ to the value $\report_\pos (\pos+1)$. Let us now describe how to insert item $t$ into rank $\estrankrel_\pos$ in array $\arr$ (Line~\ref{line:insert}). First, we assign an arbitrary real value to item $1$. Then for each item $2\le \pos \le \numitems$, we retrieve the two values of the items of rank $(\estrankrel_{\pos} -1)$ and $\estrankrel_\pos$ in the array $\arr$ respectively, and assign to item $\pos$ an arbitrary real value between these two values. In the boundary case where item $\pos$ is inserted to the beginning of the sequence with $\estrankrel_\pos-1 = 0$, we set the value of the item of rank $(\estrankrel_\pos-1)$ (in the tree of $(\pos-1)$ items) as $-\infty$. Likewise if item $\pos$ is inserted to the end of the sequence with $\estrankrel_\pos = \pos$, we set the value of the item of rank $\estrankrel_\pos$  (in the tree of $(\pos-1)$ items) as $\infty$.
Due to the structure of the order-statistics tree, each retrieval by rank takes $O(\log \numitems)$ time, and each insertion takes $O(\log\numitems)$ time.

Summing over $\pos\in [\numitems]$, the for-loop (Line~\ref{line:for_start}-\ref{line:for_end}) takes $O(\numitems\log \numitems)$ time. Finally, the absolute ranks of all the items (Line~\ref{line:sort}) can be obtained by a traversal of the tree which takes linear time.
\qed

\subsection{Proof of Theorem~\ref{thm:mle_ub}}\label{sec:proof_thm_mle}

For notational simplicity, we drop the subscript and write $\estinvperm\defn \estinvpermls$ for the least squares estimator.
Recall the definition of $\objtrue_{[\pos]}$ and $\objest_{[\pos]}$ from~\eqref{eq:ub_construct_intermediate} as the vectors consisting of the absolute ranks of the first $\pos$ items, sorted in increasing order according to $\invpermtruerestrict{\pos}$ and $\estinvpermrestrict{\pos}$, respectively. Using relation~\eqref{eq:obj_n}, we have
\begin{align*}
    \numitems^2 \losssf(\estinvperm, \invpermtrue) = \normone{\estinvperm - \invpermtrue} = \normone{\identityrank_\numitems - \invpermtrue\circ (\estinvperm)^{-1}} = \normone{\objtrue_{[\numitems]} - \objest_{[\numitems]}},
\end{align*}
so that it suffices to control the $\ell_1$ norm between the vectors $\objtrue_{[\numitems]}$ and $\objest_{[\numitems]}$. Our proof proceeds by constructing a recursive relation for the quantity $\normone{\objtrue_{[\numitems]}-\objest_{[\numitems]}}$.

Toward this goal, we define one more intermediate object  $\objpartial_{[\pos]}$ as the ``partially-corrected'' version of $\objest_{[\pos]}$, where the position of the last item (item $\pos$) is fixed, and in the remaining positions, all the other items are sorted in increasing order according to their true ranks. In the example in Eq.~\eqref{eq:ub_mle_example}, the last item (item $3$) has a true rank of $60$, so $\objpartial_{[3]}$ is obtained from $\objest_{[3]} = [60, 40, 10]$ by leaving $60$ where it is and correctly reordering $10$ and $40$. 
That is, $\objest_{[3]} = [60, 10, 40]$.

By triangle inequality, we have
\begin{align*}
    \normone{\objtrue_{[\pos]} - \objest_{[\pos]} }
    \le \normone{ \objtrue_{[\pos]} - \objpartial_{[\pos]}} + \normone{ \objpartial_{[\pos]} - \objest_{[\pos]} }.
\end{align*}
Recall that $\objtrue_{[\pos]}$ sorts the absolute ranks of the $\pos$ items in increasing order. By definitions of $\objest_{[\pos]}$ and $\objpartial_{[\pos]}$, we have $\normone{ \objpartial_{[\pos]} - \objest_{[\pos]} } = \normone{  \objtrue_{[\pos-1]} - \objest_{[\pos-1]} }$, so that
\begin{align*}
    \normone{\objtrue_{[\numitems]}- \objest_{[\numitems]} }
    & \le  \normone{ \objtrue_{[\numitems]} - \objpartial_{[\numitems]} } + \normone{\objtrue_{[\numitems-1]} - \objest_{[\numitems-1]}} \\
    & \le \ldots \\
    & \le \sum_{\pos=2}^\numitems \normone{ \objtrue_{[\pos]} - \objpartial_{[\pos]} } + \normone{ \objtrue_{[1]} - \objest_{[1]} } 
    = \sum_{\idxitem=2}^\numitems \normone{ \objtrue_{[\pos]} - \objpartial_{[\pos]}}.
\end{align*}
We have thus reduced our problem to one of analyzing $\normone{  \objtrue_{[\pos]} - \objpartial_{[\pos]} }$ for each $\pos$.

Recall that at time $\pos$, the last item (item $\pos$) has a true relative rank of $\rankreltrue_\pos$ (out of the $\pos$ items). By Algorithm~\ref{algo:ls}, the insertion position $\estrankrel_\pos$ of item $\pos$ is given by rounding $\report_\pos \cdot (\pos + 1)$ to the nearest integer within $\{1, \ldots, \pos\}$. Since the noise is bounded in $[-\noiseunifparam, \noiseunifparam]$,  the inserted position is bounded as
\begin{align}
    \estrankrel_\pos\in [\rankreltrue_\pos -\ceilnoiseposplueone,  \rankreltrue_\pos + \ceilnoiseposplueone], \label{eq:ub_insertion_pos_bound}
\end{align}
where we recall the shorthand $\posplusone = \pos + 1$. We now split the argument into two cases.
    
\begin{figure}[tb]
    \centering
    \includegraphics[width=0.65\linewidth]{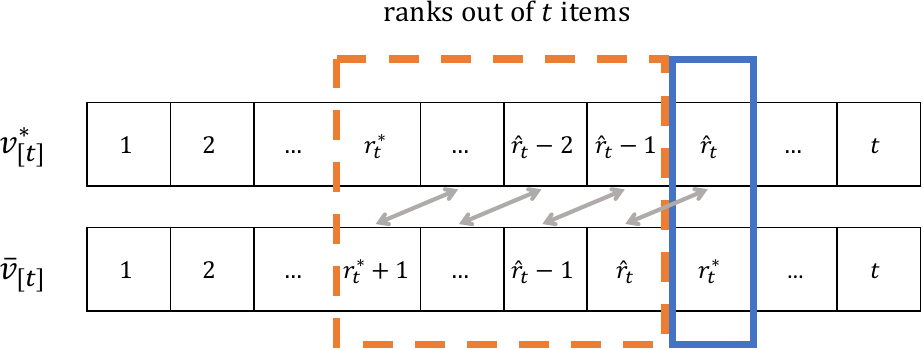}
    \caption{The relative ranks of items in $\objtrue_{[\pos]}$ and $\objpartial_{[\pos]}$.}
    \label{fig:insertion}
\end{figure}

\paragraph{Case $\estrankrel_\pos \ge \rankreltrue_\pos$.} Consider the items in positions from $\rankreltrue_\pos$ to $\estrankrel_\pos$, in the vectors $\objpartial_{[\pos]}$ and $\objtrue_{[\pos]}$. 
Since $\objtrue_{[\pos]}$ is sorted in increasing order, the items in these positions have true ranks (out of $\pos$ items) in $[\rankreltrue_\pos, \estrankrel_\pos]$. On the other hand, the partially-corrected ordering $\objpartial_{[\pos]}$ is similar to $\objtrue_{[\pos]}$, except that: (a) The current item (having true relative rank $\rankreltrue_\pos$) may not be at position $\rankreltrue_\pos$ and is instead at position $\estrankrel_\pos$, and (b)~the items having true ranks (out of $\pos$ items) between $\rankreltrue_\pos+1$ and $\estrankrel_\pos$ shift by one position. See Figure~\ref{fig:insertion} for an illustration of the ranks of the items in $\objtrue_{[\pos]}$ and $\objpartial_{[\pos]}$.
    
Recall from Section~\ref{sec:theory_noisy} our definition of the quantity $\mapreltoabs^\pos_\rankrel(\invperm)\defn \invperm\big(\invpermrestrict{\pos}^{-1}(\rankrel)\big)$, namely, the true absolute rank of the item that is the $\rankrel^\textth$ largest among the first $\pos$ items according to any ranking $\invperm$.  Note that for each fixed position $\pos$ and ranking $\invperm$, we have that $\mapreltoabs^\pos_{\rankrel}(\invperm)$ is monotonically increasing when viewed as a function of $\rankrel$, since an item with a higher relative rank has a higher absolute rank. 
     
As shown in Figure~\ref{fig:insertion}, the items in $\objtrue_{[\pos]}$ and $\objpartial_{[\pos]}$ only differ in positions $\rankreltrue_\pos$ through $\estrankrel_\pos$. We decompose the $\ell_1$ error $\normone{  \objtrue_{[\pos]} - \objpartial_{[\pos]} }$ into the following two parts:
\begin{itemize}
    \item First, we consider the position $\estrankrel_\pos$ (see the solid blue box in Figure~\ref{fig:insertion}). At position $\estrankrel_\pos$, the item in $\objtrue_{[\pos]}$ has a true rank $\estrankrel_\pos$ (out of $\pos$ items), and the item in $\objpartial_{[\pos]}$ has a true rank of $\rankreltrue_\pos$ (out of $\pos$ items). By the definition of $\mapreltoabs^\pos_{r}$, the error at this position is equal to
    \begin{align}
        \abs*{\mapreltoabs^\pos_{\estrankrel_\pos}(\invpermtrue) - \mapreltoabs^\pos_{\rankreltrue_\pos}(\invpermtrue)}
        & =  \mapreltoabs^\pos_{\estrankrel_\pos}(\invpermtrue) - \mapreltoabs^\pos_{\rankreltrue_\pos}(\invpermtrue), \label{eq:ub_err_part_one}
    \end{align}
    where we have used the monotonicity relation alluded to before.
    
    \item Next, we consider the positions $[\rankreltrue_\pos, \estrankrel_\pos-1]$ (see the dashed orange box in Figure~\ref{fig:insertion}). Each of the items is shifted by one position. Using the monotonicity relation, the $\ell_1$ error contributed by these positions is given by
    \begin{align}
        \sum_{\rankrel = \rankreltrue_\pos}^{\estrankrel_\pos-1} \abs*{\mapreltoabs^\pos_{\rankrel+1}(\invpermtrue) - \mapreltoabs^\pos_{\rankrel}(\invpermtrue)} {=}
        \sum_{\rankrel = \rankreltrue_\pos}^{\estrankrel_\pos-1} \left(\mapreltoabs^\pos_{\rankrel+1}(\invpermtrue) - \mapreltoabs^\pos_{\rankrel}(\invpermtrue)\right) 
        =  \mapreltoabs^\pos_{\estrankrel_\pos}(\invpermtrue)- \mapreltoabs^\pos_{\rankreltrue_\pos}(\invpermtrue) , \label{eq:ub_err_part_two}
    \end{align}
\end{itemize}
    
Putting~\eqref{eq:ub_err_part_one} and~\eqref{eq:ub_err_part_two} together, we see for each $\pos \in [\numitems]$, the $\ell_1$ error is upper bounded as 
\begin{subequations}\label{eq:ub_case_insert}
\begin{align}
    \normone{ \objtrue_{[\pos]} - \objest_{[\pos]} } \leq 
    2\left(\mapreltoabs^\pos_{\estrankrel_\pos}(\invpermtrue) - \mapreltoabs^\pos_{\rankreltrue}(\invpermtrue)\right)
    \stackrel{\stepone}{\le} 2\left(\mapreltoabs^\pos_{\rankreltrue_\pos + \ceil*{\noiseunifparam\posplusone}}(\invpermtrue) - \mapreltoabs^\pos_{\rankreltrue}(\invpermtrue) \right),
\end{align}
where step~\stepone is true due the monotonicity of $\mapreltoabs^\pos_\rankrel$ and the bound~\eqref{eq:ub_insertion_pos_bound} on $\estrankrel_\pos$ (recall the definition that $\mapreltoabs_\rankrel^\pos = \mapreltoabs_\pos^\pos$ for $\rankrel > \pos$). 
    
\paragraph{Case $\rankrelalt_\pos < \rankreltrue$.} By a symmetric argument to the one above, we have
\begin{align}
    \normone{ \objtrue_{[\pos]} - \objpartial_{[\pos]} } 
    \le & 2\left( \mapreltoabs^\pos_{\rankreltrue}(\invpermtrue) - \mapreltoabs^\pos_{\rankreltrue_\pos - \ceil*{\noiseunifparam\posplusone}}(\invpermtrue)\right),
\end{align}
\end{subequations}
with the convention that $\mapreltoabs_\rankrel^\pos = \mapreltoabs_1^\pos$ for $\rankrel < 1$.

\medskip
    
Finally, combining the two cases, i.e., the two parts of Eq.~\eqref{eq:ub_case_insert}, we have
\begin{align*}
     \normone{ \objtrue_{[\pos]} - \objpartial_{[\pos]} } 
     & \le 2 \left( \mapreltoabs^\pos_{\rankreltrue_\pos + \ceil*{\noiseunifparam \posplusone}}(\invpermtrue) - \mapreltoabs^\pos_{\rankreltrue_\pos - \ceil*{\noiseunifparam \posplusone}}(\invpermtrue) \right).
\end{align*}
Summing over $\pos\in [\numitems]$ and normalizing completes the proof. 
\qed

\subsection{Proof of Corollary~\ref{cor:ls_ub_uniform_prior}}\label{sec:cor-proof}

In the case of a uniform true ordering $\invpermtrue$, we apply Lemma~\ref{lem:expect_dist} from Appendix~\ref{app:proof_lem_expect_dist} to bound each error term $\mapreltoabs^\pos_{\rankreltrue_\pos + \ceil*{\noiseunifparam \posplusone}}(\invpermtrue) - \mapreltoabs^\pos_{\rankreltrue_\pos - \ceil*{\noiseunifparam \posplusone}}(\invpermtrue)$. Note that this term is $0$ when $\noiseunifparam\posplusone < 0.5$. Hence, it suffices to consider any fixed timestep $\pos$ such that $\noiseunifparam\posplusone = \noiseunifparam (\pos + 1) \ge 0.5$. 
Invoking Lemma~\ref{lem:expect_dist_perm} from Appendix~\ref{app:proof_lem_expect_dist} yields
\begin{align}
    \Expect_{\invpermtrue\sample \setperms_\numitems}\left[\mapreltoabs^\pos_{\rankreltrue_\pos + \ceilnoiseposplueone}(\invpermtrue) - \mapreltoabs^\pos_{\rankreltrue_\pos - \ceilnoiseposplueone}(\invpermtrue) \right]
    & = \Expect_{\noise_{1:\pos}}\Expect_{\rankreltrue_{1:\pos}}\left[2\cdot \ceilnoiseposplueone \cdot \frac{\numitems+1}{\pos+1} \right] \nonumber\\
    & {=} \frac{\numitems+1}{\posplusone} \cdot 2\ceilnoiseposplueone \stackrel{\stepone}{\lessorder} \numitems \noiseunifparam,\label{eq:ub_expect_diff_order_stats}
\end{align}
where step~\stepone uses the assumption that $\noiseunifparam\posplusone \ge 0.5$.
Summing~\eqref{eq:ub_expect_diff_order_stats} over $\pos\in [\numitems]$ completes the proof.
\qed

\subsection{Proof of Theorem~\ref{thm:lb_sf}}\label{sec:proof_thm_lb_sf}

Recall from~\eqref{eq:dist_inv_vec} in Section~\ref{sec:proof_notation} that the inversion vector $\rankrelvec(\invperm) \defn \{\rankrel_\pos(\invperm)\}_{\pos\in [\numitems]}$ is defined as the vector of relative ranks, and recall from~\eqref{eq:dist_inv_vec} that $\lossinvvec(\invperm, \invpermtrue)$ is defined as the (normalized) $\ell_1$ distance between the inversion vectors of two rankings. 

The standard result of Lemma~\ref{lem:expect_dist_perm} from Appendix~\ref{app:proof_lem_expect_dist} establishes a relation between the Kendall--tau error and the $\ell_1$ error on inversion vectors. Namely, for any two rankings $\invperm_1$ and $\invperm_2$, we have the inequality:
\begin{align}
    \lossinvvec(\invperm_1, \invperm_2) \le \losskt(\invperm_1, \invperm_2).\label{eq:relation_inv_vec_kt}
\end{align}
Recall that the Spearman's footrule error and the Kendall--tau error are within a constant factor from each other~\citep{diaconis1977disarray}. Relation~\eqref{eq:relation_inv_vec_kt} thus allows us to convert the desired Spearman's footrule error, an error dependent across each step $\pos\in [\numitems]$, to the $\ell_1$ error on inversion vectors, an error independent across each step. It remains to prove that
\begin{align}\label{eq:lb_inv_vec_desired}
    \inf_{\estinvperm}\Expect[\lossinvvec(\estinvperm, \invpermtrue)] \ge \const \noiseunifparam.
\end{align}

Consider any estimator $\estinvperm:\reals^\numitems\rightarrow \setperms_\numitems$. Recall the shorthand $\estrankrel_\pos \defn \rankrel_\pos(\estinvperm)$ and $\posplusone\defn \pos +1$. Slightly abusing the notation by using $\Prob$ to also denote the p.d.f. of continuous random variables, we write the expected error on the relative rank at any position $\pos$ as
\begin{align}
    \Expect \abs*{\estrankrel_\pos - \rankreltrue_\pos} & \stackrel{\stepone}{=} \int_{\report_{1:\numitems}}\sum_{\rankrel\in [\pos]} \abs*{\estrankrel_\pos(\report_{1:\numitems}) - \rankrel}\cdot \Prob(\report_{1:\numitems}, \rankreltrue_\pos = \rankrel)\dd \report_{1:\numitems},\nonumber\\
  & \stackrel{\steptwo}{=} \int_{\report_{-\pos}}
    \underbrace{
        \left[\int_{\frac{1}{\posplusone}-\noiseunifparam}^{\frac{\pos}{\posplusone} + \noiseunifparam}\sum_{\rankrel\in [\pos]}
        \abs*{\estrankrel_\pos(\report_{1:\numitems}) - \rankrel}\cdot \Prob(\report_\pos, \rankreltrue_\pos=\rankrel) \dd \report_\pos\right]
    }_{\term_\pos(\report_{-\pos})}
    \Prob(\report_{-\pos})\dd\report_{-\pos},\label{eq:lb_inv_vec_term}
\end{align}
where step~\stepone is true because it can be verified that intenral randomness cannot improve the estimator, so we only consider any deterministic estimator $\estinvperm$; step~\steptwo is true because the true relative ranks $\{\rankreltrue_\pos\}_{\pos\in [\numitems]}$ are independent under a uniform prior of $\invpermtrue$, and the noise terms $\{\noise_\pos\}_{\pos\in [\numitems]}$ are independent, so $\report_\pos$ is independent from $\report_{-\pos}\defn \report_{[\numitems]\setminus \{\pos\}}$ according to the model~\eqref{eq:model_parametric}. We now fix any value for $\report_{-\pos}\in \reals^{\numitems-1}$ and consider the term $\term_\pos(\report_{-\pos})$ defined in~\eqref{eq:lb_inv_vec_term}. We have
\begin{align}
    \Prob(\report_\pos, \rankreltrue_\pos = \rankrel) & = \Prob(\report_\pos \given \rankreltrue_\pos =\rankrel)\cdot \Prob(\rankreltrue_\pos = \rankrel)\nonumber\\
    & = \Prob(\report_\pos \given \rankreltrue_\pos = \rankrel) \cdot \frac{1}{\pos}.\label{eq:lb_inv_vec_prob_conditional}
\end{align}
Recall that the noise term $\noise_\pos$ is sampled from $\uniform[-\noiseunifparam, \noiseunifparam]$. For any $\report_\pos\in [\frac{1}{\posplusone}+\noiseunifparam, \frac{\pos}{\posplusone} - \noiseunifparam]$, we have
\begin{align}
     \Prob(\report_\pos \given \rankreltrue_\pos =\rankrel) & =\Prob\left(\noise_\pos = \report_\pos - \frac{\rankrel}{\posplusone}\;\middle|\; \rankreltrue_\pos = \rankrel\right) =\begin{cases}
        \frac{1}{2\noiseunifparam} & \text{if } \ceil*{(\report_\pos - \noiseunifparam)\posplusone}\le \rankrel \le \floor*{(\report_\pos+\noiseunifparam)\posplusone} \\
        0 & \text{otherwise.}
    \end{cases}\label{eq:lb_inv_vec_prob_expression}
\end{align}
Plugging~\eqref{eq:lb_inv_vec_prob_conditional} and~\eqref{eq:lb_inv_vec_prob_expression} back to the expression of $\term_\pos(\report_{-\pos})$ in~\eqref{eq:lb_inv_vec_term}, we have that for any estimator $\estrankrel_\pos$,
\begin{align*}
    \term_\pos(\report_{-\pos}) & \ge \int_{\frac{1}{\posplusone}+\noiseunifparam}^{\frac{\pos}{\posplusone} - \noiseunifparam} \sum_{\rankrel=\ceil*{(\report_\pos - \noiseunifparam)\posplusone}}^{\floor*{(\report_\pos + \noiseunifparam)\posplusone}} \abs*{\estrankrel_\pos(\report_{1:\numitems}) - \rankrel}\cdot \frac{1}{2\noiseunifparam\pos} \dd \report_\pos\\
    & \ge \int_{\frac{1}{\posplusone}+\noiseunifparam}^{\frac{\pos}{\posplusone} - \noiseunifparam} \frac{(\floor*{(\report_\pos + \noiseunifparam)\posplusone} - \ceil*{(\report_\pos - \noiseunifparam)\posplusone} )^2}{4} \cdot\frac{1}{2\noiseunifparam\pos}\dd \report_\pos
\end{align*}
For any $\pos \ge \frac{2}{\noiseunifparam}$, we have 
\begin{align*}
    \floor*{(\report_\pos + \noiseunifparam)\posplusone} - \ceil*{(\report_\pos - \noiseunifparam)\posplusone} & \ge (\report_\pos + \noiseunifparam)\posplusone-1 - (\report_\pos - \noiseunifparam)\posplusone-1\\
    & = 2\noiseunifparam\posplusone-2 \gtrorder\noiseunifparam\pos,
\end{align*}
and hence for any $\noiseunifparam < \frac{1}{4}$ and any $\pos \ge \frac{2}{\noiseunifparam} > 8$, we have
\begin{align}
    \term_\pos(\report_{-\pos}) & \gtrorder \left(\frac{\pos-1}{\posplusone}-2\noiseunifparam\right) \cdot\noiseunifparam\pos \gtrorder \noiseunifparam\pos,\label{eq:lb_inv_vec_term_bound}
\end{align}
Plugging~\eqref{eq:lb_inv_vec_term_bound} back to~\eqref{eq:lb_inv_vec_term}, we have that for any $\noiseunifparam < \frac{1}{4}$ and any $\pos \ge \frac{2}{\noiseunifparam}$,
\begin{align}
    \Expect \abs*{\estrankrel_\pos - \rankreltrue_\pos} & \gtrorder \noiseunifparam\pos.\label{eq:lb_inv_vec_t_bound}
\end{align}
Recall the assumption that $\numitems\ge \frac{\Const}{\noiseunifparam}$ for some positive constant $\Const$.
Setting $\Const=4$, we have $\frac{\numitems}{2} \ge \frac{2}{\noiseunifparam}$. Summing~\eqref{eq:lb_inv_vec_t_bound} over $\pos$, we have that for any $\noiseunifparam<\frac{1}{4}$,
\begin{align*}
    \Expect\normone{\rankrelvec(\estinvperm)-\rankrelvec(\invpermtrue)} \ge \sum_{\pos=\frac{\numitems}{2}}^\numitems \Expect\abs*{\estrankrel_\pos - \rankreltrue_\pos} \gtrorder \sum_{\pos=\frac{\numitems}{2}}^\numitems \noiseunifparam\pos\gtrorder \numitems^2\noiseunifparam,
\end{align*}
Normalizing by $\numitems^2$ completes the proof of~\eqref{eq:lb_inv_vec_desired}.

\qed
  

\subsection{Proof of Theorem~\ref{thm:ub_inf}}\label{sec:proof_thm_ub_inf}

For notational simplicity, we drop the subscript and write $\estinvperm\defn \estinvpermls$ for the least squares estimator. 
The goal is to bound
\begin{align*}
    \abs*{\estinvperm(\pos)- \invpermtrue(\pos)} \qquad \text{ for all } \pos\in [\numitems],
\end{align*}
and in particular, the maximum expectation of this quantity over $\pos$. Since $\estinvperm$ is a permutation, it is equivalent to bound, for all $\pos\in [\numitems]$, the quantity
\begin{align}
    \abs*{\estinvperm \circ\estinvperm^{-1}(\pos)- \invpermtrue \circ\estinvperm^{-1}(\pos)}=\abs*{\pos-\invpermtrue\circ\estinvperm^{-1}(\pos)}.\label{eq:ub_inf_expression}
\end{align}
Recall that $\estinvperm_{[\pos]}$ denotes the ranking restricted to the first $\pos$ items. Recall the definitions of $\objtrue_{[\pos]}$ and $\objest_{[\pos]}$ from Eq.~\eqref{eq:ub_construct_intermediate} that $\objtrue_{[\pos]}$ denotes the absolute ranks of the $\pos$ items in increasing order, and $\objest_{[\pos]}$ denotes the absolute ranks of the $\pos$ items sorted in the order of $\estinvperm_{[\pos]}$. Using relation~\eqref{eq:obj_n}, expression~\eqref{eq:ub_inf_expression} reduces to $\abs*{\objest_{[\numitems]} (\pos)- \objtrue_{[\numitems]}(\pos)}$.
Finally, since by definition $\estinvperm_{[\numitems]}$ is a permutation of $\numitems$ items, it thus suffices to bound, for all $t \in [n]$, the absolute value of
\begin{align}
    \vardiff{\numitems}{\pos} := \objest_{[\numitems]} \circ \estinvpermrestrict{\numitems}(\pos)- \objtrue_{[\numitems]}\circ \estinvpermrestrict{\numitems}(\pos). \label{eq:ub_inf_expression_alt}
\end{align}
For each fixed $\idxitem$ such that $\pos \le \idxitem \le \numitems$, let
\begin{subequations}
\begin{align}
    \vardiff{\idxitem}{\pos} &\defn \objest_{[\idxitem]}\circ \estinvpermrestrict{\idxitem}(\pos) - \objtrue_{[\idxitem]}\circ \estinvpermrestrict{\idxitem}(\pos)\label{eq:ub_inf_def_var_diff}\\
    & \stackrel{\stepone}{=} \invpermtrue(\pos) - \objtrue_{[\idxitem]}\circ \estinvpermrestrict{\idxitem}(\pos),\label{eq:ub_inf_def_var_diff_reduce}
\end{align}
\end{subequations}
where step~\stepone holds by definition~\eqref{eq:ub_construct_intermediate_est} of $\objest_{[\idxitem]}$. 
We expand~\eqref{eq:ub_inf_expression_alt} as
\begin{align}
    \abs*{\vardiff{\numitems}{\pos}} & = \abs*{\vardiff{\pos}{\pos} + \sum_{\idxitem=\pos+1}^\numitems \left(\vardiff{\idxitem}{\pos} - \vardiff{\idxitem-1}{\pos}\right)}\nonumber\\
    & \le \abs*{\vardiff{\pos}{\pos}} + \sum_{\idxitem=\pos+1}^\numitems \abs*{\vardiff{\idxitem}{\pos} - \vardiff{\idxitem-1}{\pos}}.\label{eq:ub_inf_sum}
\end{align}
Operationally, we bound $\abs*{\vardiff{\numitems}{\pos}}$ by considering its starting value $\vardiff{\pos}{\pos}$ when item $\pos$  was inserted, and track its change at every timestep then onward. We analyze the two terms in inequality~\eqref{eq:ub_inf_sum} separately. 

\paragraph{Term $\vardiff{\pos}{\pos}$.} Recall once again our shorthand $\posplusone = \pos+ 1$. If $\noiseunifparam\posplusone < 0.5$, then it can be verified that all insertions up to the current time in Algorithm~\ref{algo:ls} are correct, yielding $\objest_{[\idxitem]}=\objtrue_{[\idxitem]}$ and hence $\vardiff{\pos}{\pos} = 0$ by~\eqref{eq:ub_inf_def_var_diff}. It remains to consider $\noiseunifparam\posplusone \ge 0.5$. Recall that the notation $\mapreltoabs^\pos_\rankrel(\invperm)$ denotes the absolute rank of the $\rankrel^\textth$ largest item among the first $\pos$ items according to ordering $\invperm$. We consider the two terms in~\eqref{eq:ub_inf_def_var_diff_reduce}:
\begin{itemize}
    \item \textbf{Term $\invpermtrue(\pos)$.}
    Recall that $\rankreltrue_\pos$ is the true relative rank of item $\pos$, so the true absolute rank of item $\pos$, namely $\invpermtrue(\pos)$, is the absolute rank of the $(\rankreltrue_\pos)^\textth$ largest item (among the $\pos$ items). We have
    \begin{subequations}\label{eq:ub_inf_term_one_order_stats}
    \begin{align}
        \invpermtrue(\pos) = \mapreltoabs^\pos_{\rankreltrue_\pos}(\invpermtrue).
    \end{align}

    \item \textbf{Term $\objtrue_{[\pos]}\circ \estinvpermrestrict{\pos}(\pos)$.} Recall that $\estrankrel_\pos$ is the estimated relative rank of item $\pos$, so we have $\estinvpermrestrict{\pos}(\pos) = \estrankrel_\pos$, and $\objtrue_{[\idxitem]}\circ \estinvpermrestrict{\idxitem}(\pos) = \objtrue_{[\idxitem]}(\estrankrel_\pos)$ is the absolute rank of the $\estrankrel_\pos^\textth$ largest item (among the $\pos$ items). We have \begin{align}
        \objtrue_{[\idxitem]}\circ \estinvpermrestrict{\idxitem}(\pos) = \mapreltoabs^\pos_{\estrankrel_\pos}(\invpermtrue).
    \end{align}
    \end{subequations}
\end{itemize}
Since the noise term $\noise_\pos$ is bounded pointwise in the range $[-\noiseunifparam, \noiseunifparam]$, the definition of $\estrankrel_\pos$ (see Line~\ref{line:compute_rank} in Algorithm~\ref{algo:ls}) yields the pointwise relation
\begin{align}
    \estrankrel_\pos \in \left[\rankreltrue_\pos - \ceilnoiseposplueone, \rankreltrue_\pos + \ceilnoiseposplueone\right].\label{eq:ub_inf_insert_pointwise}
\end{align}
Combining the two terms in~\eqref{eq:ub_inf_term_one_order_stats}, we have the pointwise relation
\begin{align}
    \abs*{\vardiff{\pos}{\pos}} & = \abs*{\mapreltoabs^\pos_{\rankreltrue_\pos}(\invpermtrue) - \mapreltoabs^\pos_{\estrankrel_\pos}(\invpermtrue)}\nonumber \\
    & \le \mapreltoabs^\pos_{\rankreltrue_\pos + \ceil*{\noiseunifparam\posplusone}}(\invpermtrue) - \mapreltoabs^\pos_{\rankreltrue_\pos - \ceil*{\noiseunifparam\posplusone}}(\invpermtrue).\label{eq:ub_inf_term_t_bound}
\end{align}
Taking expectations on both sides of~\eqref{eq:ub_inf_term_t_bound} over the true permutation $\invpermtrue$ chosen uniformly at random and invoking Lemma~\ref{lem:expect_dist_perm} from Appendix~\ref{app:proof_lem_expect_dist} (with $h\equiv 1$) yields
\begin{align}
    \Expect_{\invpermtrue} \abs*{\vardiff{\pos}{\pos}} 
    & \le \Expect_{\rankreltrue_{1:\pos},\noise_{1:\pos}} \left[\mapreltoabs^\pos_{\rankreltrue_\pos + \ceil*{\noiseunifparam\posplusone}}(\invpermtrue) - \mapreltoabs^\pos_{\rankreltrue_\pos - \ceil*{\noiseunifparam\posplusone}}(\invpermtrue) \;\middle|\; \noise_{1:\pos}, \rankreltrue_{1:\pos}\right] \nonumber\\
    & = \frac{\numitems+1}{\pos+1}\cdot \Expect_{\rankreltrue_{1:\pos},\noise_{1:\pos}}
    \left[2\ceil*{\noiseunifparam\posplusone} \right] \nonumber\\
    & = \frac{\numitems+1}{\pos+1} \cdot 2\ceil*{\noiseunifparam\posplusone} \stackrel{\stepone}{\lessorder} \noiseunifparam\numitems,\label{eq:ub_inf_term_one}
\end{align}
where step~\stepone is true by the assumption that $\noiseunifparam\posplusone \ge 0.5$.

\begin{figure}[tb]
    \centering
    \includegraphics[width=0.5\linewidth]{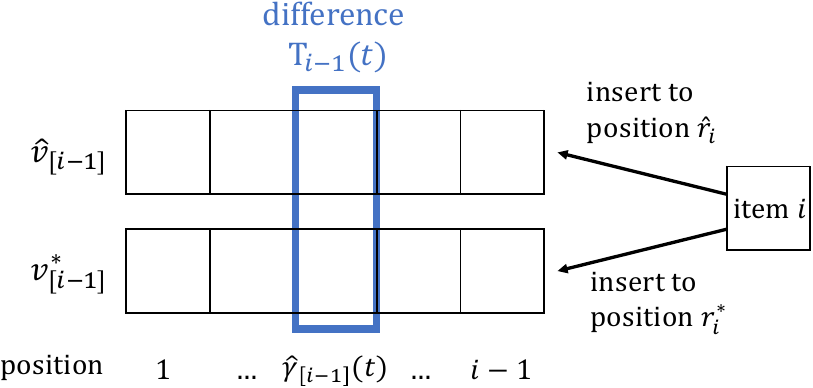}
    \caption{In bounding $\vardiff{\idxitem}{\pos}-\vardiff{\idxitem-1}{\pos}$, we track how inserting item $\idxitem$ changes the vector $\objtrue_{[\idxitem-1]}$ and the position $\estinvpermrestrict{\idxitem-1}(\pos)$.}
    \label{fig:proof_shift}
\end{figure}

\paragraph{Term $\abs*{\vardiff{\idxitem}{\pos} - \vardiff{\idxitem-1}{\pos}}$ for $\idxitem > \pos$.}
Note that if $\noiseunifparam(\idxitem+1) < 0.5$, then all the insertions by our estimator up to time $\idxitem$ are correct, and hence $\abs*{\vardiff{\idxitem}{\pos}-\vardiff{\idxitem-1}{\pos}} = 0$. In what follows, we consider the case $\noiseunifparam(\idxitem+1) \ge 0.5$.
Using the definition~\eqref{eq:ub_inf_def_var_diff_reduce} of $\vardiff{\idxitem}{\pos}$, we have
\begin{align}
    \abs*{\vardiff{\idxitem}{\pos} - \vardiff{\idxitem-1}{\pos}}
    = \abs*{\objtrue_{[\idxitem]}\circ \estinvpermrestrict{\idxitem}(\pos) - \objtrue_{[\idxitem-1]}\circ \estinvpermrestrict{\idxitem-1}(\pos) }.\label{eq:ub_inf_term_diff}
\end{align}
We track how inserting item $\idxitem$ changes the position  $\estinvpermrestrict{\cdot}(\pos)$ and the vector $\objtrue_{[\cdot]}$  in~\eqref{eq:ub_inf_term_diff} from time $(\idxitem-1)$ to time $\idxitem$. As illustrated in Figure~\ref{fig:proof_shift}, we write out the two vectors $\objtrue_{[\idxitem-1]}$ and $\objest_{[\idxitem-1]}$. The blue rectangle is located at position $\estinvpermrestrict{\idxitem-1}(\pos)$. The upper item in the blue rectangle always has a value of $\objest_{[\cdot]}\circ \estinvpermrestrict{\cdot}(\pos)=\invpermtrue(\pos)$; the lower item in the blue rectangle has a value of $\objtrue_{[\cdot]}\circ \estinvpermrestrict{\cdot}(\pos)$. We track how inserting item $\idxitem$ changes the location of the blue rectangle, and the value of the lower item.

By the definition of $\objtrue_{[\cdot]}$ and $\objest_{[\cdot]}$, item $\idxitem$ is inserted to position $\rankreltrue_\idxitem$ in vector $\objtrue_{[\idxitem-1]}$, and to position $\estrankrel_\idxitem$ in vector $\objest_{[\idxitem-1]}$. If item $\idxitem$ is inserted to the left of or at position $\estinvpermrestrict{\idxitem-1}(\pos)$, then the blue rectangle moves to the right by one position (so that the value of its upper item stays the same); if item $\idxitem$ is inserted to the right of  position $\estinvpermrestrict{\idxitem-1}(\pos)$, then the blue rectangle does not move. The term~\eqref{eq:ub_inf_term_diff} is the change in the value of the lower item from timestep $(\idxitem-1)$ to $\idxitem$.
We split the rest of the proof into three cases, depending on where item $\idxitem$ is inserted to the vectors $\objest_{[\idxitem-1]}$ and $\objtrue_{[\idxitem-1]}$. 

\noindent \textbf{Case 1: Item $\idxitem$ is inserted to the left of (or at) $\estinvpermrestrict{\idxitem-1}(\pos)$ in both $\objest_{[\idxitem-1]}$ and $\objtrue_{[\idxitem-1]}$.}

In Figure~\ref{fig:proof_shift}, since item $\idxitem$ is inserted to the left of position $\estinvpermrestrict{\idxitem-1}(\pos)$ in $\objest_{[\idxitem-1]}$, the blue rectangle moves to the right by one position:
\begin{align}
    \estinvpermrestrict{\idxitem}(\pos) = \estinvpermrestrict{\idxitem-1}(\pos) + 1.\label{eq:ub_inf_case_one_change_pos}
\end{align}
Since item $\idxitem$ is inserted to the left of position $\estinvpermrestrict{\idxitem-1}(\pos)$ in $\objtrue_{[\idxitem-1]}$, the item at position $\estinvpermrestrict{\idxitem-1}(\pos)$ in $\objtrue_{[\idxitem-1]}$ also moves to the right by one position:
\begin{align}
    \objtrue_{[\idxitem]}(\estinvpermrestrict{\idxitem-1}(\pos)+1) =\objtrue_{[\idxitem-1]}(\estinvpermrestrict{\idxitem-1}(\pos)).\label{eq:ub_inf_case_one_change_vec}
\end{align}
Combining~\eqref{eq:ub_inf_case_one_change_pos} and~\eqref{eq:ub_inf_case_one_change_vec}, we have
\begin{align*}
    \vardiff{\idxitem}{\pos}- \vardiff{\idxitem-1}{\pos}= 0.
\end{align*}

\noindent \textbf{Case 2: Item $\idxitem$ is inserted to the right of position $\estinvpermrestrict{\idxitem-1}(\pos)$ in both $\objest_{[\idxitem-1]}$ and $\objtrue_{[\idxitem-1]}$.}
The blue rectangle does not move. We again have 
\begin{align*}
    \vardiff{\idxitem}{\pos}-\vardiff{\idxitem-1}{\pos} = 0.
\end{align*}

\noindent \textbf{Case 3: Item $\idxitem$ is inserted to the left of (or at) position $\estinvpermrestrict{\idxitem-1}(\pos)$ in $\objtrue_{[\idxitem-1]}$, and to the right in $\objest_{[\idxitem-1]}$.}
The location of the blue rectangle does not change:
\begin{align*}
    \estinvpermrestrict{\idxitem}= \estinvpermrestrict{\idxitem-1}.
\end{align*}
The lower value of the rectangle at time $\idxitem$ is $\mapreltoabs^\idxitem_{\estrankrel_{[\idxitem-1]}(\invpermtrue)}$, the lower value at time $(\idxitem-1)$ moves to the right by one position: \begin{align*}
    \mapreltoabs^{\idxitem-1}_{\estrankrel_{[\idxitem-1]}}(\invpermtrue) = \mapreltoabs^{\idxitem}_{\estrankrel_{[\idxitem-1]}+1}(\invpermtrue).
\end{align*}.

\noindent\textbf{Case 4: Item $\idxitem$ is inserted to the right of position $\estinvpermrestrict{\idxitem-1}(\pos)$  in $\objtrue_{[\idxitem-1]}$, and to the left of (or at) $\estinvpermrestrict{\idxitem-1}(\pos)$  in $\objest_{[\idxitem-1]}$.}
The location of the blue rectangle moves to the right by one position:
\begin{align*}
    \estinvpermrestrict{\idxitem}= \estinvpermrestrict{\idxitem-1} +1.
\end{align*}
The lower value of the rectangle at time $\idxitem$ is $\mapreltoabs^\idxitem_{\estrankrel_{[\idxitem]}}(\invpermtrue)=\mapreltoabs^\idxitem_{\estrankrel_{[\idxitem-1]}+1}(\invpermtrue)$, and at time $(\idxitem-1)$ is $\mapreltoabs^\idxitem_{\estrankrel_{[\idxitem-1]}}(\invpermtrue)$.
Combining all the cases, the quantity $\abs*{\vardiff{\idxitem}{\pos}-\vardiff{\idxitem-1}{\pos}}$ is bounded by
\begin{align}
    \abs*{\vardiff{\idxitem}{\pos} -\vardiff{\idxitem-1}{\pos}} \le \mapreltoabs^\idxitem_{\estinvpermrestrict{\idxitem-1} + 1}(\invpermtrue) - \mapreltoabs^\idxitem_{\estinvpermrestrict{\idxitem-1}}(\invpermtrue).\label{eq:ub_inf_diff_bound}
\end{align}
Now we consider the possible values of $\rankreltrue$ for these cases. Recall the pointwise relation~\eqref{eq:ub_inf_insert_pointwise} between $\estrankrel_\pos$ and $\rankreltrue$. 

\noindent\textbf{When $\rankreltrue_\pos < \estinvpermrestrict{\idxitem-1}(\pos) - \ceil*{\noiseunifparam(\idxitem+1)}$:} By~\eqref{eq:ub_inf_insert_pointwise} the inserted position of item $\idxitem$ is bounded pointwise as
\begin{align*}
    \estrankrel_\idxitem \le \rankreltrue_\idxitem + \ceil{\noiseunifparam(\idxitem+1)} < \estinvpermrestrict{\idxitem-1}(\pos),
\end{align*}
suggesting that item $\idxitem$ is inserted to the left of both vectors $\objest_{[\idxitem-1]}$ and $\objtrue_{[\idxitem-1]}$, and hence Case 1 is active.

\noindent\textbf{When $\rankreltrue_\pos > \estinvpermrestrict{\idxitem-1}(\pos) + \ceil*{\noiseunifparam(\idxitem+1)}$:} By~\eqref{eq:ub_inf_insert_pointwise} again, we have
\begin{align*}
    \estrankrel_\idxitem \ge \rankreltrue_\idxitem-\ceil*{\noiseunifparam(\idxitem+1)} > \estinvpermrestrict{\idxitem-1}(\pos),
\end{align*}
suggesting that item $\idxitem$ is inserted to the right of both vectors $\objest_{[\idxitem-1]}$ and $\objtrue_{[\idxitem-1]}$, and hence Case 2 is active.
\medskip

We denote the event
\begin{align*}
    \event \defn \Big\{\rankreltrue_\pos \in \left[ \estinvpermrestrict{\idxitem-1}(\pos) - \ceil*{\noiseunifparam(\idxitem+1)},\estinvpermrestrict{\idxitem-1}(\pos) + \ceil*{\noiseunifparam(\idxitem+1)}\right]\Big\}.
\end{align*}
Note that $\estinvpermrestrict{\idxitem-1}$ is a deterministic function of $(\rankreltrue_{1:\idxitem-1}, \noise_{\idxitem-1})$, so $\indicator\{\event\}$ is a deterministic function of $(\rankreltrue_{1:\idxitem}, \noise_{1:\idxitem})$. Taking an expectation on both sides of Eq.~\eqref{eq:ub_inf_diff_bound} over the true permutation $\invpermtrue$ chosen uniformly at random, and invoking Lemma~\ref{lem:expect_dist_perm} from Appendix~\ref{app:proof_lem_expect_dist} with $h = \indicator\{\event\}$, we have
\begin{align}
    \Expect_{\invpermtrue}  \,\abs*{\vardiff{\idxitem}{\pos}- \vardiff{\idxitem-1}{\pos}}  \stackrel{\stepone}{=} \Expect_{\invpermtrue}  \,\big[\abs*{\vardiff{\idxitem}{\pos}- \vardiff{\idxitem-1}{\pos}}\cdot \indicator\{\event\}\big]  = \frac{\numitems+1}{\idxitem+1} \cdot \Expect_{\rankreltrue_{1:\idxitem}, \noise_{1:\idxitem}}\left[\indicator\{\event\}\right], \label{eq:ub_inf_diff_expression}
\end{align}
where step~\stepone is true since conditional on $\setcomplement{\event}$, Case 1 or Case 2 must be true, and we have $\abs*{\vardiff{\idxitem}{\pos}- \vardiff{\idxitem-1}{\pos}}=0$. 
Note that $\rankreltrue_\idxitem$ is independent from $(\rankreltrue_{1:\idxitem-1}, \noise_{1:\idxitem-1})$. By the law of iterated expectation:
\begin{align}
    \Expect_{\rankreltrue_{1:\idxitem}, \noise_{1:\idxitem}} \, [\indicator\{\event\}] & = \Expect_{\rankreltrue_{1:\idxitem-1}, \noise_{1:\idxitem}}\, \Expect_{\rankreltrue_\idxitem} \left[\indicator\{\event\} \;\middle|\; \rankreltrue_{1:\idxitem-1}, \noise_{1:\idxitem}\right] \nonumber\\
    & \stackrel{\stepone}{\le} \frac{2\ceil*{\noiseunifparam(\idxitem+1)}+1}{\idxitem}\stackrel{\steptwo}{\lessorder} \frac{\noiseunifparam(\idxitem+1)}{\idxitem},\label{eq:ub_inf_diff_prob}
\end{align}
where step~\stepone is true by the definition of the event $\event$ and the fact that $\rankreltrue_\idxitem$ is uniformly at random from $[\idxitem]$; step~\steptwo holds since $\noiseunifparam(\idxitem+1) > 0.5$. Substituting~\eqref{eq:ub_inf_diff_prob} into Eq.~\eqref{eq:ub_inf_diff_expression} yields
\begin{align}
    \Expect_{\invpermtrue}\, \abs*{\vardiff{\idxitem}{\pos}-\vardiff{\idxitem-1}{\pos}} \lessorder  \frac{\numitems+1}{\idxitem+1}\cdot \frac{\noiseunifparam(\idxitem+1)}{\idxitem} %
    = \frac{\noiseunifparam(\numitems+1)}{\idxitem}\lessorder \frac{\noiseunifparam\numitems}{\idxitem}.\label{eq:ub_inf_term_two}
\end{align}
~\\
Substituting the two terms from Eqs.~\eqref{eq:ub_inf_term_one} and~\eqref{eq:ub_inf_term_two} back to~\eqref{eq:ub_inf_sum}, we have
\begin{align*}
    \Expect_{\invpermtrue}\,\abs*{\vardiff{\numitems}{\pos}} \lessorder \noiseunifparam\numitems+ \noiseunifparam\numitems \sum_{\idxitem=\pos+1}^\numitems \frac{1}{\idxitem} 
    \lessorder \noiseunifparam\numitems\log\numitems,
\end{align*}
as desired.
\qed

\subsection{Proof of Proposition~\ref{thm:lb_inf}}\label{sec:proof_thm_lb_inf}

Set $\noiseunifparam_0 = \frac{1}{40}$ and consider any $\noiseunifparam < \noiseunifparam_0$. From the parametric model~\eqref{eq:model_parametric_param}, the noiseless score of item $1$ is always $\frac{1}{2}$, independent of its true rank. Since the noise $\noise_1$ is deterministically bounded in $[-\noiseunifparam, \noiseunifparam]$, the observed score for item $1$ is deterministically bounded in $[\frac{1}{2} - \noiseunifparam, \frac{1}{2} + \noiseunifparam]$. In what follows, we show that there are absolute constants $c_1, c_2 \in (0, 1)$ such that the following is true: (a) $\const_1 \numitems$ of the items have observed scores greater than $\frac{1}{2} + \noiseunifparam$ with high probability, and hence the rank of item $1$ induced by the scores is at most $(1-\const_1)\numitems$; (b) when $\invpermtrue$ is chosen at random, the probability that the true rank of item $1$ being above $(1-\frac{\const_1}{2})\numitems$ is greater than $c_2$, and hence the score-induced ranking $\estinvperminduced$ incurs error proportional to $\numitems$ just on position $1$.

In this argument, we decompose the dependencies between the rank of item $1$ and the scores of the rest of the items, by constructing a new sequence where item $1$ is removed from this sequence, and items $2$ through $\numitems$ appear in the same order as in the original sequence. The noiseless score of each item is re-computed in the new sequence due to the removal of item $1$. For the noise term, we couple the randomness in the two sequences, so that the noise on each item has the same realized value across the two sequences. For each $2\le \pos \le \numitems$, we denote $\rankrelminusonetrue_\pos$ as the relative rank of each item $\pos$ among $(\pos-1)$ items according to $\invpermtrue$, in the new sequence without item $1$. That is, for each $\pos \ge 2$, we define 
\begin{align*}
    \rankrelminusonetrue_\pos \defn \abs*{\left\{\idxitem\in \naturals, 2\le \idxitem \le \pos: \, \invpermtrue(\idxitem) \le \invpermtrue(\pos)\right\}}.
\end{align*}
Comparing with the analogous definition~\eqref{eq:def_rel_rank}, we see that $\rankreltrue_\pos \geq \rankrelminusonetrue_\pos$.
Now define the random variable $\varinsertpos_\pos$ as
\begin{align*}
    \varinsertpos_\pos \defn \indicator\left\{\rankrelminusonetrue_\pos \ge \big(\frac{1}{2}+2\noiseunifparam\big)(\pos+1)\right\}.
\end{align*}
Since the true ordering $\invpermtrue$ is chosen uniformly at random, each item $\pos$ is inserted in this new sequence to any of the $(\pos-1)$ positions uniformly at random and independently from all other insertions. For each $\pos \ge 4$, we have
\begin{align*}
     \Prob(\varinsertpos_\pos = 1) = \Prob\left(\rankrelminusonetrue_\pos \ge \big(\frac{1}{2}+2\noiseunifparam\big)(\pos+1)\right)
    & \ge \frac{1}{t-1}\left[(\pos-1) -\big(\frac{1}{2} + 2\noiseunifparam\big)(\pos+1)\right]\\
    & = 1 - \frac{\pos+1}{\pos-1}\big(\frac{1}{2} + 2\noiseunifparam\big)\\
    & \ge \frac{1}{6} - \frac{10}{3}\noiseunifparam \stackrel{\stepone}{>} \frac{1}{12},
\end{align*}
where step~\stepone is due to the assumption that $\noiseunifparam < \frac{1}{40}$. Note that the random variables $\{\varinsertpos_\pos\}_{\pos=2}^\numitems$ are independent. By Hoeffding's inequality, there exists a universal constant $\const > 0$ such that
\begin{align*}
    \Prob\left(\sum_{\pos=4}^\numitems \varinsertpos_\pos \le \frac{\numitems-3}{24}\right) \le e^{-\const (\numitems-3)}.
\end{align*}
Denote the event $\event\defn \left\{\sum_{\pos=4}^\numitems \varinsertpos_\pos \ge \frac{\numitems}{30}\right\}$. There exists a universal constant $\numitemslb >0$, such that for each $\numitems \ge \numitemslb$, we have
\begin{align}
    \Prob(\event) = \Prob\left(\sum_{\pos=4}^\numitems \varinsertpos_\pos \ge \frac{\numitems}{30}\right) \ge \frac{1}{2}.\label{eq:lb_inf_num_greater_half}
\end{align}
The error on item $1$ is bounded as
\begin{align}
    \Expect\abs*{\estinvperminduced(1) - \invpermtrue(1)} \ge \Expect\left[\abs*{\estinvperminduced(1) - \invpermtrue(1)} \;\middle|\; \event\intersect \left\{\invpermtrue(1) > \frac{59\numitems}{60}\right\}\right] \cdot \Prob\left(\event\intersect \left\{\invpermtrue(1) > \frac{59\numitems}{60}\right\}\right). \label{eq:inf_lb_conditional_expression}
\end{align}
We analyze the two terms in~\eqref{eq:inf_lb_conditional_expression} separately. For the second term in~\eqref{eq:inf_lb_conditional_expression}, the relative ranks of items $2$ through $\numitems$ (in the new sequence excluding item $1$) are independent from the absolute rank of item $1$ in the original sequence. Hence, 
\begin{align}
    \Prob\left(\event \intersect \left\{\invpermtrue(1) > \frac{59\numitems}{60}\right\}\right) = \Prob(\event) \cdot \Prob\left(\invpermtrue(1) > \frac{59\numitems}{60}\right) \stackrel{\stepone}{\ge} \frac{1}{2}\cdot \frac{1}{60}= \frac{1}{120},\label{eq:lb_inf_conditional_prob}
\end{align}
where step~\stepone is true by combining~\eqref{eq:lb_inf_num_greater_half} with the assumption that the true rank $\invpermtrue(1)$ of item $1$ is uniformly at random.
Now we consider the first term in~\eqref{eq:inf_lb_conditional_expression}. By the definition of event $\event$, there are at least $\frac{\numitems}{30}$ items with $\rankrelminusonetrue_\pos \ge (\frac{1}{2} + 2\noiseunifparam)(\pos+1)$, so that $\rankreltrue_\pos \ge \rankrelminusonetrue_\pos\ge (\frac{1}{2} + 2\noiseunifparam)(\pos+1)$. Conditional on the event $\{\varinsertpos_\pos = 1\}$, its observed score is bounded as
\begin{align*}
    \report_\pos = \frac{\rankreltrue_\pos}{\pos+1}+ \noise_\pos \stackrel{\stepone}{>} \frac{1}{2} + \noiseunifparam,
\end{align*}
where step~\stepone holds because the noise $\noise_\pos$ is bounded in $[-\noiseunifparam, \noiseunifparam]$.
Recall that the score of item $1$ is bounded in $[\frac{1}{2} - \noiseunifparam, \frac{1}{2} + \noiseunifparam]$, so conditional on $\event$, at least $\frac{\numitems}{30}$ items have their observed scores strictly higher than the observed score of item $1$, and thus the rank of item $1$ induced by the scores satisfies
$\estinvperminduced(1) \le \frac{29\numitems}{30}$.
Conditioning on the event $\event\intersect \left\{\invpermtrue(1) > \frac{59\numitems}{60}\right\}$, we thus have the pointwise relation
\begin{align}
    |\estinvperminduced(1) - \invpermtrue(1)| \ge \abs*{\frac{29\numitems}{30} - \frac{59\numitems}{60}} = \frac{\numitems}{60}. \label{eq:lb_inf_conditional_expect}
\end{align}
Substituting Eqs.~\eqref{eq:lb_inf_conditional_prob} and~\eqref{eq:lb_inf_conditional_expect} back to~\eqref{eq:inf_lb_conditional_expression}, we have
$\Expect\abs*{\estinvperminduced(1) - \invpermtrue(1)}\gtrorder \numitems$, as desired.
\qed

\subsection*{Acknowledgments}
JW was supported in part by the Ronald J. and Carol T. Beerman President’s Postdoctoral Fellowship and the ARC (Algorithms \& Randomness Center) Postdoctoral Fellowship at Georgia Tech. AP was supported in part by the National Science Foundation grants CCF-2107455 and DMS-2210734, and gifts/awards from Adobe, Amazon, and Mathworks. We thank Ramesh Johari, Cheng Mao, Juba Ziani, Yuqing Kong, and the anonymous referees for helpful comments and discussions, and the Simons Institute for the Theory of Computing for their hospitality when part of this work was performed.

\small
\bibliographystyle{plain}
\bibliography{references}

\normalsize
\appendix
\section{Auxiliary results}\label{app:proof_lem_expect_dist}

In this section, we collect a few standard results, and include their proofs for completeness.

\paragraph{Expectation of order statistics.}
The first standard result establishes the expectation of order statistics obtained from sampling integers without replacement.
\begin{lemma}\label{lem:expect_dist}
    Consider the set of the natural numbers $[n] = \{1, \ldots, n\}$.  Suppose we select $k$ numbers uniform at randomly without replacement from $[n]$, and denote the order statistics of these numbers as $X^{(1)} < \ldots < X^{(k)}$. Then for each $r\in [k]$, we have
  \begin{align*}
         \Expect[X^{(r)}] = \frac{n+1}{k+1} \cdot r.
     \end{align*}
\end{lemma}
The proof of this lemma is provided in Appendix~\ref{sec:proof_lem_expect_dist}.
Building on top of Lemma~\ref{lem:expect_dist}, we have the following lemma involving certain random quantities under a true ranking $\invpermtrue$ that is sampled uniformly at random. Recall from Section~\ref{sec:theory_noisy} that $\mapreltoabs^\pos_\rankrel(\invperm)$ denotes the absolute rank of the $\rankrel^\textth$ largest item (among the first $\pos$ items) according to the ranking $\invperm$. 
\begin{lemma}\label{lem:expect_dist_perm}
    Consider the non-parametric model~\eqref{eq:model_nonparametric}. Let $\pos\in [\numitems]$ be any timestep. Let $\funcrank$ and $h$ be deterministic functions of the tuple $(\pos, \noiseunifparam, \rankreltrue_{1:\pos}, \noise_{1:\pos})$. Suppose that the true ranking $\invpermtrue\in \setperms_\numitems$ is sampled uniformly at random. Then
    \begin{align*}
        \Expect_{\invpermtrue}[\mapreltoabs^\pos_\funcrank(\invpermtrue)\cdot h] =\frac{\numitems+1}{\pos +1}\cdot  \Expect_{\rankreltrue_{1:\pos},\noise_{1:\pos}} \left[\funcrank\cdot h\right].
    \end{align*}
\end{lemma}
The proof of this lemma is provided in Appendix~\ref{sec:proof_lem_expect_dist_perm}.

\paragraph{Relation between ranking metrics.}
Recall from~\eqref{eq:dist_inv_vec} in Section~\ref{sec:proof_notation} that $\lossinvvec$ denotes the (normalized) $\ell_1$ distance between the inversion vectors of two rankings. The following result from~\citet{jiang2010modulation} provides a relation between $\lossinvvec$  and $\losskt$, the Kendall--tau distance.
\begin{lemma}[Corollary 9 of~\citealp{jiang2010modulation}]\label{lem:kt_vs_rankrel}
For any two rankings $\invperm_1$ and $\invperm_2$, we have the relation
\begin{align*}
    \lossinvvec(\invperm_1, \invperm_2) \le \losskt(\invperm_1, \invperm_2).
\end{align*}
\end{lemma}
The proof of this lemma is provided in Appendix~\ref{sec:proof_lem_kt_vs_rankrel}. 

\subsection{Proof of Lemma~\ref{lem:expect_dist}}\label{sec:proof_lem_expect_dist}

We use the following generative procedure to sample the order statistics. We consider $\numitems$ balls, consisting of $k$ red balls and $(\numitems-k)$ blue balls. We order the balls in a line according to an ordering sampled uniformly at random. Then the order statistics $X^{(r)}$ is equivalent to the position of the $r^\textth$ red ball.

Equivalently, we start with $k$ red balls, and then insert the $(\numitems-k)$ blue balls into the sequence one-by-one. The first blue ball is inserted uniformly at random to the $(k+1)$ possible positions; the second blue ball is inserted uniformly at random to the $(k+2)$ possible position, and so on. Each insertion of a blue ball is independent from all other insertions.

Consider the final sequence after all blue balls are inserted. For each $r\in [1, k-1]$, we denote the number of blue balls between the $r^\textth$ and the $(r+1)^\textth$ red balls as the random variable $L_r \defn X^{(r+1)}-X^{(r)}-1$. Moreover, we denote $L_0 \defn X^{(1)} - 1$ as the number of blue balls to the left of the first red ball, and denote $L^{(k)} = n - X^{(k)}$ as the number of blue balls to the right of the $k^\textth$ (last) red ball. By symmetry of the insertion process, we have
\begin{align}
    \Expect[L_0] = \Expect[L_1] = \cdots = \Expect[L_k].\label{eq:order_stats_expect_equal}
\end{align}
Moreover, we have $(\numitems-k)$ blue balls in total. That is,
\begin{align}
    \sum_{i=0}^k L_i = n-k.\label{eq:order_stats_sum}
\end{align}
Combining~\eqref{eq:order_stats_expect_equal} and~\eqref{eq:order_stats_sum} yields
$\Expect[L_0]= \Expect[L_1] = \cdots = \Expect[L_k] = \frac{n-k}{k+1}$.
Hence, the expected position of the $r^\textth$ red ball is
\begin{align*}
    \Expect[X^{(r)}] = r + \sum_{i=0}^{r-1} \Expect[L_i] = r + r\cdot \frac{(n-k)}{k+1} = r\cdot \frac{n+1}{k+1},
\end{align*}
completing the proof.
\qed

\medskip 

\subsection{Proof of Lemma~\ref{lem:expect_dist_perm}}\label{sec:proof_lem_expect_dist_perm}

For the purposes of this proof, for any true ranking $\invpermtrue$ sampled uniformly at random, it is useful to think of the absolute ranks of its first $\pos$ items, $\{\invpermtrue(1), \ldots, \invpermtrue(\pos)\}$ as sampled using the following procedure, where the steps~\ref{step:sample_noise}-\ref{step:sample_rankrel} are executed independently of one another: 
\begin{enumerate}[label=(\arabic*)]
    \item \label{step:sample_noise}
    Sample the noise terms $\noise_{1:\pos}$ i.i.d. from $\Uniform[-\noiseunifparam, \noiseunifparam]$;
    \item \label{step:sample_val}
    Sample $\pos$  numbers from $[\numitems]$ uniformly at random without replacement, and denote them by $X^{(1)} < \ldots < X^{(\pos)}$;
    \item \label{step:sample_rankrel}
    Sample the relative ranks $\rankreltrue_{1:\pos}$ of the $\pos$ items independently from one another, where for each $\idxitem\in [\pos]$, the relative rank $\rankreltrue_\idxitem$ is sampled uniformly at random from $[\idxitem]$;
    \item From the relative ranks $\rankreltrue_{1:\pos}$, we obtain the ranking $\invpermtruerestrict{\pos}$ restricted to the first $\pos$ items. Then we set
    \begin{align*}
        \invpermtrue(\idxitem) = X^{(\invpermtruerestrict{\pos}(\idxitem))} \qquad \text{for each }\idxitem\in [\pos].
    \end{align*}
\end{enumerate}

It can be verified that marginally for any fixed $\pos$, the distribution of the tuple $(\invpermtrue(1), \ldots, \invpermtrue(\pos))$ under a uniform ranking $\invpermtrue$ is identical to the distribution of the tuple $(\invpermtrue(1), \ldots, \invpermtrue(\pos))$ obtained by this alternative sampling procedure.

Recall that $\mapreltoabs^\pos_\rankrel(\invperm)$ denotes the absolute rank of the $\rankrel^\textth$ largest item (among the first $\pos$ items) according to $\invperm$. Using this alternative sampling procedure, we have
\begin{align*}
    \mapreltoabs^\pos_\funcrank(\invpermtrue) =\invpermtrue({\invpermtruerestrict{\pos}}^{-1}(\funcrank)) = X^{\left(\invpermtruerestrict{\pos} {\invpermtruerestrict{\pos}}^{-1} (\funcrank)\right)} = X^{(\funcrank)}.
\end{align*}
By the law of iterated expectation, we have
\begin{align*}
    \Expect_{\invpermtrue}\left[\mapreltoabs^\pos_\funcrank(\invpermtrue)\cdot h \right] & \stackrel{\stepone}{=} \Expect_{\rankreltrue_{1:\pos},\noise_{1:\pos}}\;  \Expect_{X^{(1:\pos)}}\left[ X^{(\funcrank)}\cdot h \;\middle|\; \noise_{1:\pos}, \rankreltrue_{1:\pos} \right]\\
    & \stackrel{\steptwo}{=} \Expect_{\rankreltrue_{1:\pos},\noise_{1:\pos}} \; \left[ \Expect_{X^{(1:\pos)}}\left[X^{(\funcrank)}\;\middle|\; \noise_{1:\pos}, \rankreltrue_{1:\pos}\right]\cdot h  \right]\\
    & \stackrel{\stepthree}{=}\Expect_{\rankreltrue_{1:\pos},\noise_{1:\pos}}\;\left[ \funcrank\cdot \frac{\numitems+1}{\pos+1}\cdot h \right].
\end{align*}
where step~\stepone holds due to the independence among steps~\ref{step:sample_noise}-\ref{step:sample_rankrel} in the sampling procedure; step~\steptwo is true by the assumption that $h$ is a deterministic function of $(\rankreltrue_{1:\pos},\noise_{1:\pos})$; step~\stepthree is true by invoking Lemma~\ref{lem:expect_dist} and using the assumption that $\funcrank$ is a deterministic function of $(\rankreltrue_{1:\pos},\noise_{1:\pos})$.
\qed

\subsection{Proof of Lemma~\ref{lem:kt_vs_rankrel}}\label{sec:proof_lem_kt_vs_rankrel}
We call the $\numitems$-dimensional vector $\left(\invperm^{-1}(1), \ldots, \invperm^{-1}(\numitems)\right)$ as the inverse ranking of $\invperm$. The (unnormalized) Kendall--tau distance between two rankings $\invperm_1$ and $\invperm_2$ is equivalently defined as the minimum number of pairwise adjacent transpositions required to bring the inverse ranking $\left(\invperm_1^{-1}(1), \ldots, \invperm_1^{-1}(\numitems)\right)$ to $\left(\invperm_2^{-1}(1), \ldots, \invperm_2^{-1}(\numitems)\right)$~\citep[Eq. 1.4]{diaconis1977disarray}. It remains to show that the $\ell_1$ distance between the inversion vectors $\rankrelvec(\invperm_1)$ and $\rankrelvec(\invperm_2)$ is less than or equal to the this number of pairwise adjacent transpositions.

Consider any $\rankrel\in [\numitems-1]$. An adjacent transposition between the positions $\rankrel$ and $(\rankrel+1)$  on an inverse ranking $\left(\invperm^{-1}(1), \ldots, \invperm^{-1}(\numitems)\right)$ is equivalent to the following operations on the ranking $\invperm$:
\begin{enumerate}[label=(\arabic*)]
    \item We find the positions of the items of ranks $\rankrel$ and $(\rankrel+1)$ in the ranking $\invperm$. Namely, we find $\pos\defn \invperm^{-1}(\rankrel)$ and $\posalt \defn \invperm^{-1}(\rankrel+1)$.\label{item:equiv_operation_find}
    \item In the ranking $\invperm$, we set $\invperm(\pos)\leftarrow \rankrel+1$ and $\invperm(\posalt)\leftarrow \rankrel$.\label{item:equiv_operation_flip}
\end{enumerate}
    
Now consider the (unnormalized) $\ell_1$ distance between the inversion vectors. We track this $\ell_1$ distance when executing each pairwise adjacent transposition. Let $\pos^*\defn \max\{\pos, \posalt\}$. By the equivalent operations~\ref{item:equiv_operation_find}-\ref{item:equiv_operation_flip} above, it can be verified that the relative rank of item $\pos^*$, namely $\rankrel_{\pos^*}(\invperm)$, changes by $\pm 1$ after the pairwise adjacent transposition, whereas the relative ranks of all the other $(\numitems-1)$ items remain the same. Hence, the $\ell_1$ distance between the inversion vectors is at most the minimum number of pairwise adjacent transpositions in the inverse rankings, namely the Kendall--tau distance.
\qed

\end{document}